# Analysis of proposed PDE-based underwater image enhancement algorithms

U. A. Nnolim

*Department of Electronic Engineering, University of Nigeria Nsukka, Enugu, Nigeria*

## Abstract

*This report describes the experimental analysis of proposed underwater image enhancement algorithms based on partial differential equations (PDEs). The algorithms perform simultaneous smoothing and enhancement due to the combination of both processes within the PDE-formulation. The framework enables the incorporation of suitable colour and contrast enhancement algorithms within one unified functional. Additional modification of the formulation includes the combination of the popular Contrast Limited Adaptive Histogram Equalization (CLAHE) with the proposed approach. This modification enables the hybrid algorithm to provide both local enhancement (due to the CLAHE) and global enhancement (due to the proposed contrast term). Additionally, the CLAHE clip limit parameter is computed dynamically in each iteration and used to gauge the amount of local enhancement performed by the CLAHE within the formulation. This enables the algorithm to reduce or prevent the enhancement of noisy artifacts, which if present, are also smoothed out by the anisotropic diffusion term within the PDE formulation. In other words, the modified algorithm combines the strength of the CLAHE, AD and the contrast term while minimizing their weaknesses. Ultimately, the system is optimized using image data metrics for automated enhancement and compromise between visual and quantitative results. Experiments indicate that the proposed algorithms perform a series of functions such as illumination correction, colour enhancement correction and restoration, contrast enhancement and noise suppression. Moreover, the proposed approaches surpass most other conventional algorithms found in the literature.*

## 1. Introduction

There are numerous algorithms for image enhancement in the literature and range from simple to complex. These algorithms may or may not combine various domains to accomplish the task of image enhancement. However, a greater majority of algorithms were initially developed for greyscale image enhancement. Presently, the huge explosion of digital, high-definition visual colour media has led to a relatively recent but active research into colour image processing. Thus, these current devices and the content they generate or process are mostly consumed by humans rather than machines. Thus, colour has become a very vital component that can no longer be discarded when applying Computer Vision algorithms.

Image enhancement algorithms usually work well for a narrow set of images based on their nature of formulation. For example, the linear, statistics-based contrast enhancement algorithms are fundamentally dependent on image histogram manipulation and other image statistics. Additionally, they are mostly suited to faded, grey, low contrast images. The algorithms based on the logarithmic image processing (LIP) framework utilize the multiplicative reflectance-illumination model for their operation and are suited to dark, under-exposed images or those with poor or uneven illumination. The filter-based algorithms perform mainly edge enhancement or sharpening due to their isotropic, localized gradient operations. However, they are also prone to enhancing intrinsic noise embedded within the image. This results in the visual manifestation of noisy artifacts around sharpened edges. Thus, the gain and cut-off frequency of the filters must be tuned to reduce or minimize this effect. This is easier to achieve in the frequency domain, which involves global operations using the Fourier transform. Adjusting such filters in the spatial domain is much harder and effects may vary from image to image due to the local nature of the enhancement.

In spite of all these developments, most of these algorithms do not perform adequately for most colour images and may require some additional processing to yield desired results. This is the case with several of the algorithms utilizing the LIP model such as Homomorphic filtering [1], Multiscale Retinex, (MSR) [2], etc. The application of these algorithms usually lead to faded colours or grey tint in the processed images, which violate the grey-world assumption of these algorithms. However, solutions to these problems have been proposed and include the processing of images in non-linear, perceptual colour spaces such as (Hue-Saturation-Value) HSV and (Hue-Saturation-Intensity) HSI colour coordinate systems [3]. Consequently, most new approaches to image enhancement such as Particle Swarm Optimization (PSO), Genetic Algorithm (GA), Wavelets, Dynamic Stochastic Resonance (DSR) [4] operate in the HSV colour spaces to yield adequate processed colour images.

The linear, statistics- and histogram-based contrast stretching algorithms mostly distort colours in processed RGB colour images due to their singular focus on contrast enhancement without taking into account colour channel pixel relationships when performed on each colour image channel. Hue-preserving algorithms such as histogram specification have also been proposed to reduce or eliminate this problem. However, due to the uniqueness of each new image, it is difficult to expect a consistent result for all images processed with the algorithm. Furthermore, there has been some interesting work on the utilization of Quaternion Fourier Transform for colour enhancement processing [5] [6], there are still relatively few works on colour image processing relative to greyscale image processing. However, most of the conventional algorithms for colour image processing do not usually perform well or have been tested in multiple application areas.

Moreover, most of the work in colour image enhancement also deals with natural images acquired on land. Consequently, there is relatively little published work on underwater image processing compared to land-based image processing. Thus it is not surprising that the amount of proposed algorithms for underwater colour image processing are also fewer compared to the ones on land-based image processing. However, some of the techniques used in the processing land-based images are also applied to underwater images. This is due to the similarities of some degradation processes that occur in both land-based and underwater images. Some of these degradations include blurring, poor or uneven illumination, colour cast effects due to the effects of underwater optics [7] [8]. It should be added that though work in this area is relatively small, it is growing gradually as evidenced by published works in the literature [7] [8] [9] [10] [11] [12] [13] [14] [15] [16] [17] [18] [19] [20] [21] [22] [23] [24] [25] [8] [26] [6] [27] [28] [29] [30] [31] [32].

The various algorithms and approaches are documented in [7] [8] and indicate that most of the results are visually evaluated rather than with quantitative metrics. Additionally, these algorithms are not shown to be tested with a wide variety of underwater images that could be captured in such aquatic environments. Though it is impossible to expect an image processing algorithm to resolve all the image degradation problems it is confronted with, it should at least yield reasonable results for a wide range of images within its application domain or ideally perform several image processing functions. Algorithms such as the CLAHE [33], MSR and Homomorphic filter achieve this to some extent, thus making them, their modifications and variations popular choices evidenced by the volume of available literature on them. However, these algorithms have limitations and are more or less fixed or constricted in the operation. This means that they mostly yield a particular result for a particular image all the time. Though this consistency is favoured when results are good, they can be limiting when results are not acceptable. For machine interpretation, the parameters can be optimized and fixed for numerical evaluation. However, for humans, the range of quality results is subtler and not absolute. Thus, there is need to devise an algorithm that can not only yield a particular result when required but also enable the user to decide the type of result that is acceptable.

Application of partial differential equations (PDEs) to image processing is now an established and ever-growing field. The seminal works by Perona and Malik on Anisotropic Diffusion (AD) [34], in addition to Rudin, Osher and Fatemi's work on Total Variation Regularization [35] and Shock filter [36] implementation of stabilized reverse-diffusion are well-known. Though these algorithms were designed for the filtering of Additive White Gaussian Noise (AWGN) and morphological operations, they have been adopted and applied to other areas in image processing research [37] [38].

In this work, several underwater image enhancement algorithm variants mainly composed of a contrast term and the AD term are utilized in the processing of underwater images is proposed. However, the developed algorithms are suited not just to underwater images but also land-based images. Based on experiments, the algorithms can adequately process several colour images suffering from similar degradations such as poor or uneven illumination, colour cast distortion, low contrast, fading colours or noise artifacts.

It should be noted that the incorporation of PDEs in image enhancement is not a new idea and relevant works can be found in [39]. However, in this work, we present several forms of the proposed framework in addition to the use of a contrast term based on the Probability Mass Function (PMF) [40] [41] of an image. This is in addition to the dynamic evaluation of image statistics used to guide some of the control parameters for the base algorithm utilized within the PDE framework. Additionally, the only PDE-based approaches similar to the base formulation proposed here for underwater images is that of [31].

### 1.1 Motivation for the proposed approach

The motivation for this work includes the development of an algorithm that would operate adequately on several images; colour and greyscale without resorting to colour space conversions. Additionally, the algorithm should be flexible, enabling the user to control not just the rate of processing but also the amount of processing. This is achieved in the PDE-based framework by regulating the extent or contributions of the various processes encapsulated within the framework using control parameters. Moreover, the addition of a noise suppression term eliminates or minimizes the visual effects of intrinsic noise enhancement associated with the more popular, histogram-based image contrast enhancement algorithms.

### 1.2 Anisotropic Diffusion

Anisotropic diffusion is obtained from the modification of the isotropic (equal energy in all directions) diffusion heat equation given as [34] [42];

$$\frac{\partial U(x,y,t)}{\partial t} = D\nabla U(x,y) \quad (1)$$

into an anisotropic form [34] [42], yielding the expression in (2);

$$\frac{\partial U(x,y,t)}{\partial t} = \nabla.(D(|\nabla U|)\nabla U(x,y)) \qquad (2)$$

In both equations (1) and (2), $D$ is the diffusion coefficient and $\nabla U(x,y)$ is the image gradient, though $D$ is a constant in (1) while it is a function of $|\nabla U|$ in (2). The functions proposed by Perona and Malik for $D$ in their work [34] are given as shown in (3) and (4).

$$D(|\nabla U|) = e^{-\left(\frac{|\nabla U|}{\kappa}\right)^2} \quad (3)$$

$$D(|\nabla U|) = \frac{1}{1+\left(\frac{|\nabla U|}{\kappa}\right)^2} \quad (4)$$

In (3) and (4), the parameter, $\kappa$ is the diffusion threshold parameter [34] [42]. There have been several proposed modifications of the standard approach over the years as mentioned in [37]. Due to the fact that the problem is ill-posed, the image gradient is obtained after smoothing (convolving) the image, $U(x,y)$, with a Gaussian smoothing kernel, $G_\sigma$, (with width or standard deviation, σ) in the form [43] shown in (5);

$$|\nabla U_\sigma| = |\nabla G_\sigma * U| \quad (5)$$

## 2. PDE formulation for image enhancement

The fundamental framework proposed by [39] [44] is described in this section for a continuous initial image, $U(x,y,t)$. Two processes, in this case, smoothing, $F_s(\,)$ and enhancement $F_e(\,)$functions can be defined in the PDE framework as in (6) and (7) while the combined process is shown in (8) with $\lambda$ as a control parameter that regulates the amount of smoothing with respect to enhancement.

$$\frac{\partial U(x,y,t)}{\partial t} = F_s\big(U(x,y,t)\big) \qquad (6)$$

$$\frac{\partial U(x,y,t)}{\partial t} = F_e\big(U(x,y,t)\big) \qquad (7)$$

$$\frac{\partial U(x,y,t)}{\partial t} = \lambda F_r\big(U(x,y,t)\big) + F_e\big(U(x,y,t)\big) \qquad (8)$$

For equation (6), we now define the actual function that defines the restoration process, $F_r\big(U(x,y,t)\big)$, which is the isotropic diffusion (ID) term. However, in this case, the ID term can also be reformulated in the form of the standard mean curvature motion equation [39] expressed as shown in (9);

$$\frac{\partial U(x,y,t)}{\partial t} = \|\nabla U(x,y)\| \mathrm{div}\left(\frac{\nabla U(x,y)}{\|\nabla U(x,y)\|}\right) = U(x,y)_{\xi\xi} \qquad (9)$$

In the expression in (9), ξ is the direction perpendicular to the image gradient, $\nabla U(x,y)$ parallel to the image edges [39] while *div* is the divergence operator. For the enhancement function, $F_e\big(U(x,y,t)\big)$, the PDE formulation from the literature is given as;

$$\frac{\partial U(x,y,t)}{\partial t} = f\big(U(x,y)\big) - U(x,y) \qquad (10)$$

The function, $f\big(U(x,y)\big)$ in (10) can be any contrast enhancement function though in the original formulation proposed by [39] it is a histogram modification or equalization (HE) transformation function. Other functions, both simple and complex, are employed by other authors to achieve contrast enhancement [45] [46] [47] [48] [6]-[9].

Thus, combining the two functions as before in (8), we obtain;

$$\frac{\partial U(x,y,t)}{\partial t} = \lambda \|\nabla U(x,y,t)\| \mathrm{div}\left(\frac{\nabla U(x,y)}{\|\nabla U(x,y)\|}\right) + f\big(U(x,y)\big) - U(x,y) \qquad (11)$$

However, as noted in the literature, the addition of an edge stopping function to the isotropic diffusion term leads to anisotropic diffusion [39]. Thus, the subsequent non-constant diffusion coefficient function, $D(\|\nabla U(x,y)\|)$ controls the degree of smoothing [39]. The complete generalized PDE framework for image enhancement now becomes;

$$\frac{\partial U(x,y,t)}{\partial t} = \lambda D(\|\nabla U(x,y)\|)\operatorname{div}\left(\frac{\nabla U(x,y)}{\|\nabla U(x,y)\|}\right) + f\big(U(x,y)\big) - U(x,y) \qquad (12)$$

This particular PDE-based contrast enhancement framework is utilized in enhancing images with non-uniform illumination and to improve contrast. Its features include better control and gradual enhancement in addition to simultaneous enhancement and de-noising processes [39].

However, despite this general advantage, the PDE-based formulation does not yield consistently good results for RGB colour images, due in part to the default weakness in the chosen contrast enhancement function (histogram modification in [39]). As mentioned before, most of these statistics-based algorithms focus primarily on contrast enhancement at the expense of colour relationships of the R, G and B channels. Their local isotropic nature of enhancement also results in enhancement of noise artifacts, which are visually manifested.

Additionally, most of the previous enhancement functions have little to no colour restoration ability to counter colour distortions normally encountered in underwater images. Experiments were performed using various histogram-based variants in the PDE formulation and results were not encouraging [49]. Based on the outcome of these experiments, alternatives were sought to add extra features in the PDE-based enhancement framework. The devised approach should maximize the advantages of the combined processes while minimizing or suppressing their individual weaknesses.

## 3. Proposed PDE model

With the generalized model established in section 2, we now present the proposed algorithm utilizing this model in detail. The first proposed model is different from previous approaches by not utilizing a fidelity term, though more control is added to regulate the contrast enhancement in certain cases. This in addition to a modified, more image feature-sensitive contrast enhancement term with more control.

### 3.1 Selection of contrast enhancement function

Prior to settling on the chosen contrast term and parameters, experiments were performed utilizing several approaches and a sample of results are shown in Fig. 1. A series of experiments were performed to obtain a suitable and relatively stable contrast enhancement function that can gradually and slowly converge to a steady state value. The function must not be too fast or unstable in order to suit our purposes. Based on previous work and experiments [50], the candidate algorithms ranged from the simple contrast stretching functions to complex CLAHE algorithms. The results of the various algorithms in Fig. 1 indicate that the contrast stretching approaches appear to be promising and this is also observed in underwater image enhancement algorithms from the literature [9] [17] [24] [27].

### 3.2 Formulation of global image enhancement algorithm (PA-1)

Based on the assumption of a considerably decelerated contrast enhancement term, the PDE-based contrast enhancement process is reformulated as;

$$\frac{\partial U(x,y,t)}{\partial t} = f\big(U(x,y)\big) \qquad (13)$$

And combining the two processes, leads to the basic formulation of the proposed system given as;

$$\frac{\partial U(x,y,t)}{\partial t} = \lambda D(\|\nabla U(x,y)\|)\operatorname{div}\left(\frac{\nabla U(x,y)}{\|\nabla U(x,y)\|}\right) + f\big(U(x,y)\big) \qquad (14)$$

In this case, $f\big(U(x,y)\big)$, which maps the image to its new dynamic range is given as;

$$f\big(U(x,y)\big) = \frac{U(x,y) - \mu}{\sigma} \qquad (15)$$

Where in (15), $\mu$ is the global mean while $\sigma$ is the global standard deviation of the image, $U(x,y)$. Subsequently, the equation in (14) represents the total flow that minimizes the expression similar to the one in [18] as;

$$\lambda \int_{\Omega} \|\nabla U(x,y)\| dxdy + E\big(U(x,y)\big) \quad (16)$$

In (16), Ω is the image domain while the energy [18] is given by; $E\big(U(x,y)\big) = \frac{\Omega}{\sigma\mu}\int_{\Omega}\left[U(x,y) - \frac{\mu}{\sigma}\right]^2$ dxdy. In equations (8), (11) to (14), λ is the balance factor and it controls the degree of restoration by regulating the rate of processing for time step, *dt*. Discretizing the expression in (16), the implementation becomes;

$$U(x,y,t+1) = U(x,y,t) + \left[\lambda\left(D(\|\nabla U(x,y)\|)\mathrm{div}\left(\frac{\nabla U(x,y)}{\|\nabla U(x,y)\|}\right)\right)\right]\Delta t + f\big(U(x,y)\big) \quad (17)$$

The expression in (17) is designated as algorithm 1 (PA-1) and the main features of the proposed PDE-based enhancement algorithm include colour correction, colour and contrast enhancement in addition to illumination correction/normalization and noise suppression.

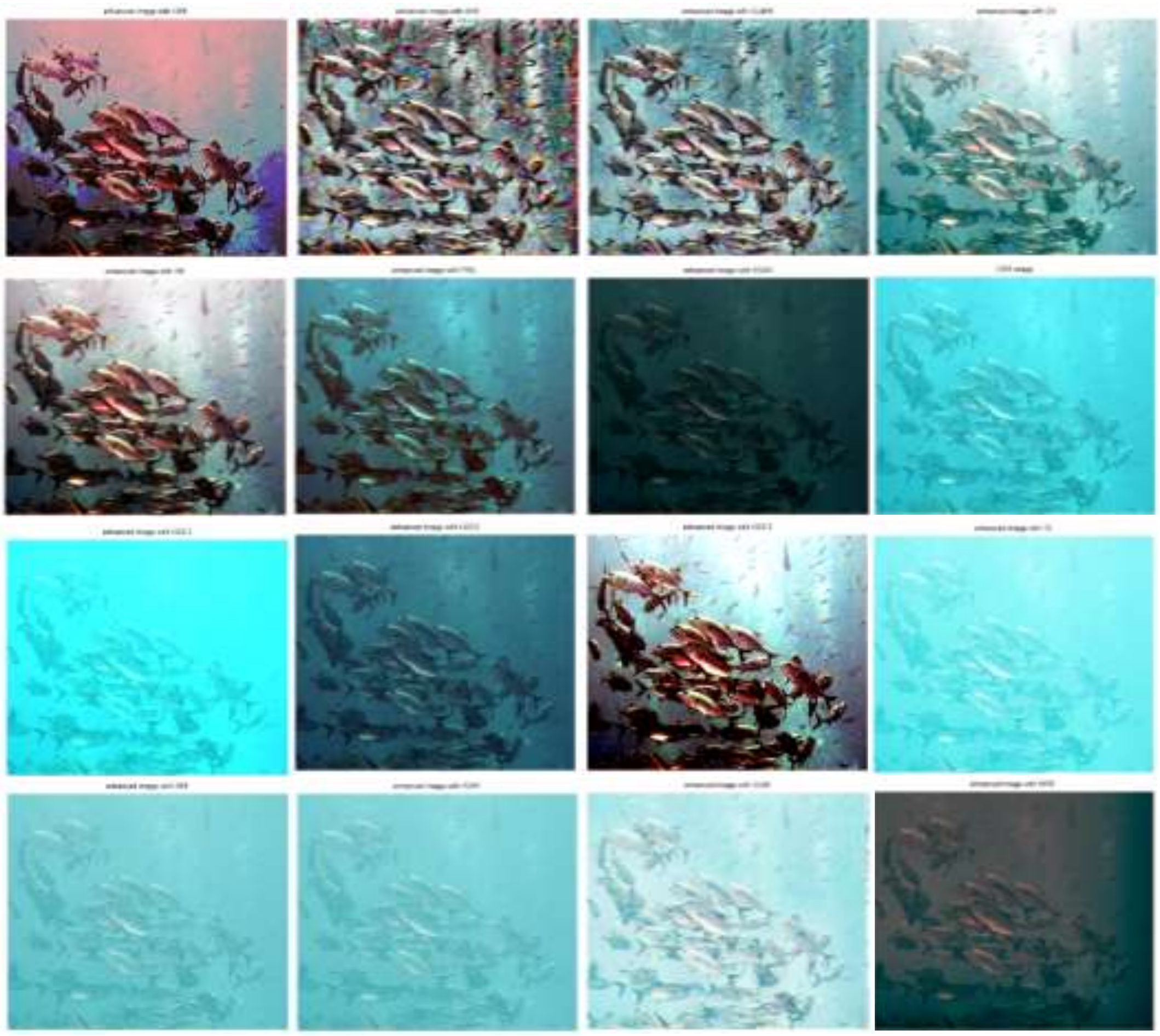

(a)

| | | | |
|---|---|---|---|
| GHE | AHE | CLAHE | CS |
| HS | PWL | SSAR | DSR |
| GOC1 | GOC2 | GOC3 | TC |
| SHF | FDHF | GUM | MSR |

(b)

**Fig. 1** (a) fish image processed with various algorithms (b) key to figures

### 3.2 Further improvements and additions to the model

Subsequent experiments performed to evaluate the performance of the proposed algorithm led to some observations. Firstly, the algorithm is harsh on certain images and there is a sharp contrast transition rather than the gradual changes observed in other images. Additionally, there is inadequate colour correction and in fact at times colour distortion occurs where processed images acquired a greenish tint.

Extensive adjustment of parameters did not yield much improvements in the visual results for some images. It was discovered that the mean is unreliable as a parameter in this case to process the images and subsequent experiments led to the adoption of the mode parameter. This appears to be related to the issue that modal statistics reveal information about the probability mass function (PMF) of the image pixel intensity distribution [40] [41]. Since the modal value is much more stable than the mean, it is observed that contrast changes are much more balanced than when the mean is utilized, since the latter is affected by outliers.

Additionally, numerical experiments concerning the mean and mode-based contrast terms were performed to ascertain the reasons for obtaining better results with the mode than the mean parameter. Another analysis of the transfer function profile of the GOC3 compared with eliminating the offset is shown to understand how to modify the term to further reduce this grey effect. The comparison between the gain offset correction and the mean and mode-based contrast term transfer functions are shown in Fig. 2. It is observed that there are a lot of values that are negative without the offset and these values will be set to zero when being displayed. Even with the offset, there are still some values below zero and will be lost when primed for display. The usage of the modal value shifts all values to positive. However, a large number of values exceed the maximum allowed value of 1. Thus this means that a large number of values will be set to 1, leading to oversaturation. However, this is less of a problem since the mode is fairly constant with increasing iterations. Using both modal and mean values in the proposed algorithm (PA-1), we plot the results over several iterations for various images. The results are shown in Fig.3. Experimental results appear to support this assumption as the mean and mode are plotted for each iteration showing their long-term stability.

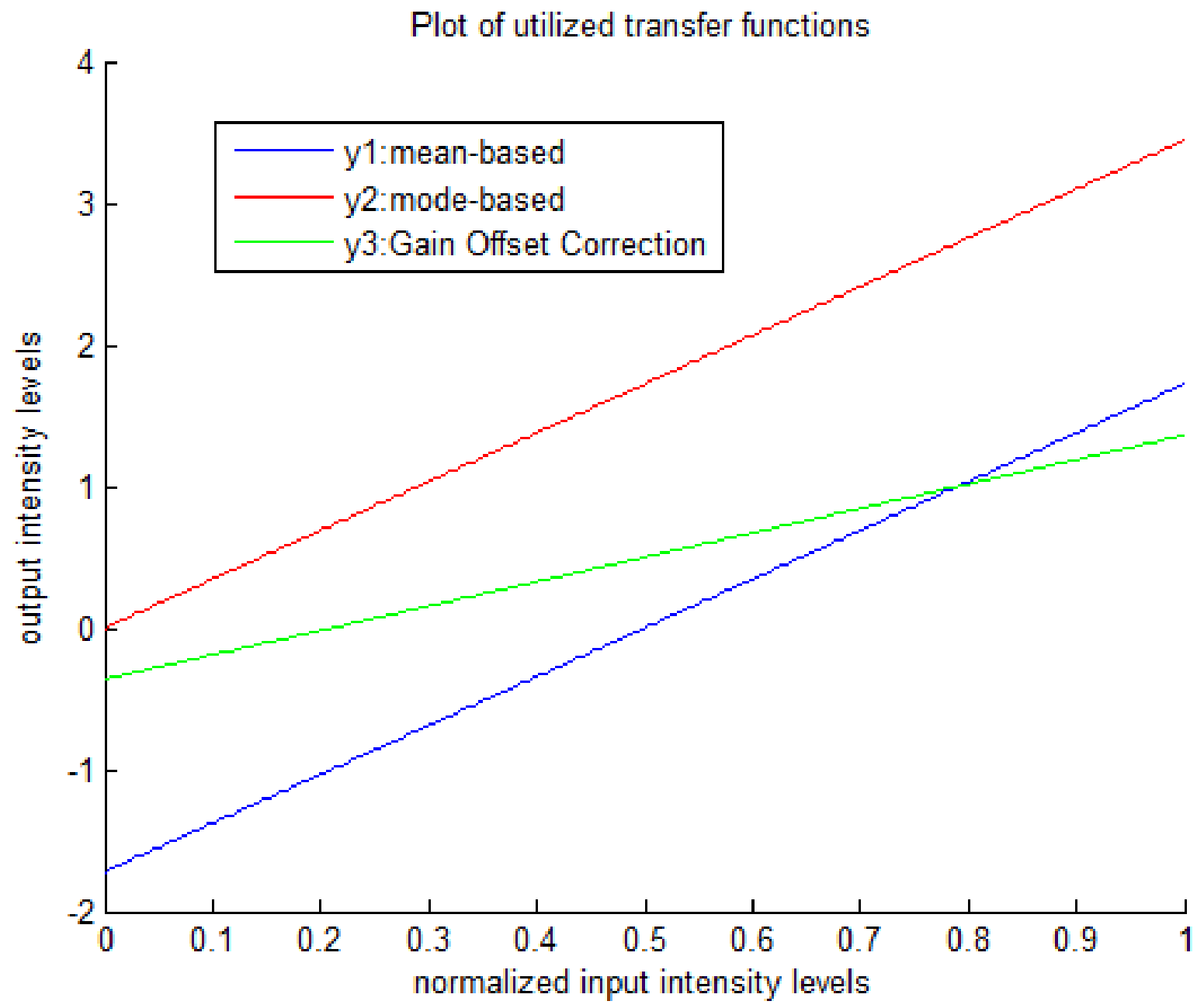

**Fig. 2** Transfer function profile of the modifications of the GOC3 contrast enhancement function

Observing the results in Fig. 3, the mode is either constant or linearly increasing or decreasing whether the modal or mean value is used for the contrast term in PA-1. Conversely, the mean is erratic when the mean value is used in the contrast term or PA-1 and is linearly increasing or decreasing when the modal value is used. Thus, using mode in the contrast term makes the system much more predictable and enables the usage of the contrast term without a fidelity term in the formulation for PA-1. Consequently, the modified term for the PDE contrast enhancement is now given as;

$$f\big(U(x,y)\big) = \frac{U(x,y)-m}{\sigma} \qquad (18)$$

In the modified expression in equation (18) for $f\big(U(x,y)\big)$, $m$ is the mode of the image. We compare visual results using both terms for processing the sea plants image as shown in Fig. 4. The modal value gives better colours though there is darkening due to drop in mean brightness. Conversely, the mean value yields brighter images due to increase in mean brightness, leading to faded colours.

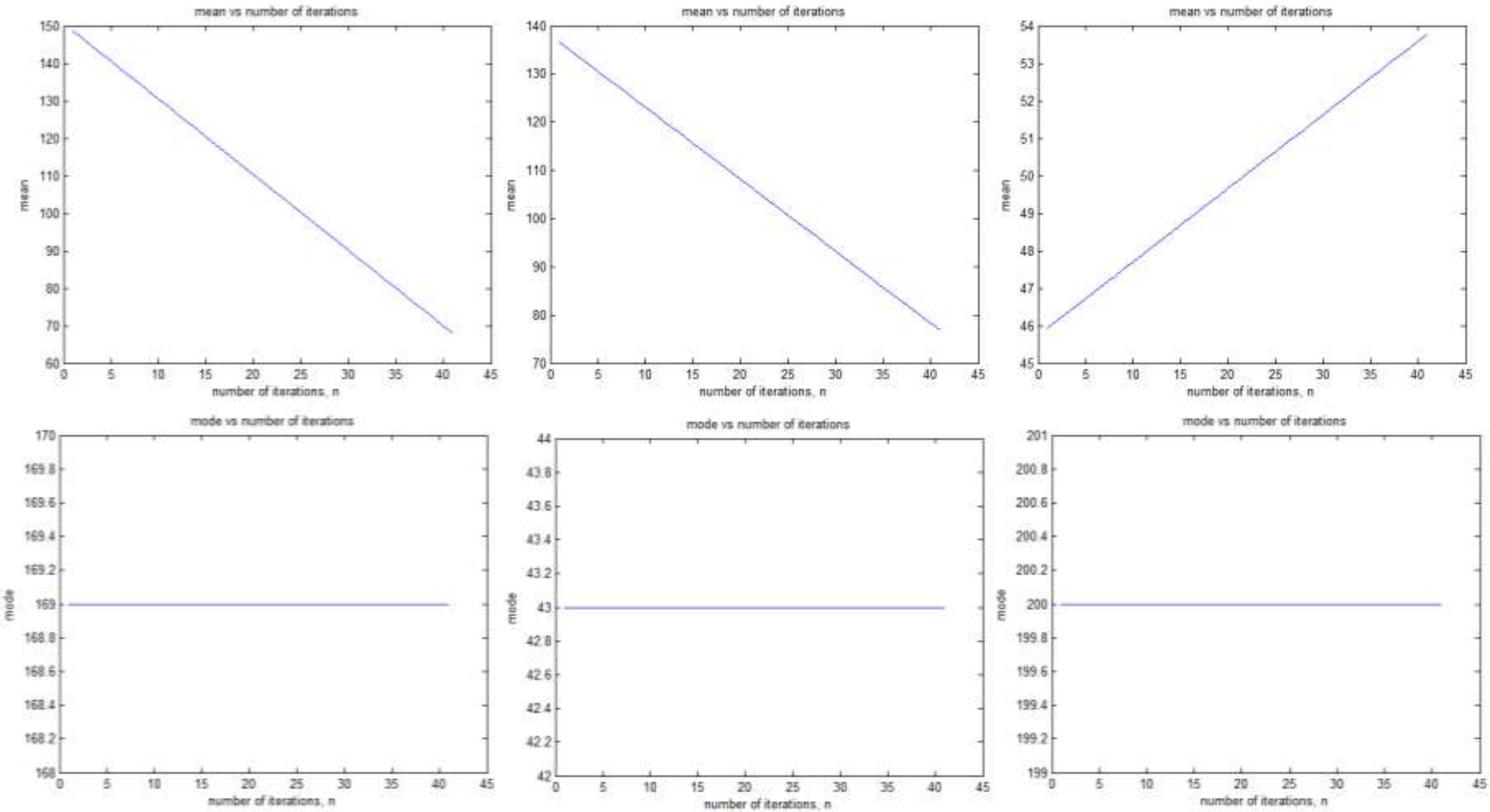
mean vs number of iterations
mean
number of iterations, n
mean vs number of iterations
mean
number of iterations, n
mean vs number of iterations
mean
number of iterations, n
mode vs number of iterations
mode
number of iterations, n
mode vs number of iterations
mode
number of iterations, n
mode vs number of iterations
mode
number of iterations, n

(a)

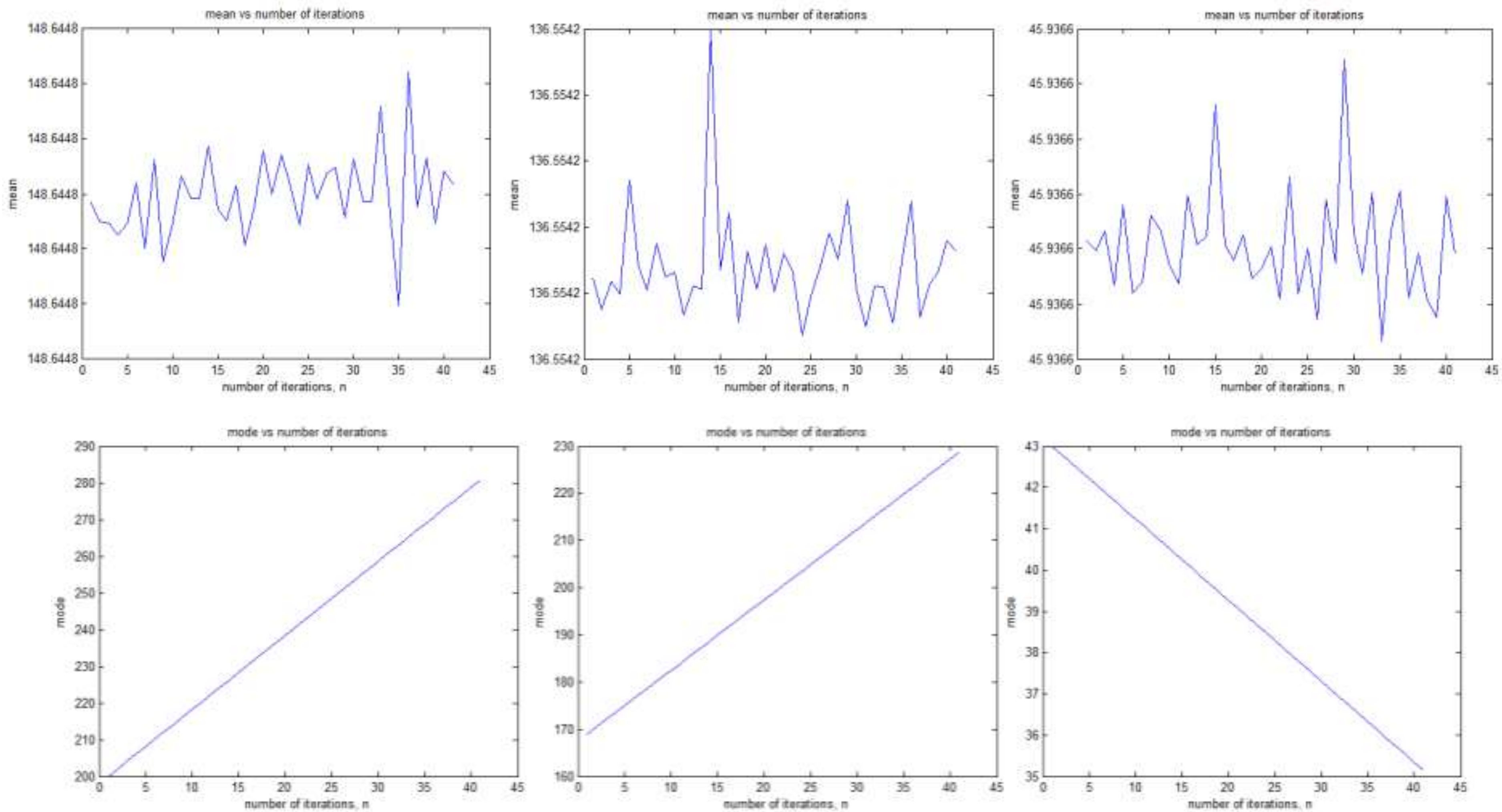

(b)

**Fig. 3** Mean and mode plots of *sea plants* image when using (a) modal value and (b) mean value for contrast term in PA-1

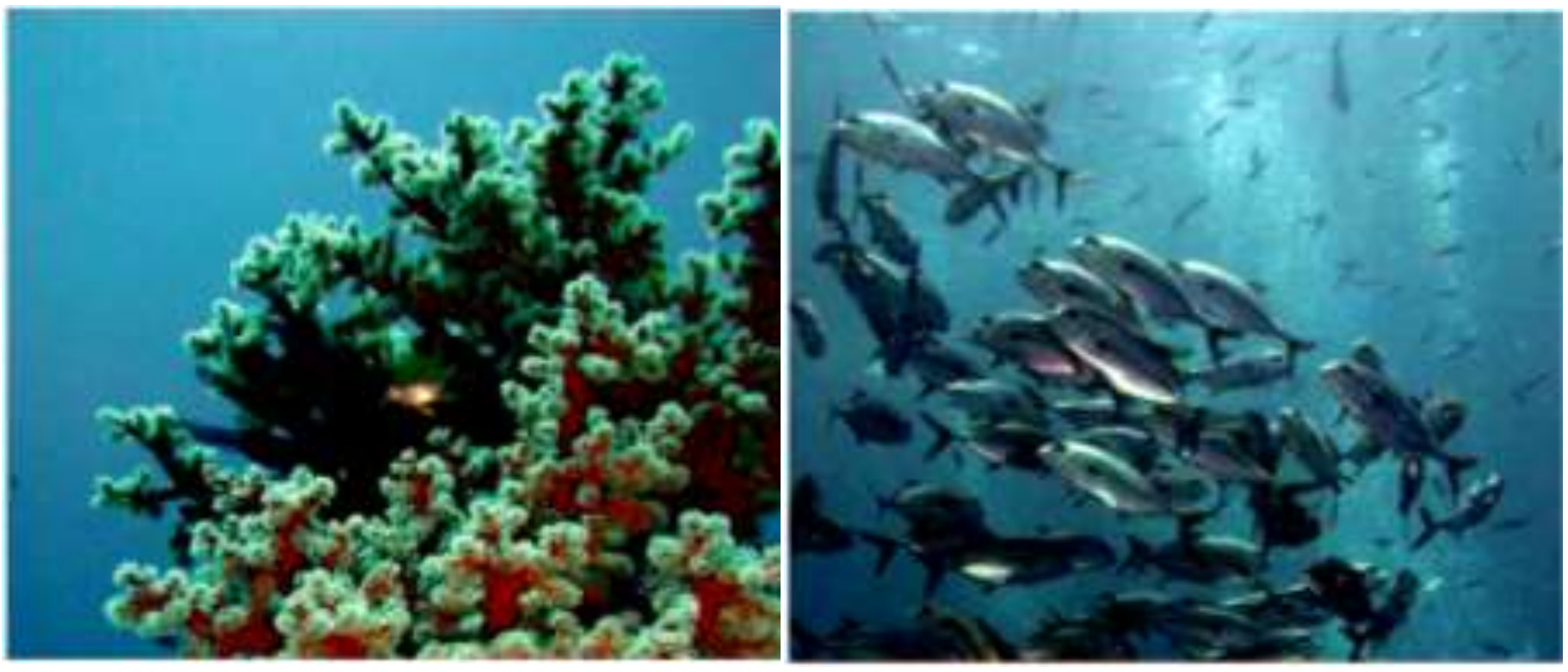

(a)

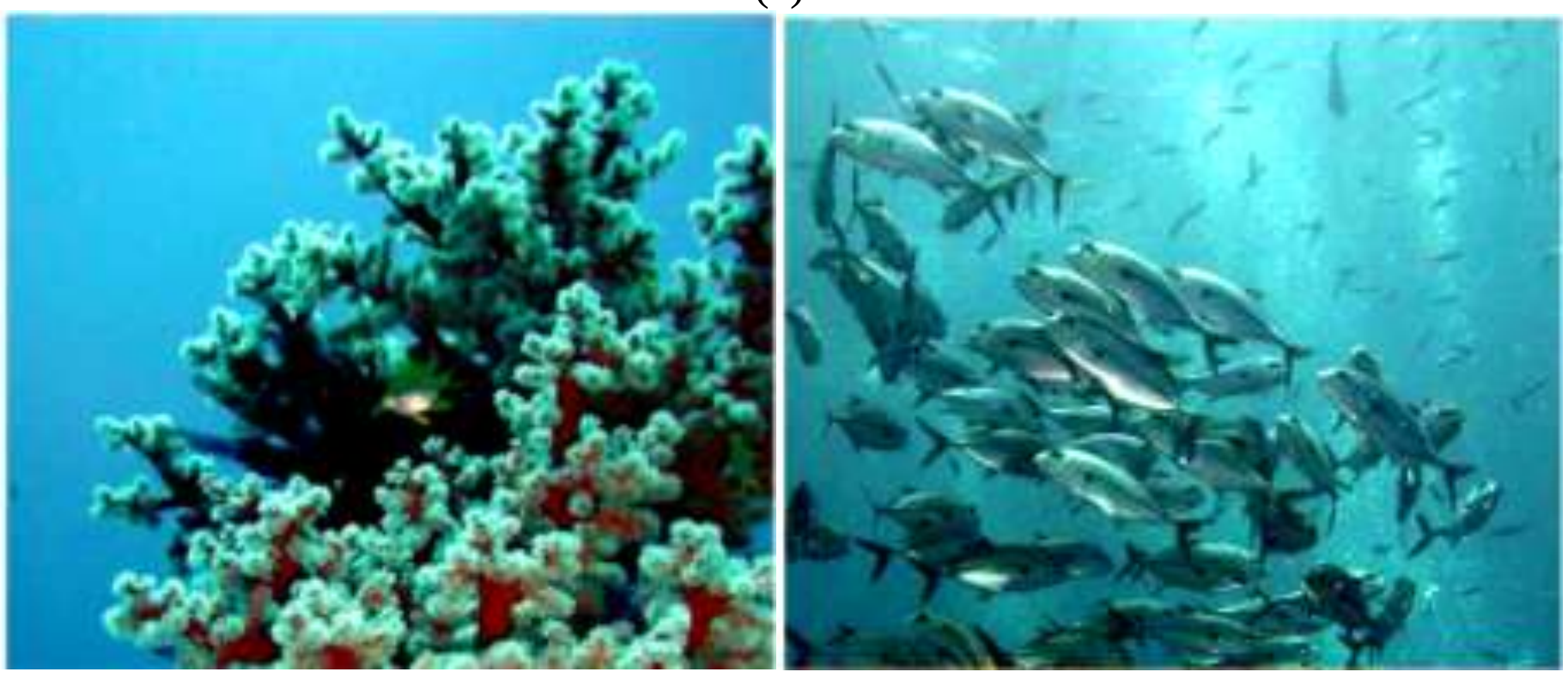

(b)

**Fig. 4** Results for *sea plants* image processed with PA-1 using (a) modal and (b) mean value for contrast term

However, though results are improved, it is observed that there is over-exposure in the bright regions of certain images, leading to loss of detail as a result of global contrast enhancement. Thus, it would be better to perform some sort of local contrast enhancement along with the global process. Thus, the next proposal would be to combine the effects of the CLAHE method, with PA-1. However, the CLAHE has certain problems such as colour distortions and increase in intrinsic noise enhancement. Thus, this addition would be meaningful if it enhances the strengths of the CLAHE while minimizing its weaknesses. This could be done by controlling the clip limit parameter of the CLAHE algorithm. Thus the combined PDE-based CLAHE and the proposed algorithm contrast enhancement process yields the formulation shown in (19);

$$\frac{\partial U(x,y,t)}{\partial t} = \lambda D(\|\nabla U(x,y)\|)\operatorname{div}\left(\frac{\nabla U(x,y)}{\|\nabla U(x,y)\|}\right) + f_g\big(U(x,y)\big) + f_l\big(U(x,y)\big) - U(x,y) \qquad (19)$$

In (19), $f_g\big(U(x,y)\big)$ is the global contrast operator while $f_l\big(U(x,y)\big)$ is the local contrast operator and the discretization leads to the following expression in (20);

$$U(x,y,t+1) = U(x,y,t) + \left[\lambda\left(D(\|\nabla U(x,y)\|)\operatorname{div}\left(\frac{\nabla U(x,y)}{\|\nabla U(x,y)\|}\right)\right) + f_l\big(U(x,y)\big) - U(x,y) + f_g\big(U(x,y)\big)\right]\Delta t \qquad (20)$$

The expression in (20) is designated as PA-2 and it enables better control of the effects of the CLAHE as shown in Fig. 5, where PA-2 yields more enhanced detail in 5(f). The image in Fig. 5(b) shows both overexposure and colour distortion due to over brightness using PA-1. Using CLAHE alone yields good local contrast enhancement but still showing hazy regions where details are still not properly enhanced, even with reduction of tile size and increase in clip limit parameter. However, there is still colour distortion in the results of PA-2 due to the CLAHE, whose effects still need further control.

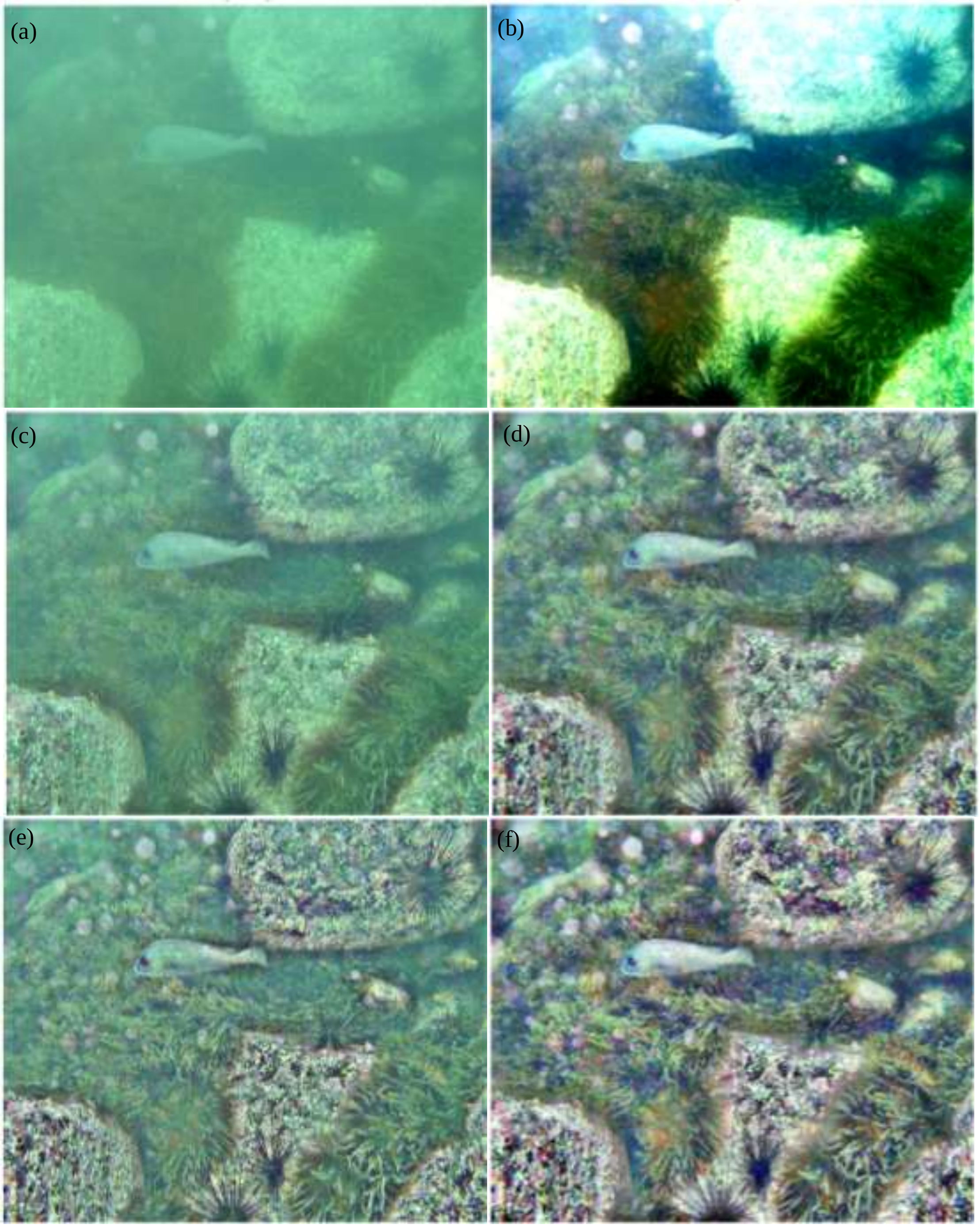

**Fig. 5** (a) Original image processed with (b) PA-1, (c) CLAHE (clip limit = 0.03, tile size = 64, bins = 256), (d) CLAHE (clip limit = 0.012, tile size = 64, bins = 256), (e) CLAHE (clip limit = 0.03, tile size = 32, bins = 256) (f) PA-2 (with clip limit = 0.012, tile size = 64, bins = 256)

### 3.3 PDE-based multi-scale algorithm with adaptive clip limit computation

On observation of results for images processed with algorithm 2 (PA-2), it is not possible to dynamically modify the clip limit since it is usually fixed, even in the conventional PDE-CLAHE algorithm. The proposed scheme in this section enables the adaptive computation of the clip limit within the CLAHE in algorithm 2 or each iteration. The measures of the variance, $\sigma^2$ and mean, $\mu$ are utilized in the formulation of the expression for the dynamic computation of the clip limit as;

$$clip\ limit = \frac{\mu}{\sigma^2} \quad (21)$$

This expression for the clip limit in (21) leads to interesting results when values are computed for each iteration. Using this information about the clip limit, we can use it as a guide to control or regulate the amount of enhancement further within the CLAHE and addition to the control parameters in the PDE framework. Further experiments were performed to observe the nature of this modification using underwater images.

**3.3.1 Experiments involving dynamic clip limit computation for evolving image**

In this section we justify the approach based on dynamic clip limit computation to enable it perform better than the fixed version of the basic algorithm 2. We optimize the algorithm using dynamic clip limit based on entropy maximization. The images tested are shown in Fig. 6A while their respective entropy and dynamic clip limit profiles results are shown in Fig. 6B. Observing the plots for various images, it becomes clear that the dynamic clip limit value stabilizes after a certain number of iterations whether the image has a monotonically increasing or decreasing function profile. The end point is similar in almost all cases. Thus, this tallies with experimental results where best results are obtained with the CLAHE using clip limits of about 0.01. Thus, for good results, the clip limit should not exceed the range of 0.03 or lower than 0.01. This is implemented within the proposed algorithm that utilizes the adaptive clip limit term to stabilize results and to reduce number of iterations.

Thus, the result is the same almost every image tested, indicating consistency. However, some images may converge quicker than others and some will be continuously enhanced while others will stabilize.

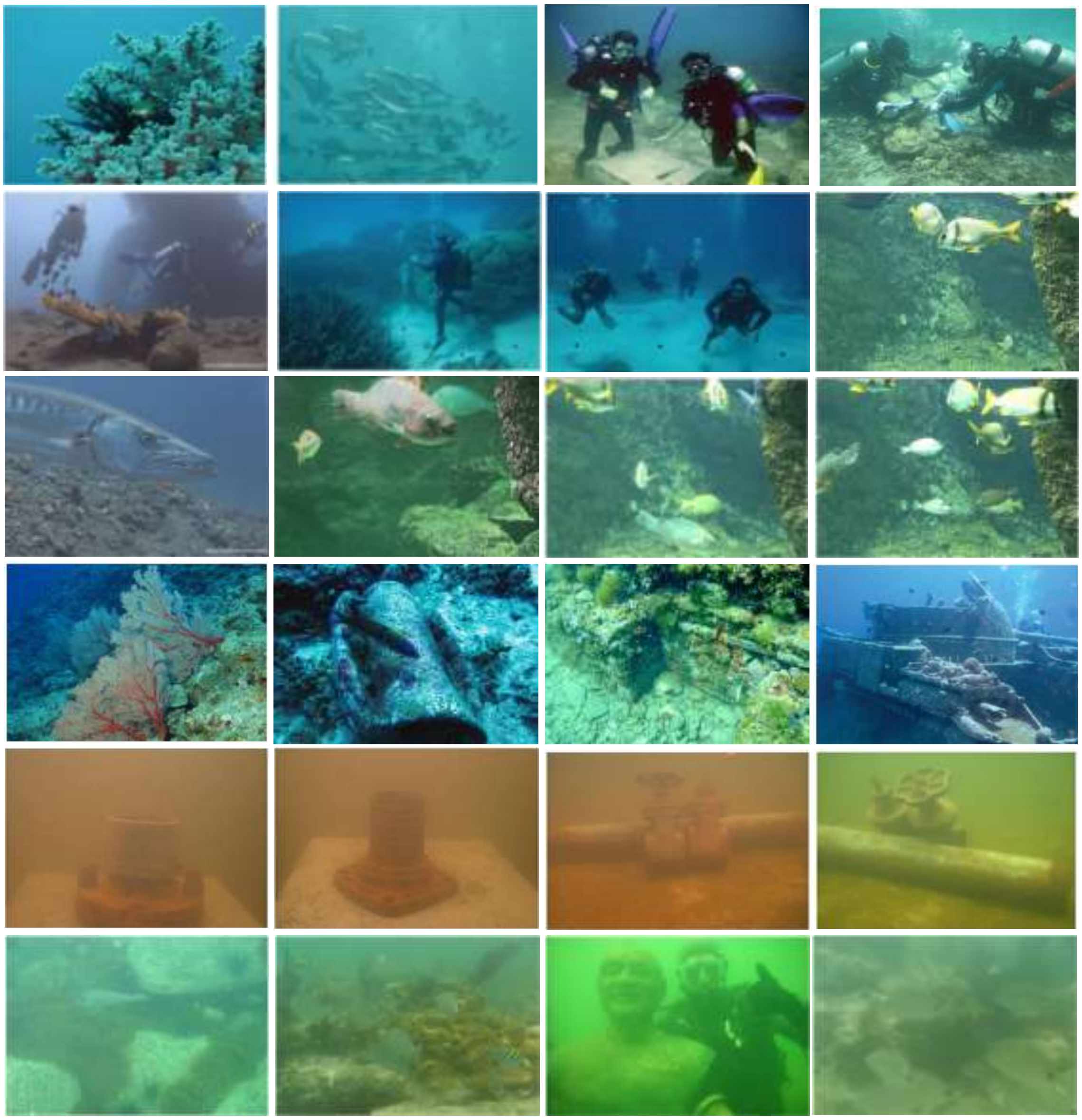

(a)

| | | | |
|---|---|---|---|
| (a) sea plants | (b) fishes1 | (c) divers1 | (d) divers2 |
| (e) divers3 | (f) diver | (g) divers4 | (h) fishes2 |
| (i) barracuda | (j) fishes3 | (k) fishes4 | (l) fishes5 |
| (m) seaplants2 | (n) ocean jar | (o) ocean floor | (p) ship wreck |
| (q) object1 | (r) object2 | (s) object3 | (t) object4 |
| (u) various | (v) fishes6 | (w) statue | (x) fishes7 |

**KEY**

(b)

**Fig. 6A** (a) Underwater images used in experiments for dynamic clip limit evaluation (b) key to figures

Based on visual results, we discover that the best settings for number of iterations is between 3 to 6 iterations using the dynamically computed clip limit for the CLAHE in the modified PA-2 algorithm.

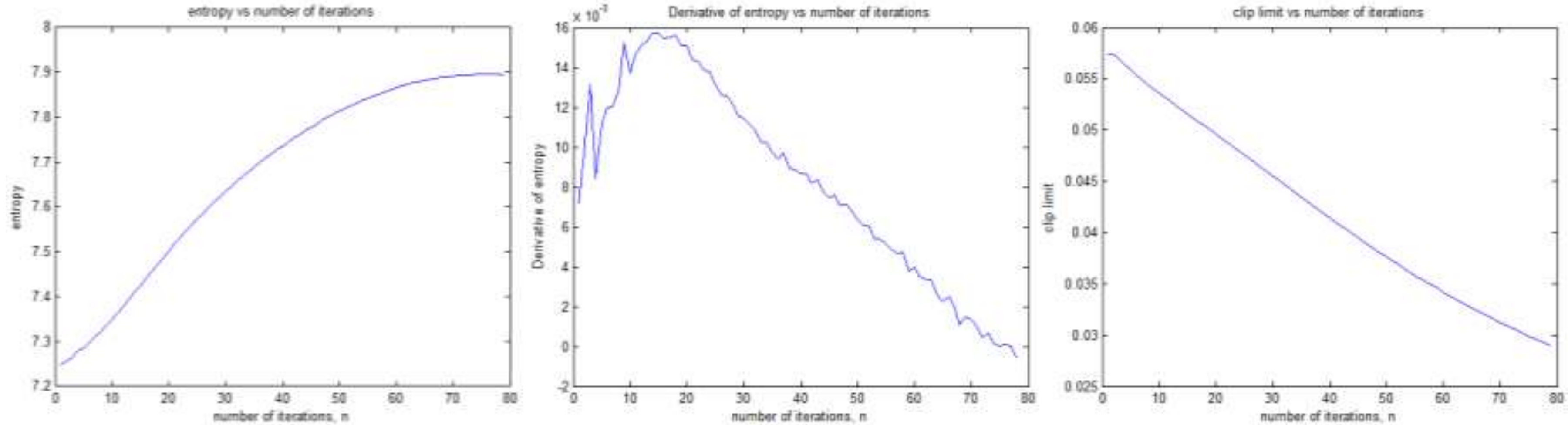
entropy vs number of iterations
entropy
number of iterations, n
Derivative of entropy vs number of iterations
Derivative of entropy
number of iterations, n
clip limit vs number of iterations
clip limit
number of iterations, n

(a)

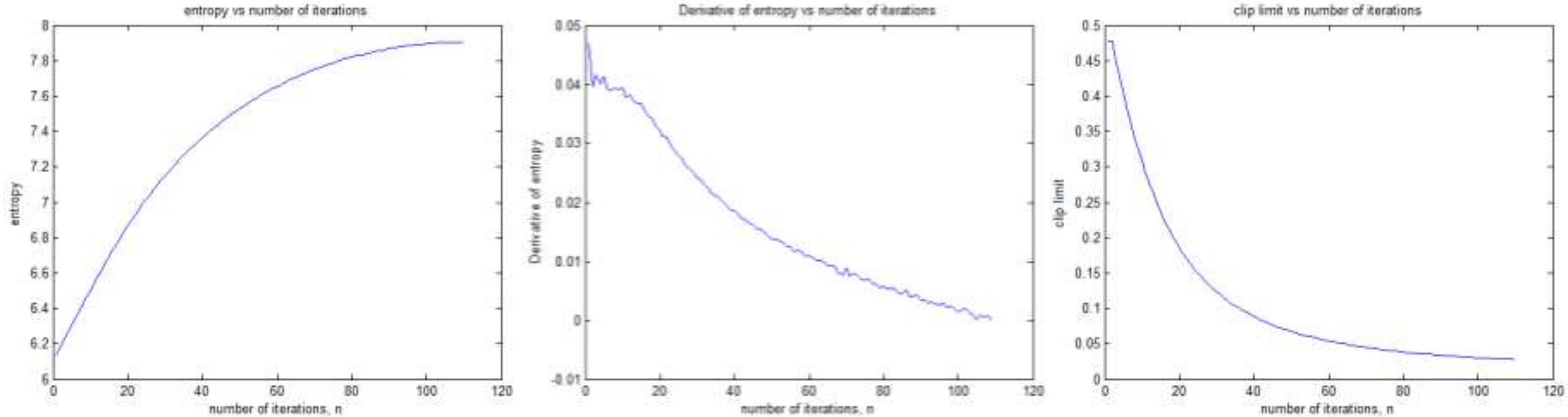
entropy vs number of iterations
entropy
number of iterations, n
Derivative of entropy vs number of iterations
Derivative of entropy
number of iterations, n
clip limit vs number of iterations
clip limit
number of iterations, n

(b)

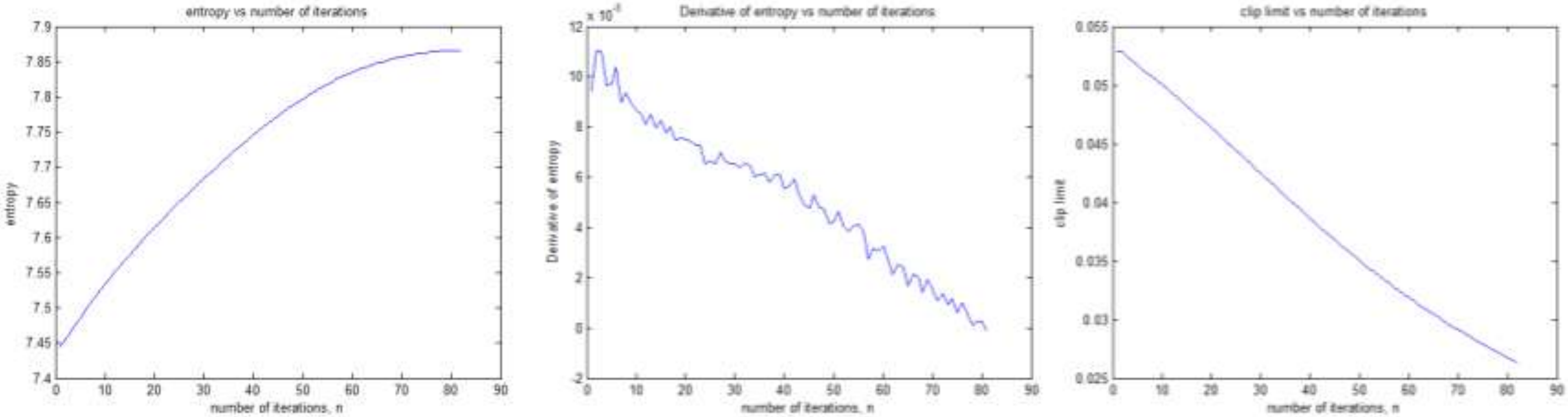
entropy vs number of iterations
entropy
number of iterations, n
Derivative of entropy vs number of iterations
Derivative of entropy
number of iterations, n
clip limit vs number of iterations
clip limit
number of iterations, n

(c)

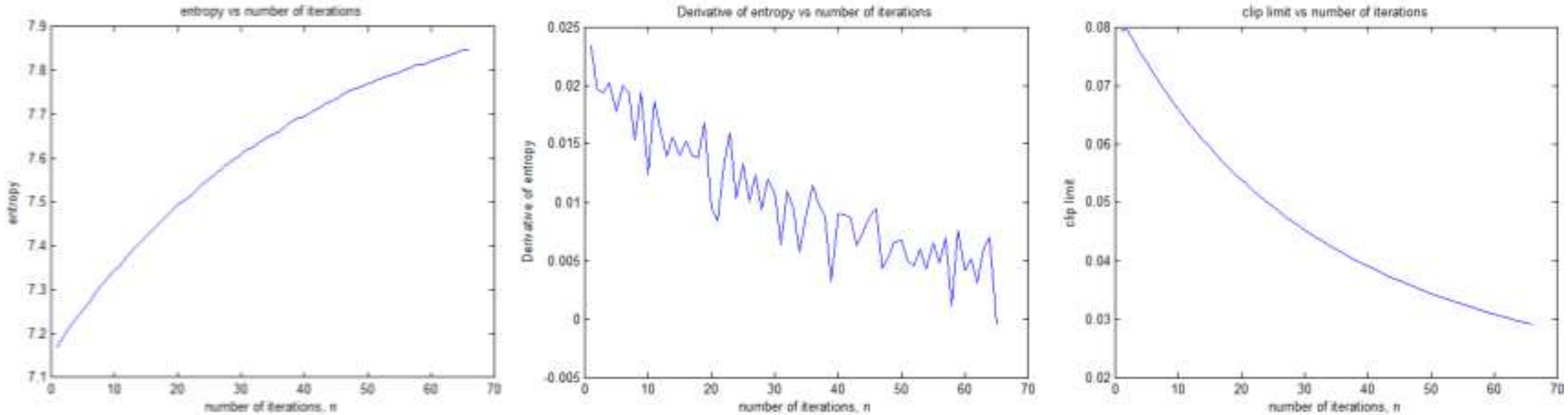
entropy vs number of iterations
entropy
number of iterations, n
Derivative of entropy vs number of iterations
Derivative of entropy
number of iterations, n
clip limit vs number of iterations
clip limit
number of iterations, n

(d)

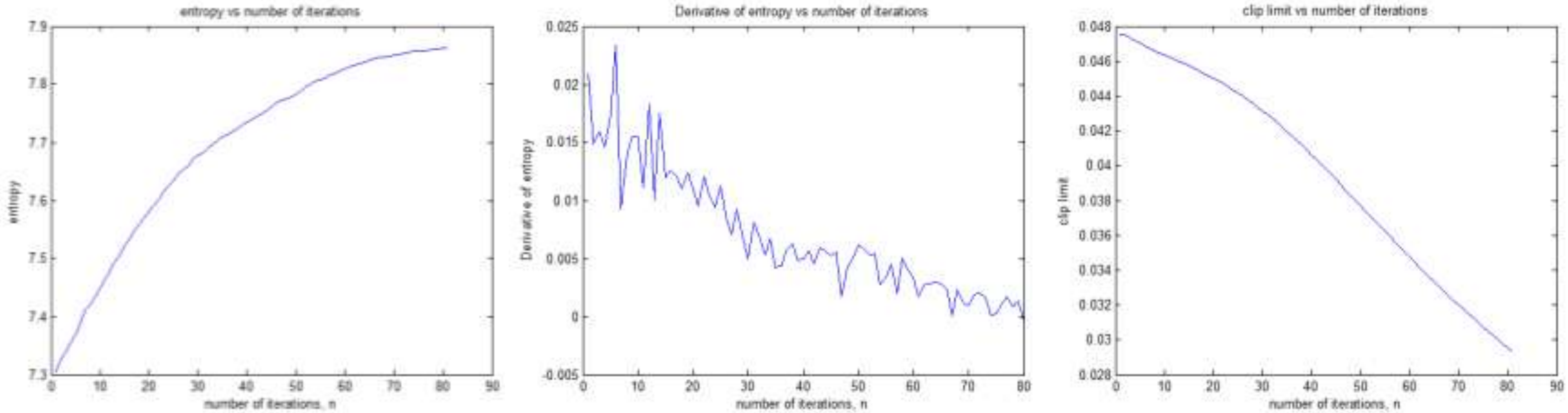
entropy vs number of iterations
entropy
number of iterations, n
Derivative of entropy vs number of iterations
Derivative of entropy
number of iterations, n
clip limit vs number of iterations
clip limit
number of iterations, n

(e)

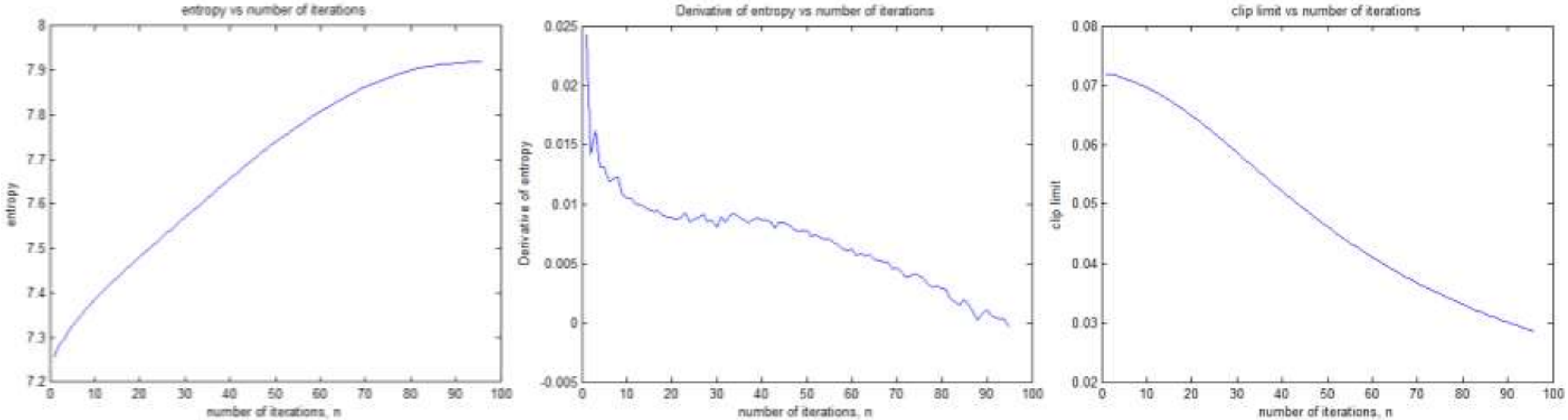
entropy vs number of iterations
entropy
number of iterations, n
Derivative of entropy vs number of iterations
Derivative of entropy
number of iterations, n
clip limit vs number of iterations
clip limit
number of iterations, n

(f)

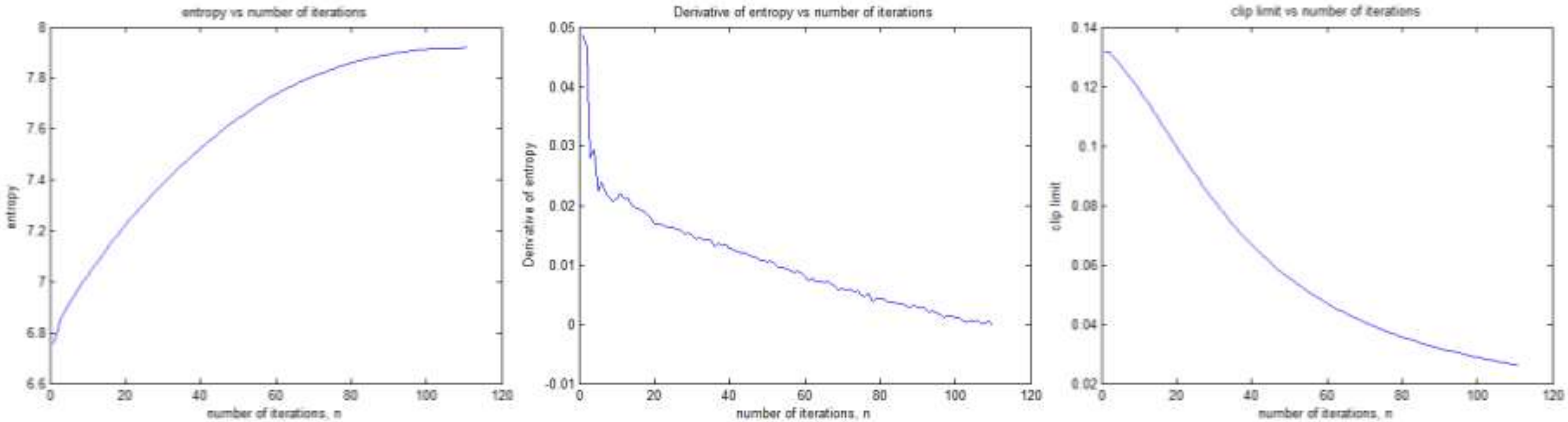
entropy vs number of iterations
entropy
number of iterations, n
Derivative of entropy vs number of iterations
Derivative of entropy
number of iterations, n
clip limit vs number of iterations
clip limit
number of iterations, n

(g)

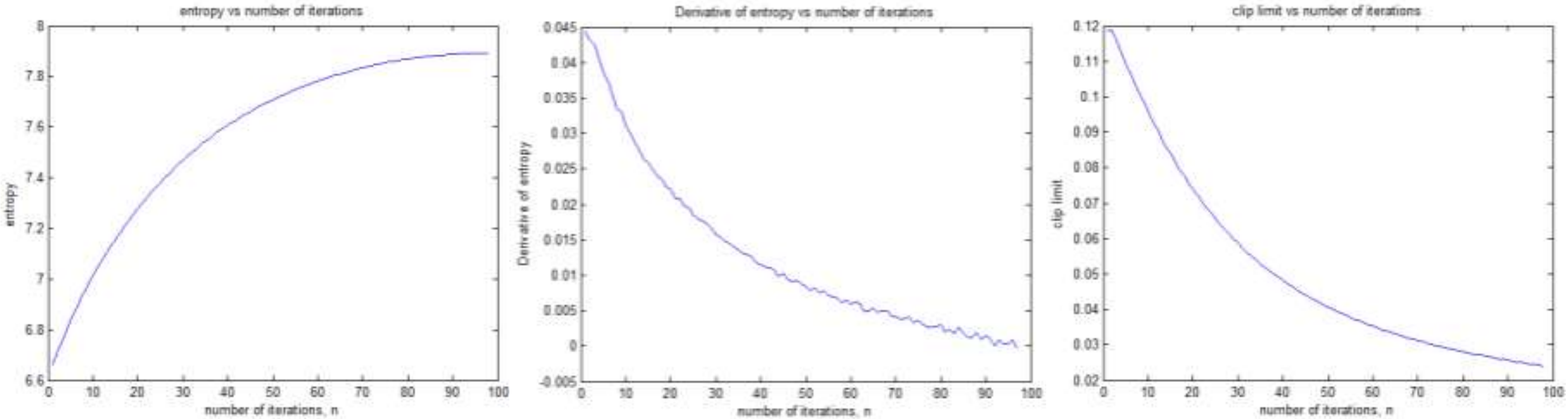
entropy vs number of iterations
entropy
number of iterations, n
Derivative of entropy vs number of iterations
Derivative of entropy
number of iterations, n
clip limit vs number of iterations
clip limit
number of iterations, n

(h)

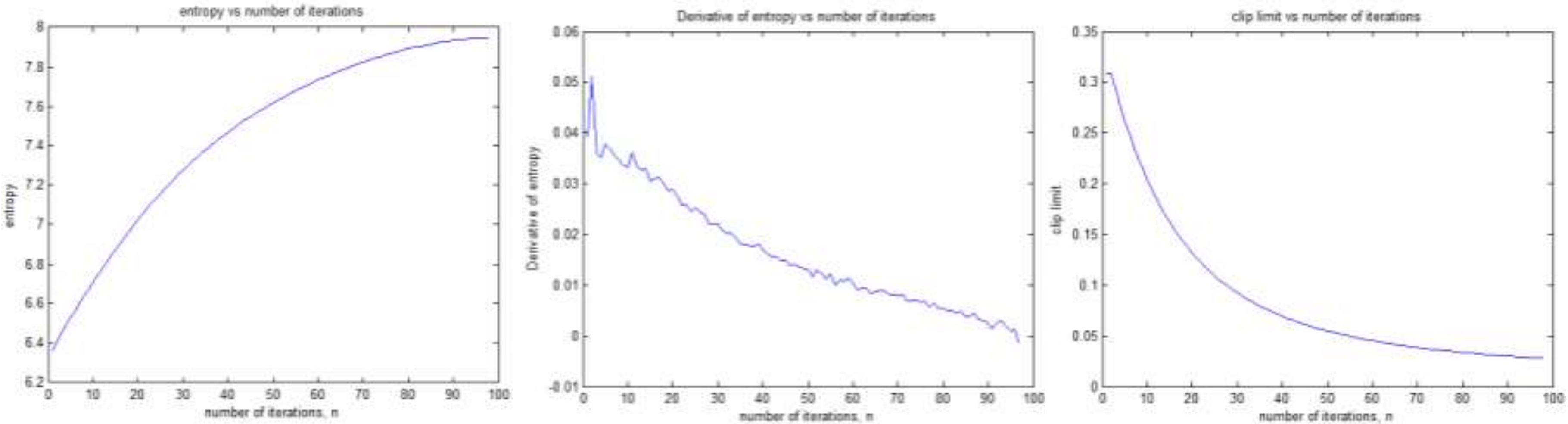
entropy vs number of iterations
entropy
number of iterations, n
Derivative of entropy vs number of iterations
Derivative of entropy
number of iterations, n
clip limit vs number of iterations
clip limit
number of iterations, n

(i)

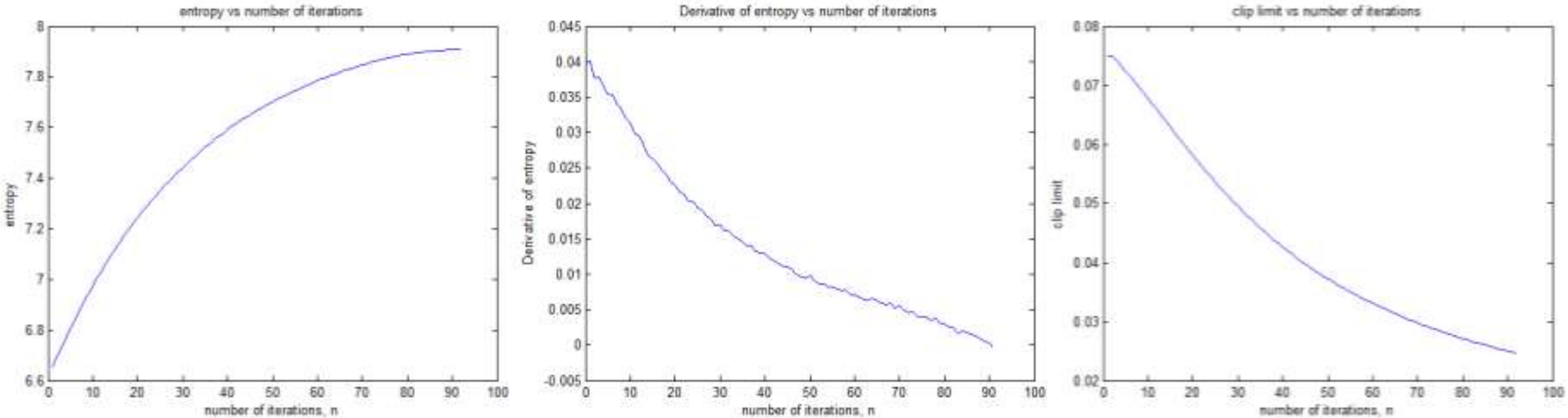
entropy vs number of iterations
entropy
number of iterations, n
Derivative of entropy vs number of iterations
Derivative of entropy
number of iterations, n
clip limit vs number of iterations
clip limit
number of iterations, n

(j)

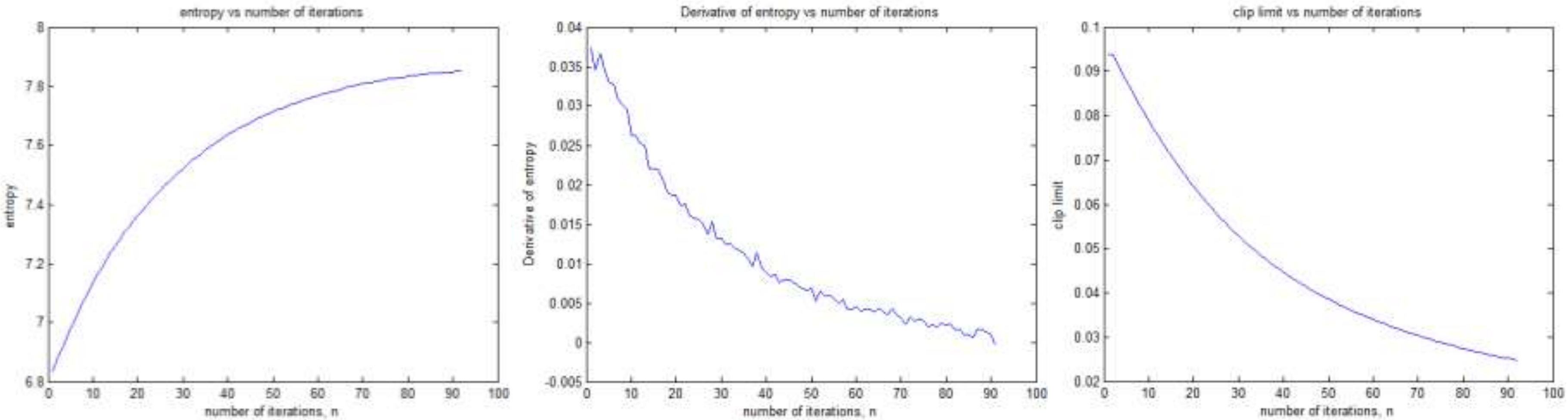
entropy vs number of iterations
entropy
number of iterations, n
Derivative of entropy vs number of iterations
Derivative of entropy
number of iterations, n
clip limit vs number of iterations
clip limit
number of iterations, n

(k)

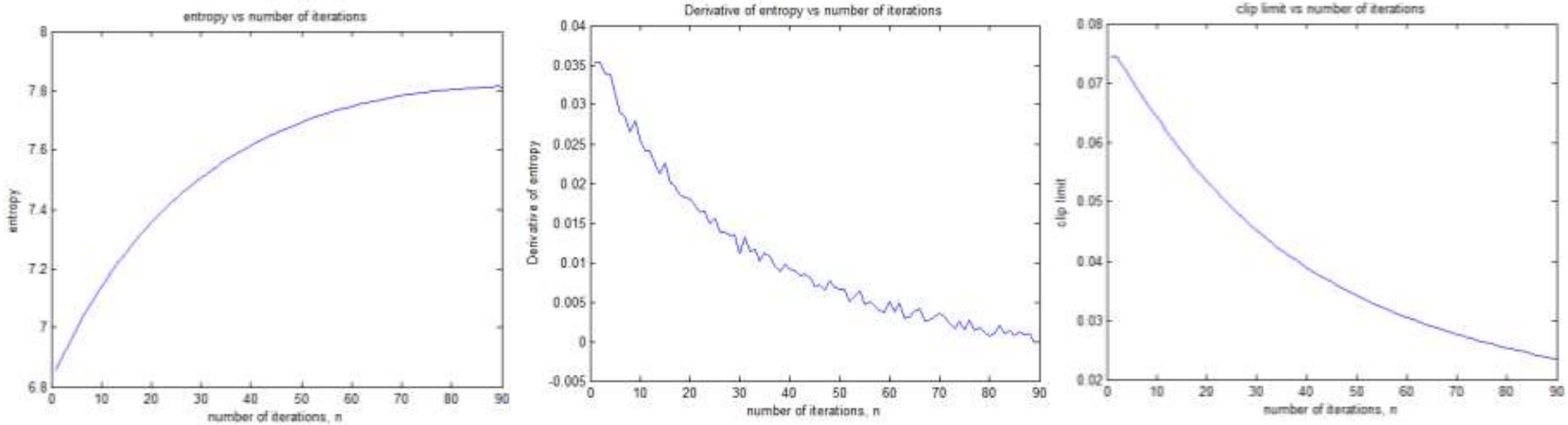
entropy vs number of iterations
entropy
number of iterations, n
Derivative of entropy vs number of iterations
Derivative of entropy
number of iterations, n
clip limit vs number of iterations
clip limit
number of iterations, n

(l)

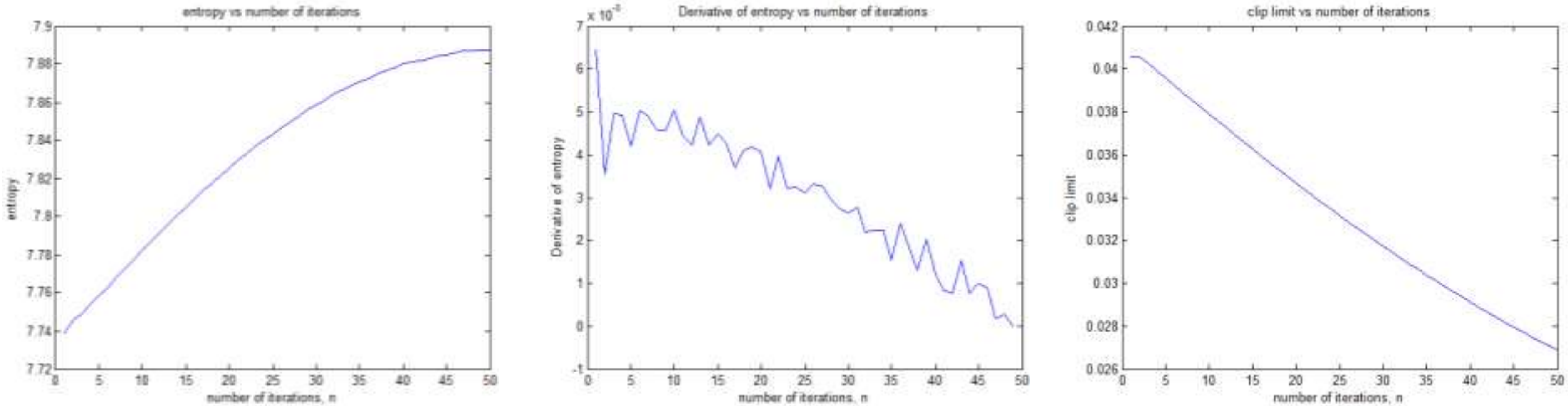
entropy vs number of iterations
entropy
number of iterations, n
Derivative of entropy vs number of iterations
x 10^-3
Derivative of entropy
number of iterations, n
clip limit vs number of iterations
clip limit
number of iterations, n

(m)

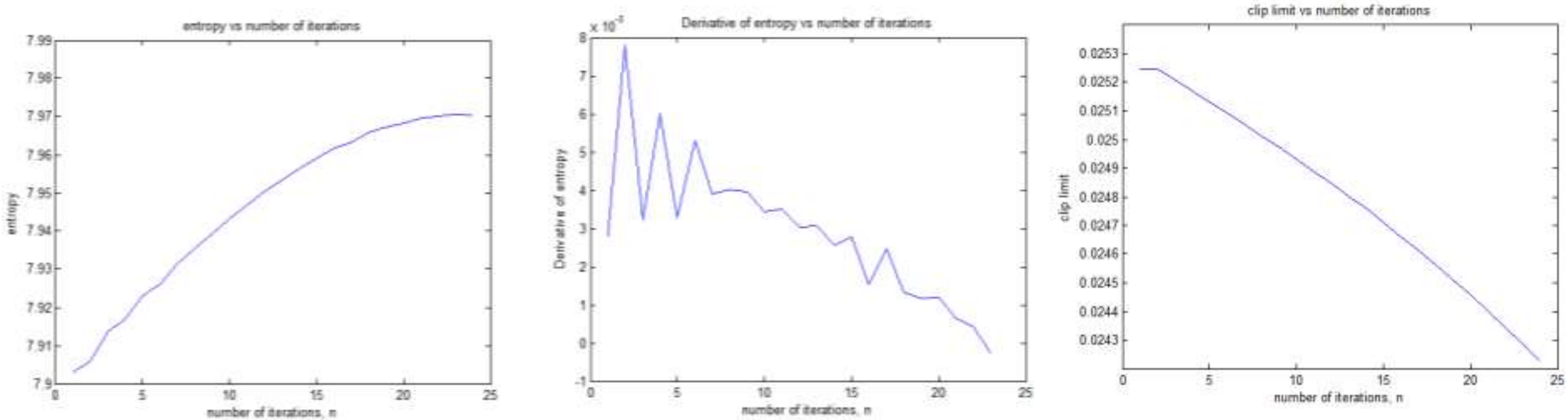
entropy vs number of iterations
entropy
number of iterations, n
Derivative of entropy vs number of iterations
x 10^-3
Derivative of entropy
number of iterations, n
clip limit vs number of iterations
clip limit
number of iterations, n

(n)

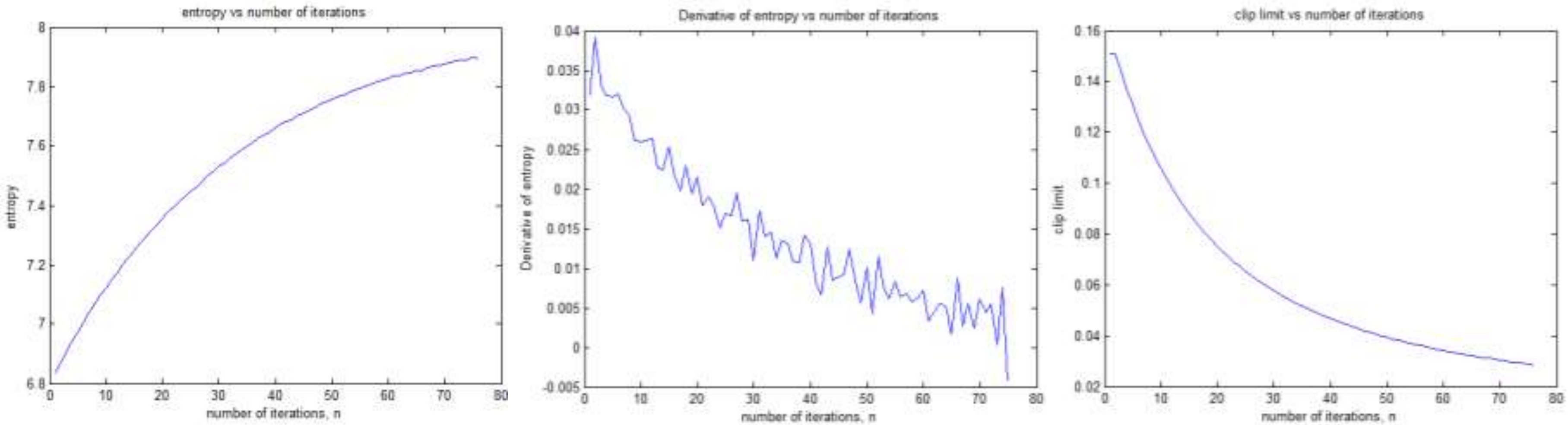
entropy vs number of iterations
entropy
number of iterations, n
Derivative of entropy vs number of iterations
Derivative of entropy
number of iterations, n
clip limit vs number of iterations
clip limit
number of iterations, n

(o)

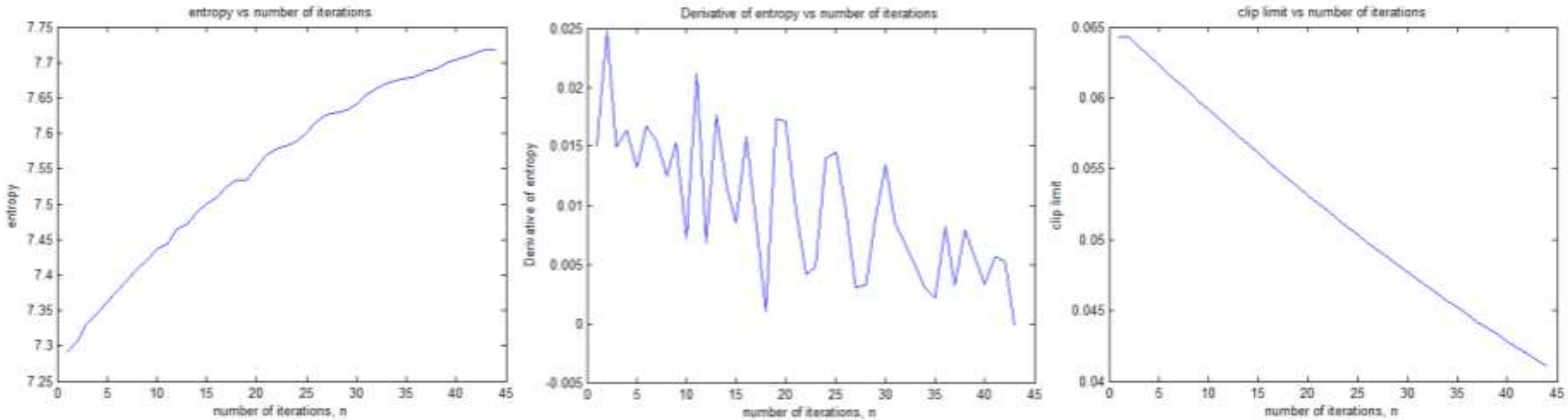
entropy vs number of iterations
entropy
number of iterations, n
Derivative of entropy vs number of iterations
Derivative of entropy
number of iterations, n
clip limit vs number of iterations
clip limit
number of iterations, n

(p)

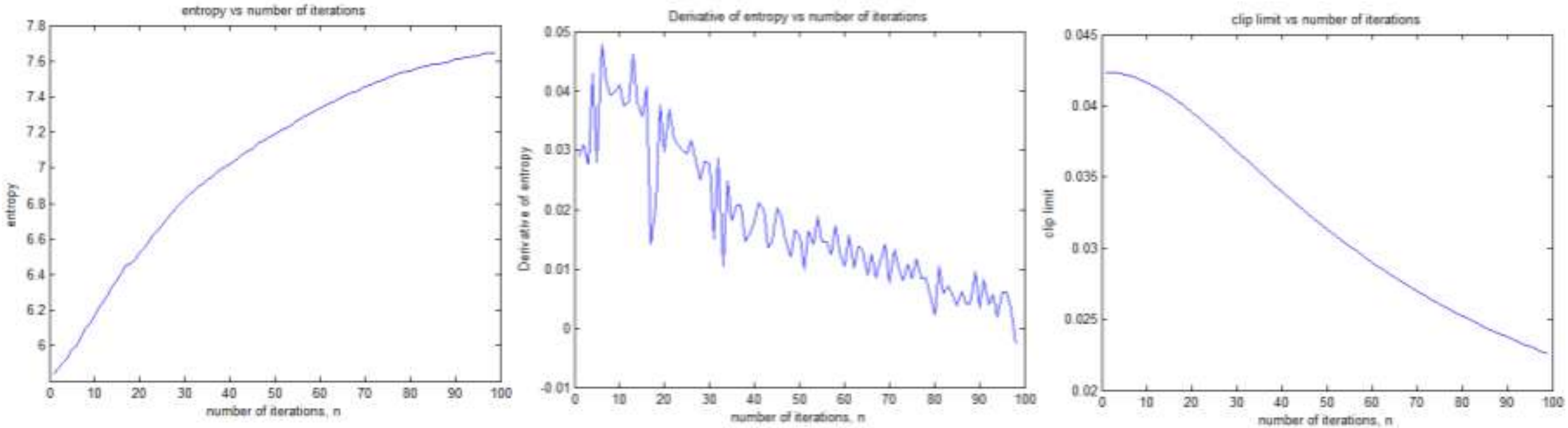
entropy vs number of iterations
entropy
number of iterations, n
Derivative of entropy vs number of iterations
Derivative of entropy
number of iterations, n
clip limit vs number of iterations
clip limit
number of iterations, n

(q)

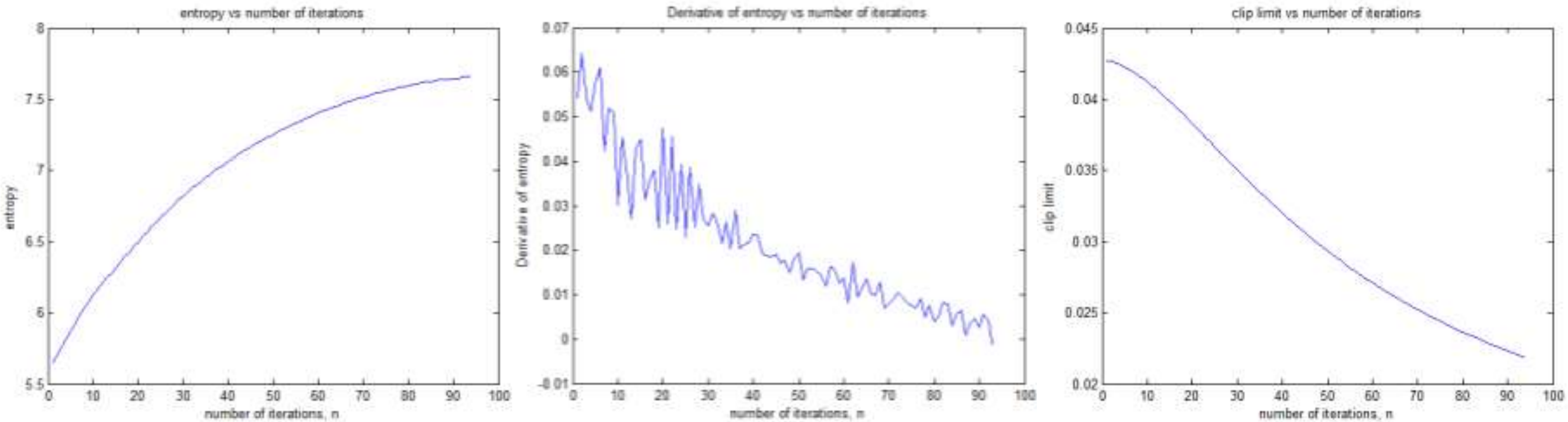
entropy vs number of iterations
entropy
number of iterations, n
Derivative of entropy vs number of iterations
Derivative of entropy
number of iterations, n
clip limit vs number of iterations
clip limit
number of iterations, n

(r)

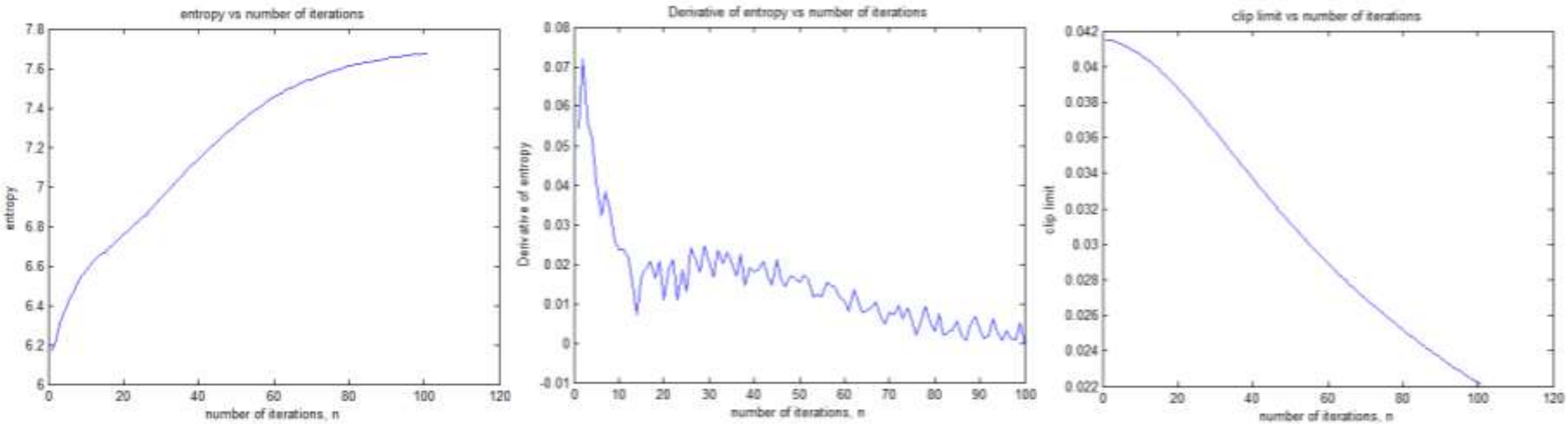
entropy vs number of iterations
entropy
number of iterations, n
Derivative of entropy vs number of iterations
Derivative of entropy
number of iterations, n
clip limit vs number of iterations
clip limit
number of iterations, n

(s)

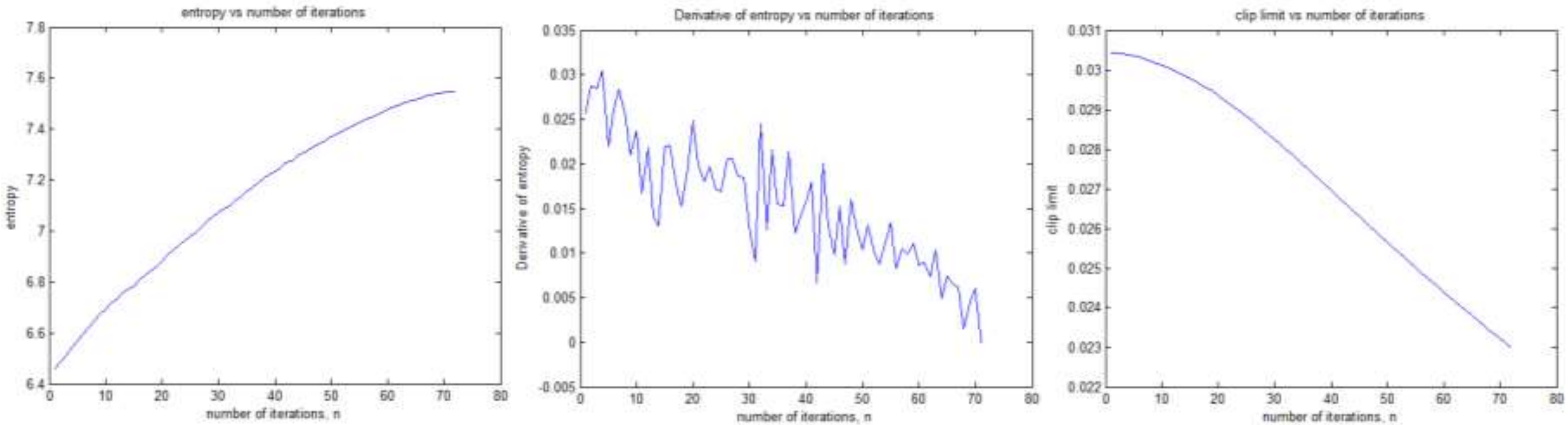
entropy vs number of iterations
entropy
number of iterations, n
Derivative of entropy vs number of iterations
Derivative of entropy
number of iterations, n
clip limit vs number of iterations
clip limit
number of iterations, n

(t)

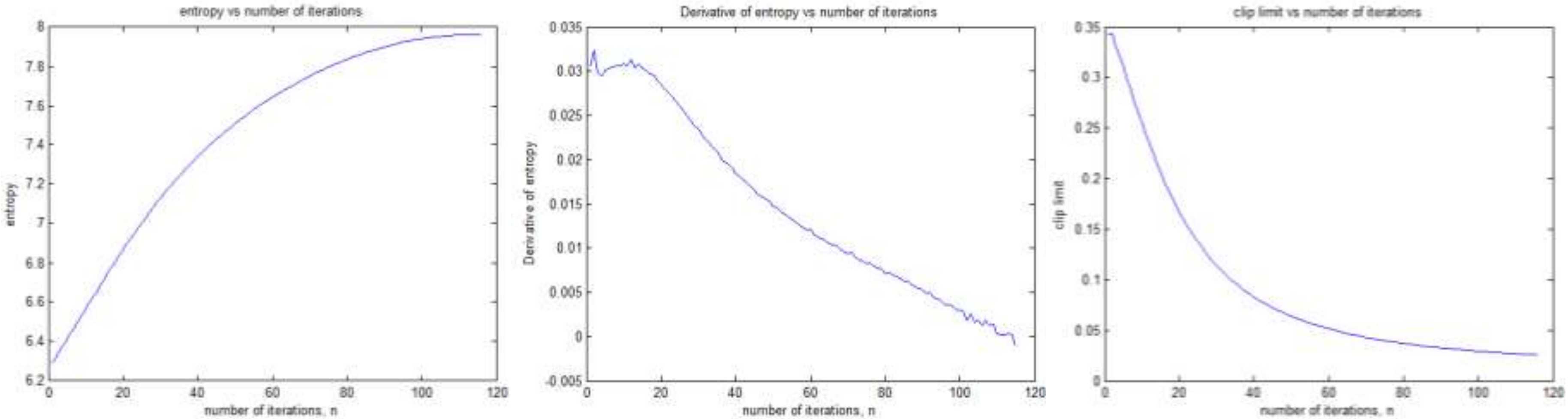
entropy vs number of iterations
entropy
number of iterations, n
Derivative of entropy vs number of iterations
Derivative of entropy
number of iterations, n
clip limit vs number of iterations
clip limit
number of iterations, n

(u)

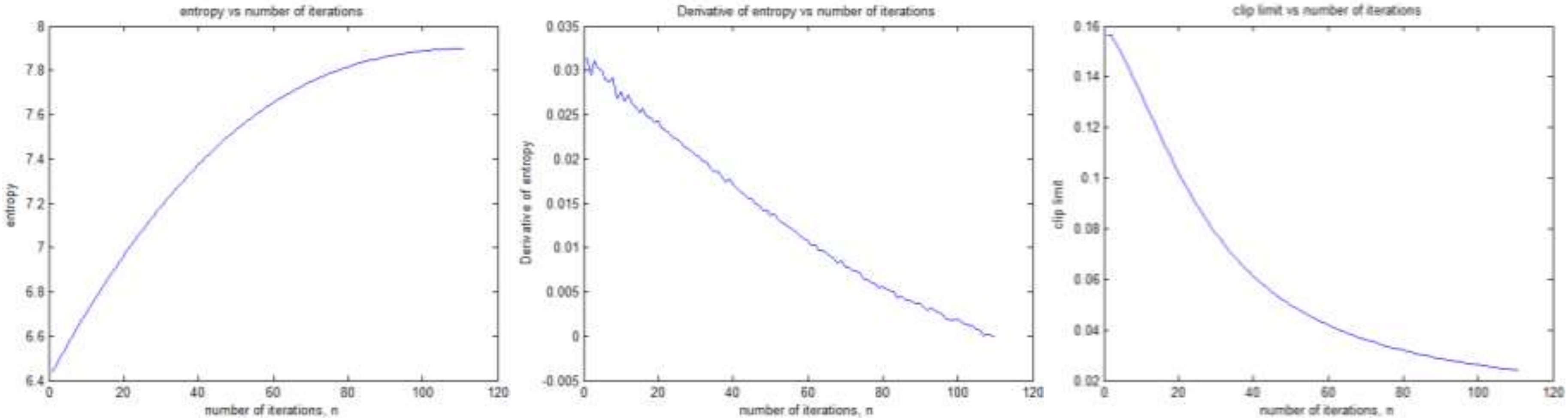
entropy vs number of iterations
entropy
number of iterations, n
Derivative of entropy vs number of iterations
Derivative of entropy
number of iterations, n
clip limit vs number of iterations
clip limit
number of iterations, n

(v)

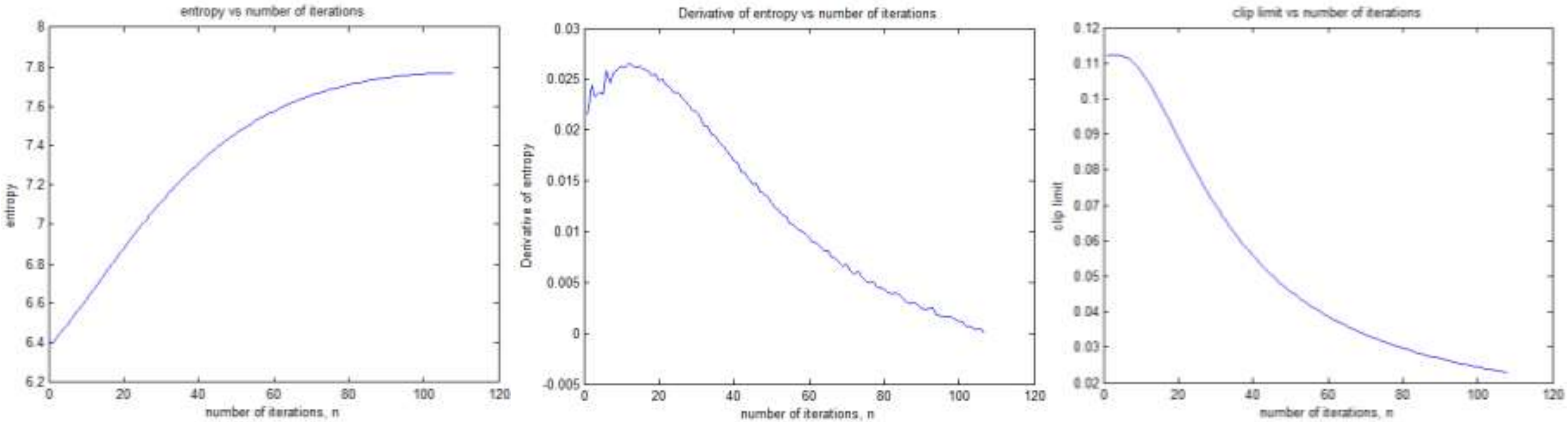

(w)

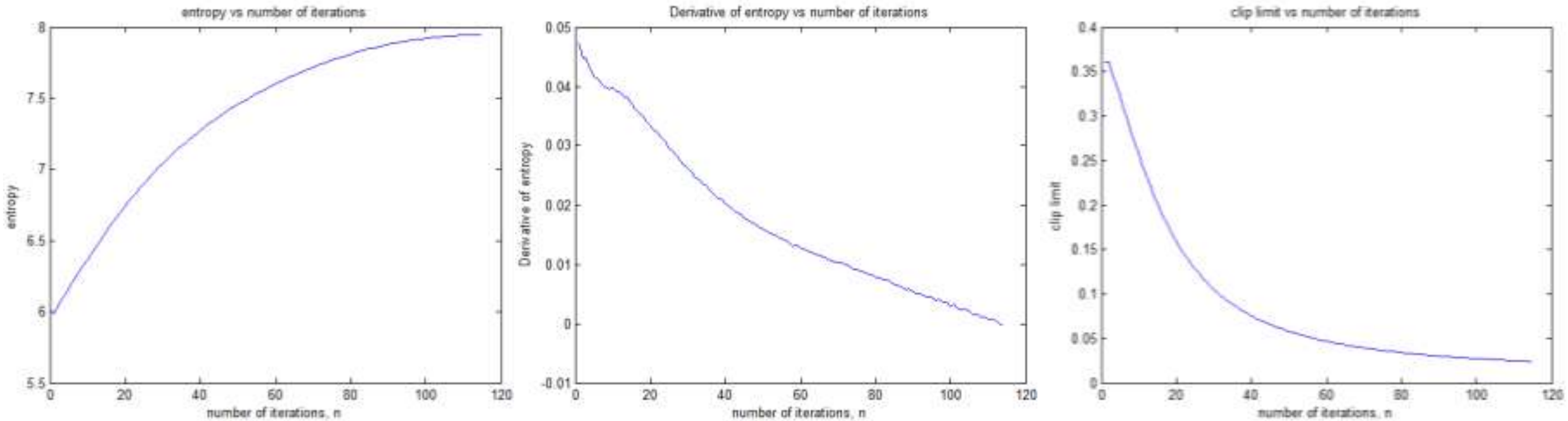

(x)

**Fig. 6B** Plots of entropy, derivative of entropy and clip limit versus the number of iterations for underwater images in 6A

### 3.4 Entropy optimized GOC2-CLAHE PA-2 (PA-2B)

Due to the need for automation and reduction of computation time in addition to trends observed in experiments using the dynamic clip computation, an entropy optimized and GOC2 regularized version of PA-2 was realized. The results are shown in Fig. 7 using different values for β and the results are strikingly different from those obtained with all other algorithms. For images of this nature, PA-2B yielded the best results with reduced computation compared with the adaptive clip limit version.

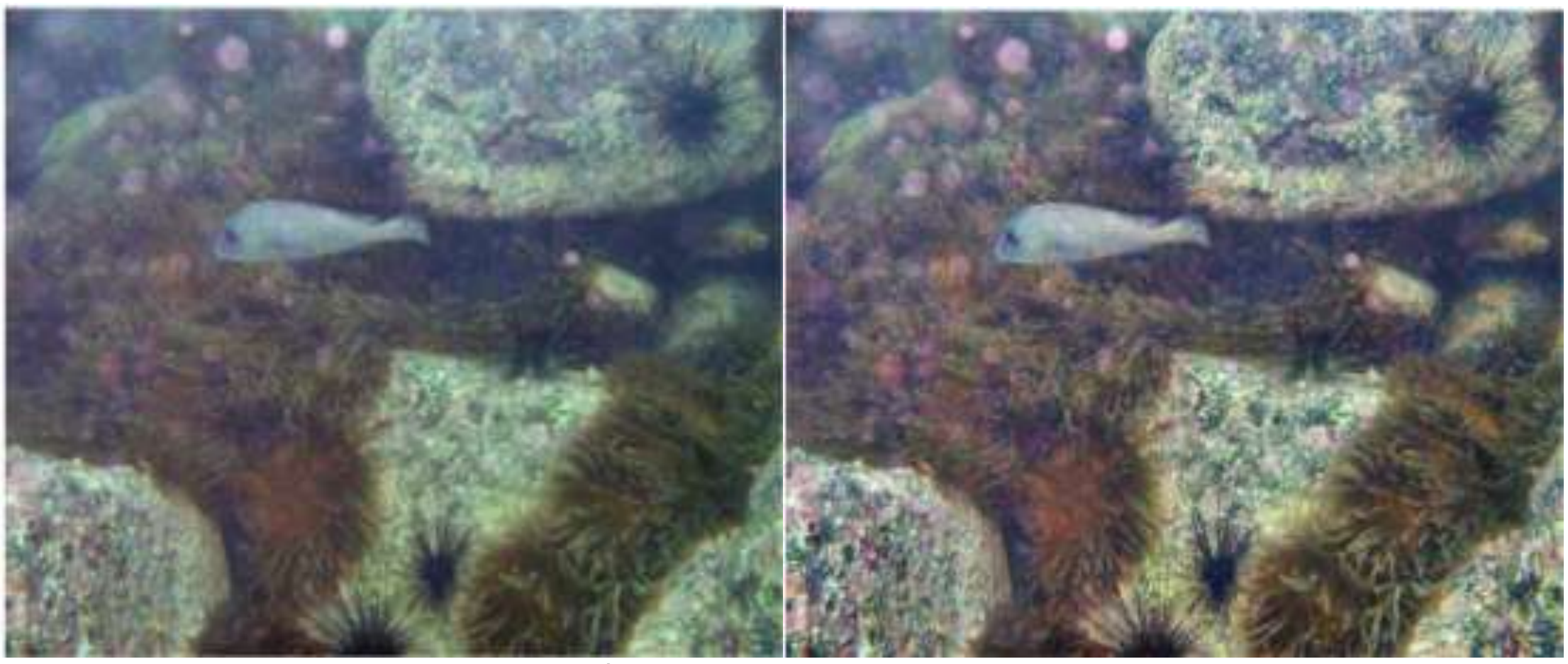

**Fig. 7** Results of using PA-2B with (a) $\beta = 0.5$ (b) $\beta = 1$

## 4 Experiments and results

This section presents results of the proposed algorithms compared with several algorithms from the literature and their PDE-based formulations. Though the results are varied, in general, PA-2B yields the most consistent and balanced results. The contrast stretching and histogram-based algorithms' results vary widely from quite good to very bad in terms of high contrast and good colour correction coupled with overexposure and colour distortion. Conversely, the local enhancement properties of the AHE and CLAHE show high local contrast and noise enhancement with reduced colour correction. The Retinex methods yield dark images or with grey hue while the GUM and HF approaches yield over-bright and faded colour images. PA-1 variants yield mild results for images due to its soft approach while PA-2 variants attempt both local and global contrast enhancement with colour correction while avoiding noise over enhancement.

Based on the study of image colour histograms shown in Fig. 8, PA-1, PA-2 and its variants work best when R, G and B histograms are closely overlapping whereas PA-2 and its variants perform both local contrast enhancement and colour correction while avoiding overexposure in bright regions. The conventional contrast stretching and histogram equalization approaches yield images with over-exposed bright regions even though there is excellent colour correction and global contrast enhancement.

Images with wide disparity between the R, G and B histograms are difficult to process in terms of proper colour correction. In other words, images with either R, G or B histograms completely out of alignment or distant from the other histograms and having a relatively flat shape make colour correction difficult using the proposed approaches. For example, images with thick green or blue haze means that the green or blue channel histogram is furthest from the red and blue or red and green channel histograms. Thus, the PA-1 and PA-2 fail for these images with very flat distributions while conventional contrast stretching algorithms yield varying results as noted earlier. Further study will involve processing in various colour spaces to ascertain the best choice and or combination of processes to mitigate these drawbacks.

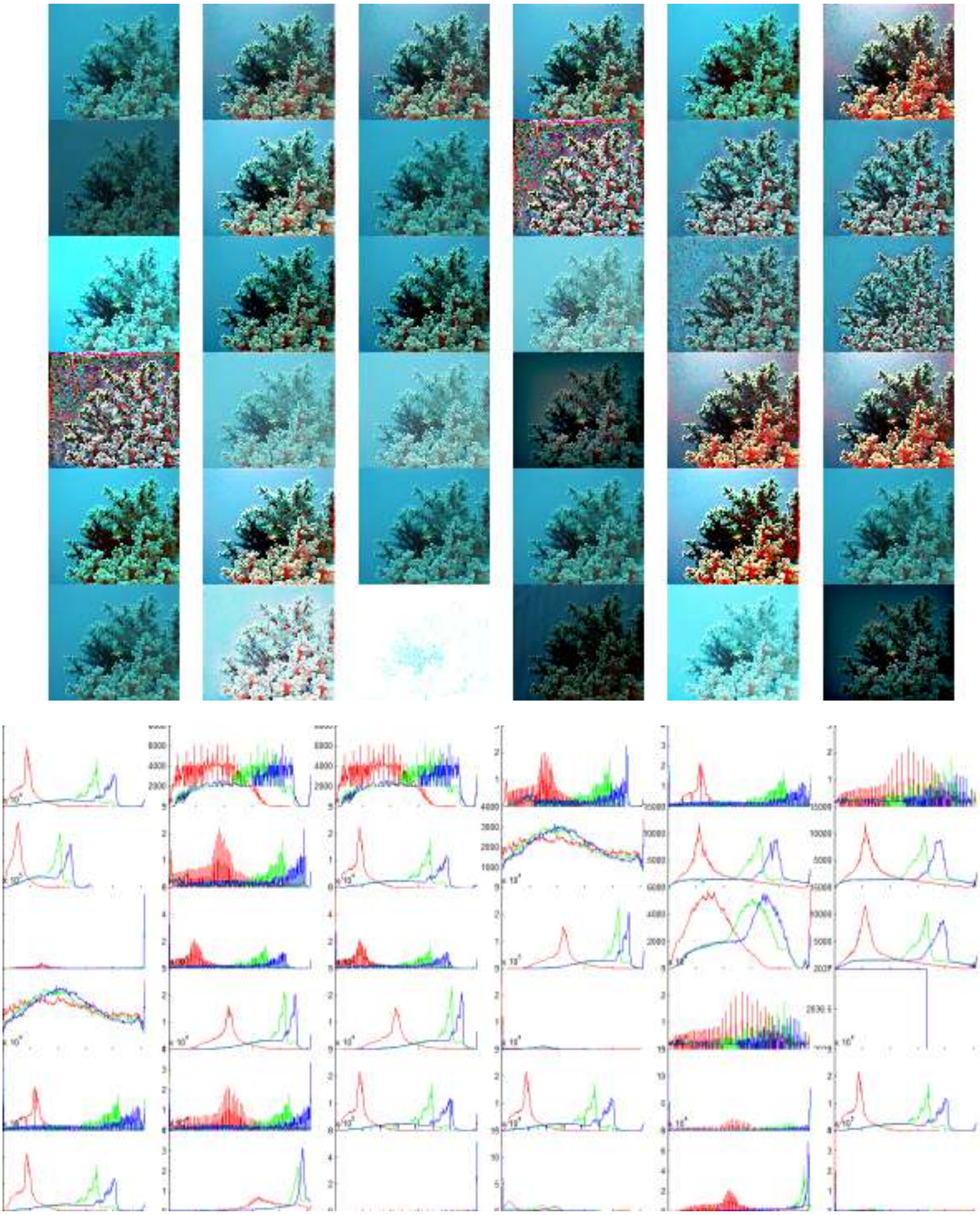

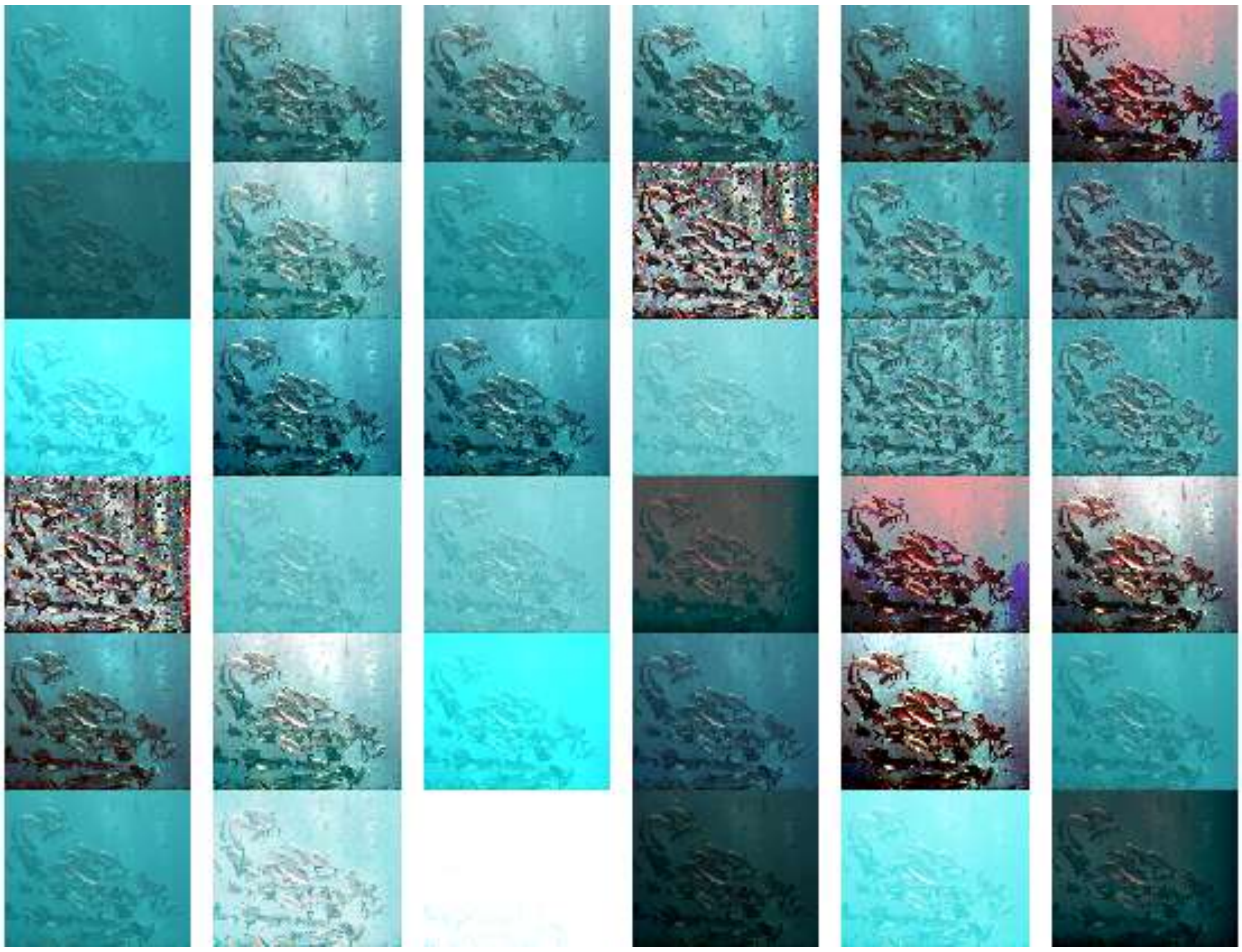

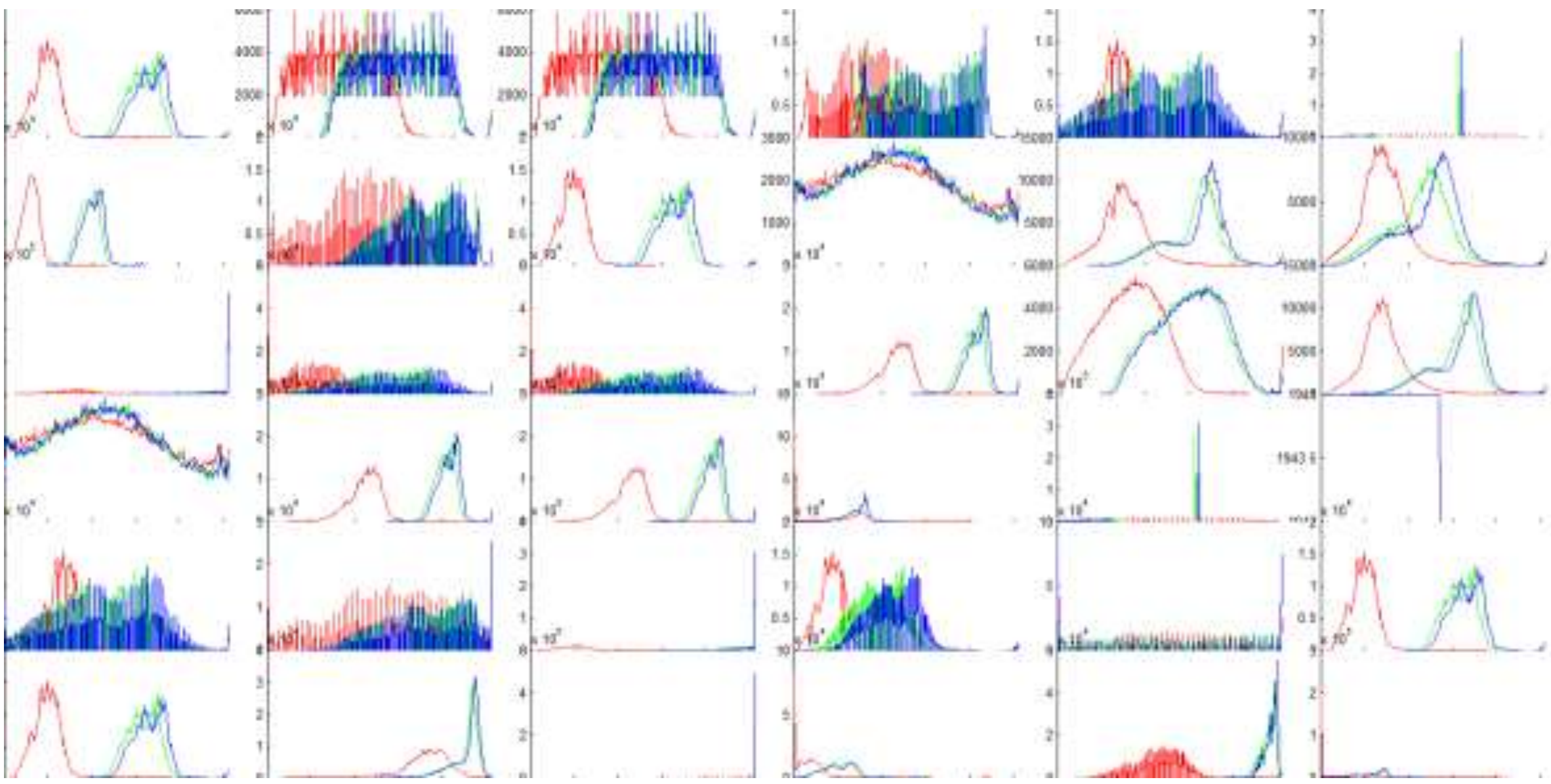

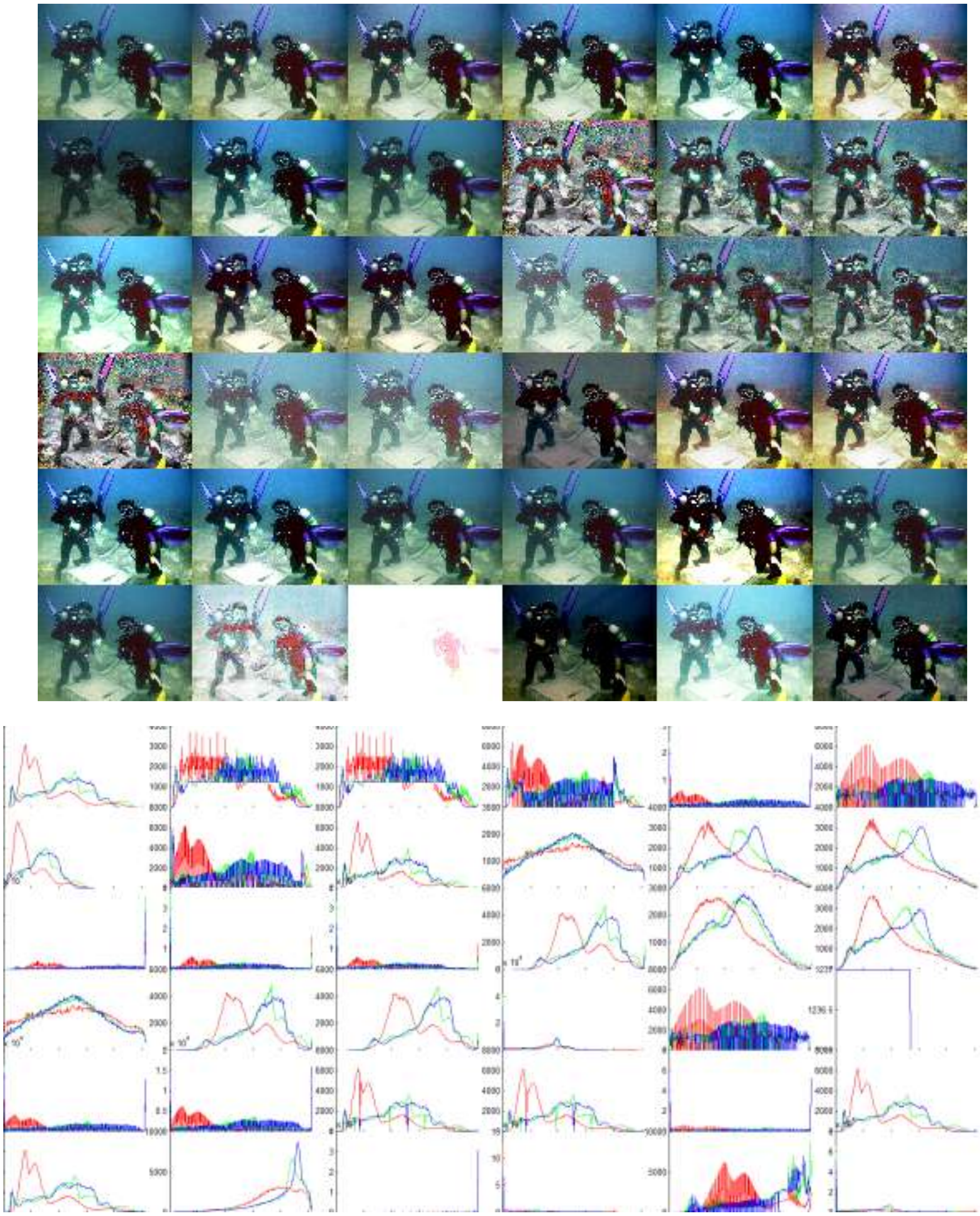

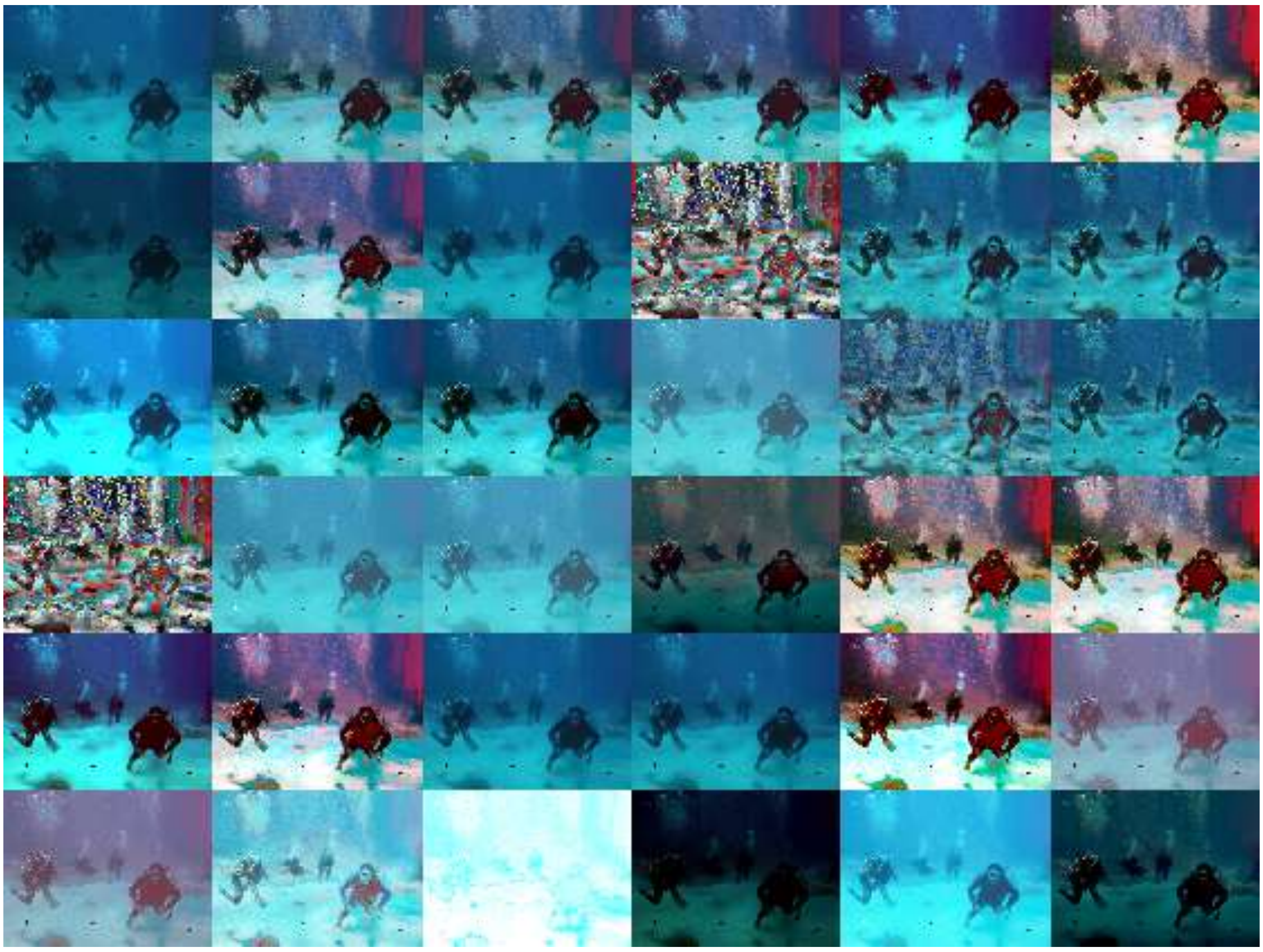

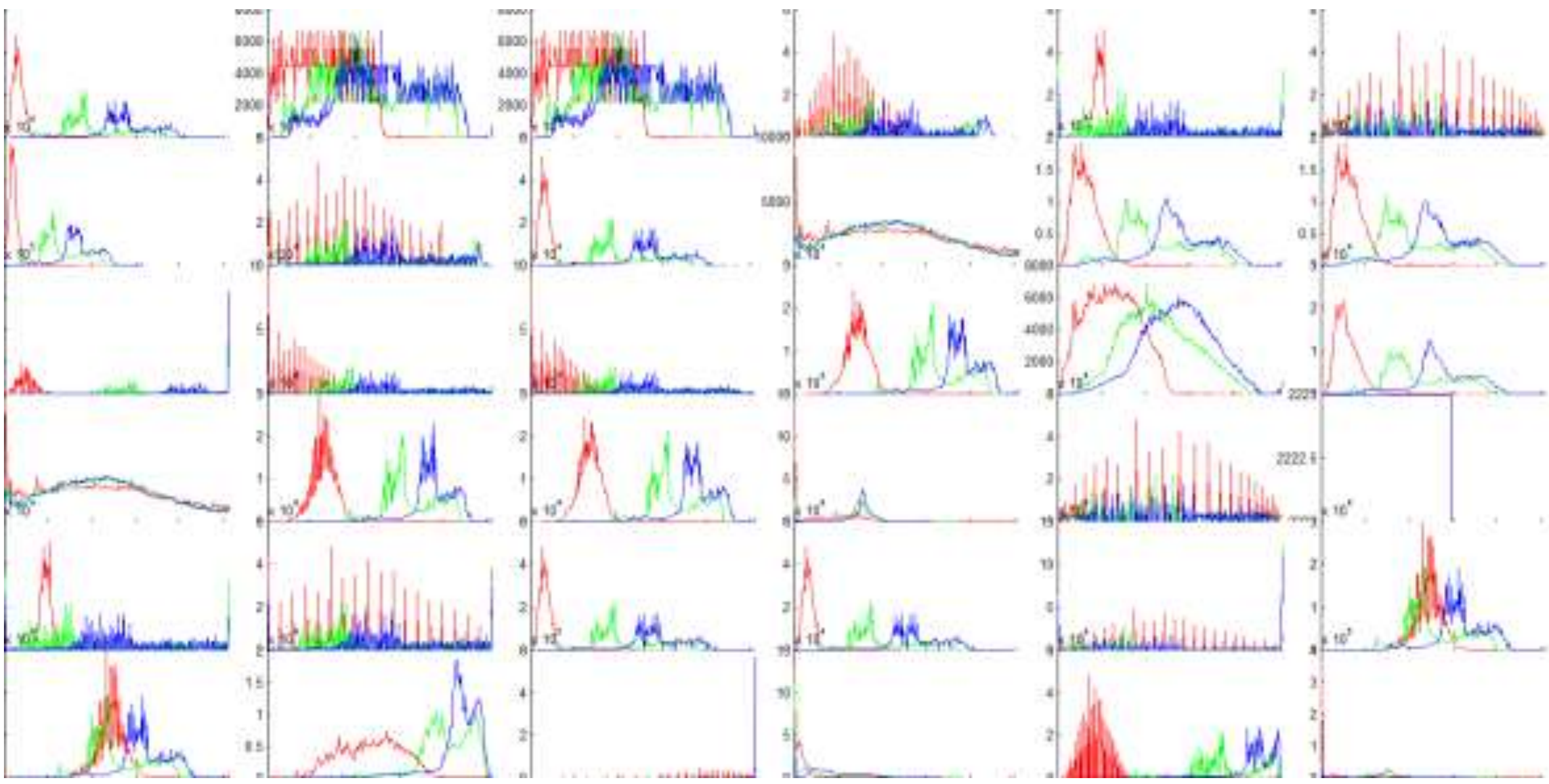

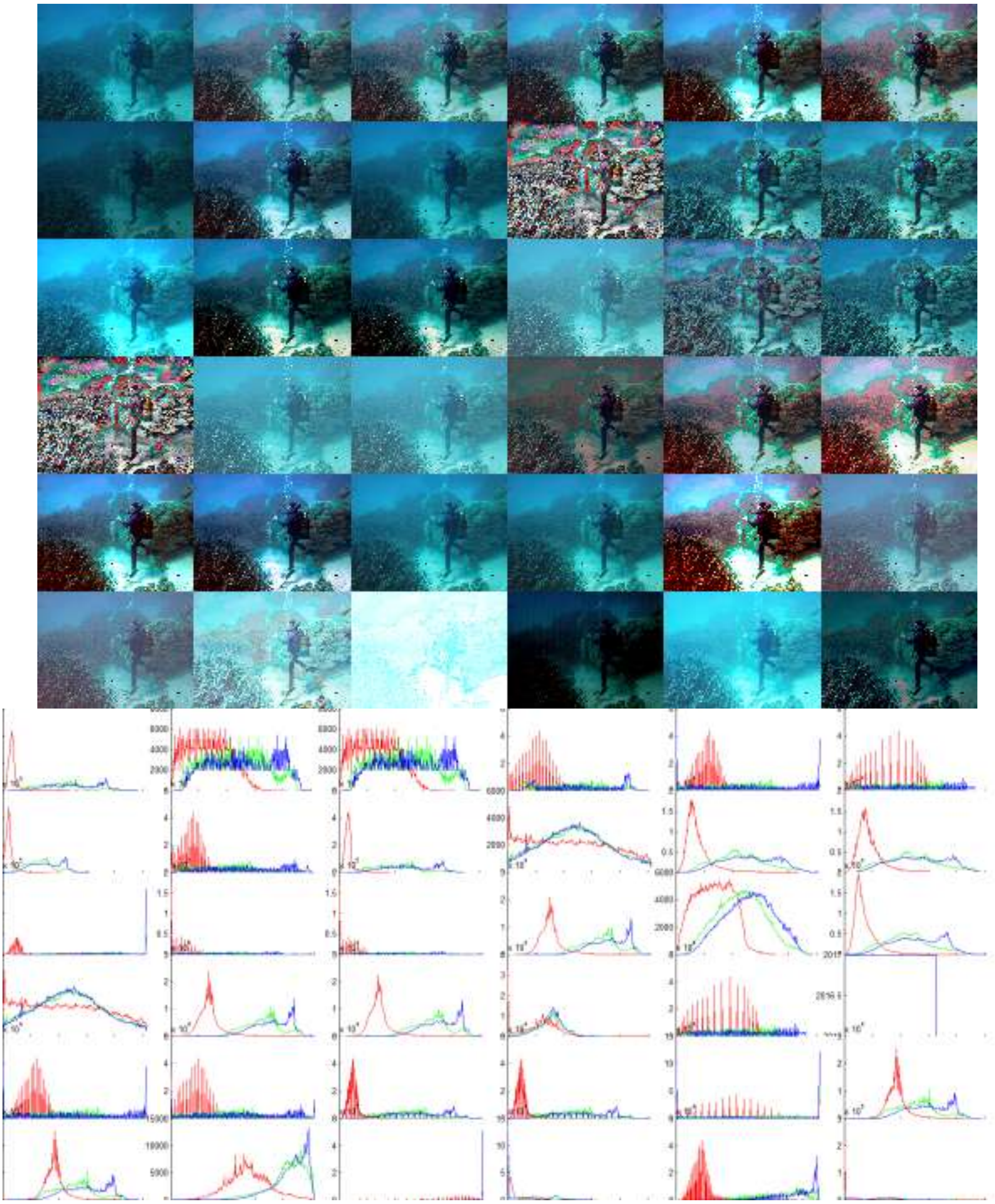

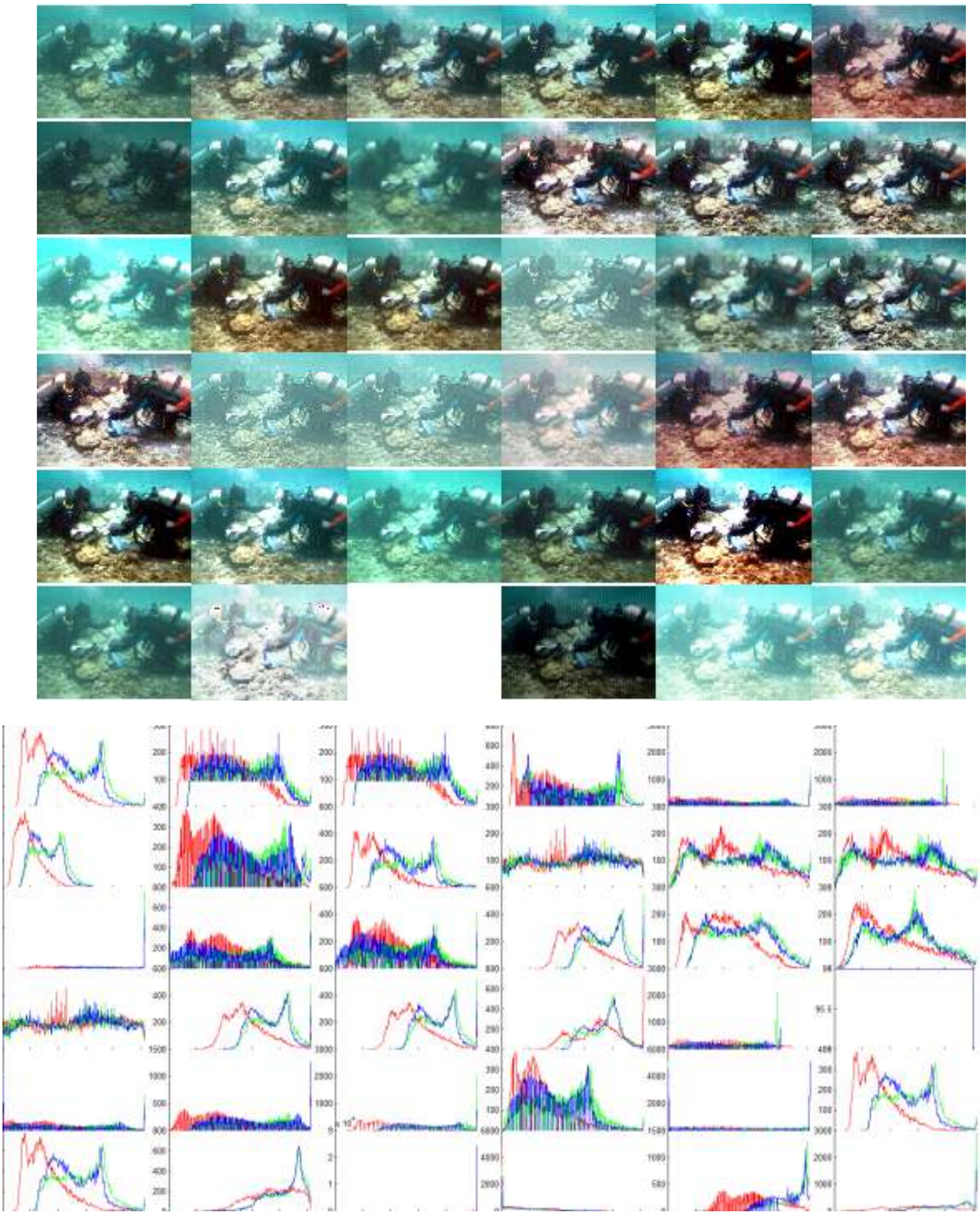

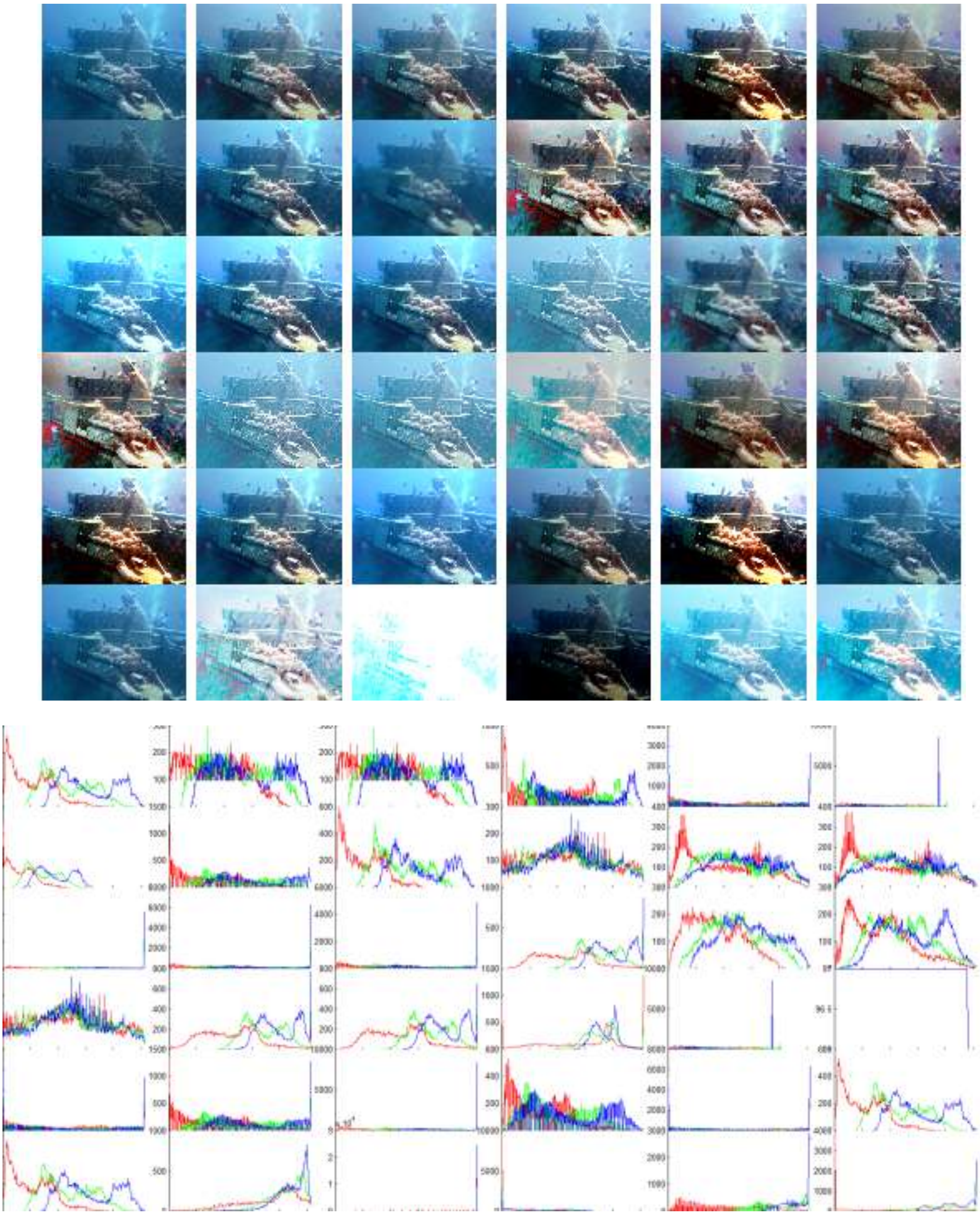

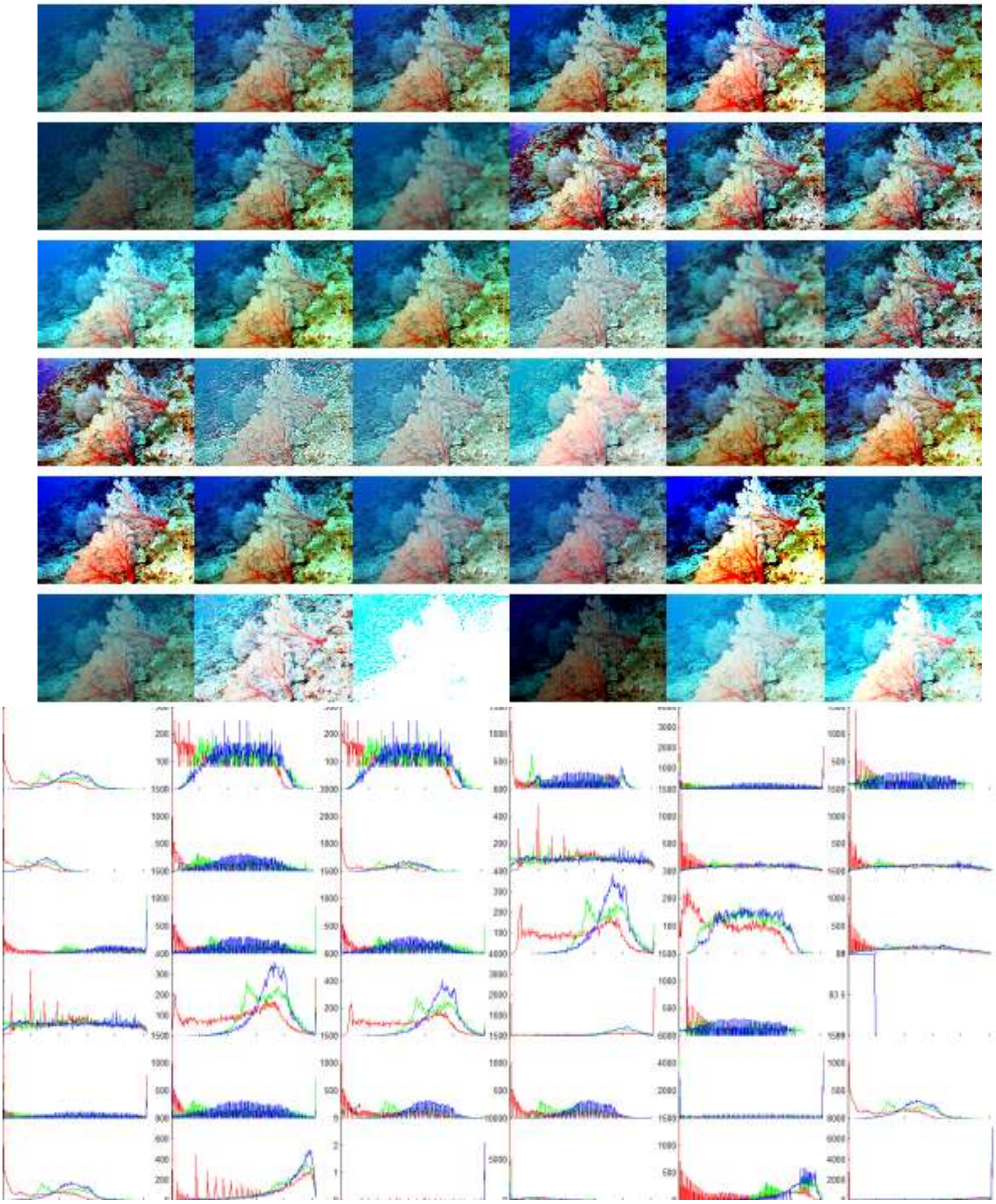

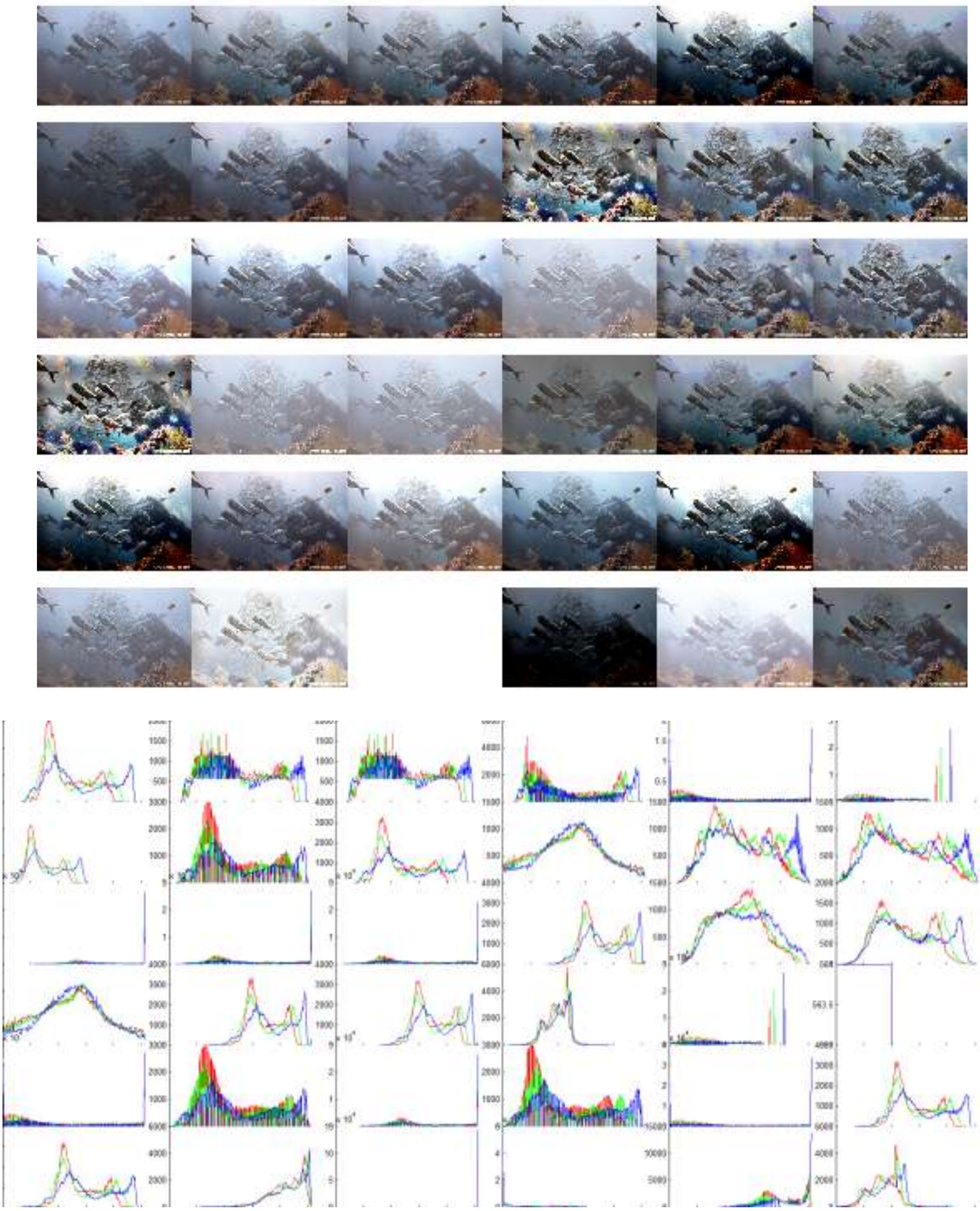

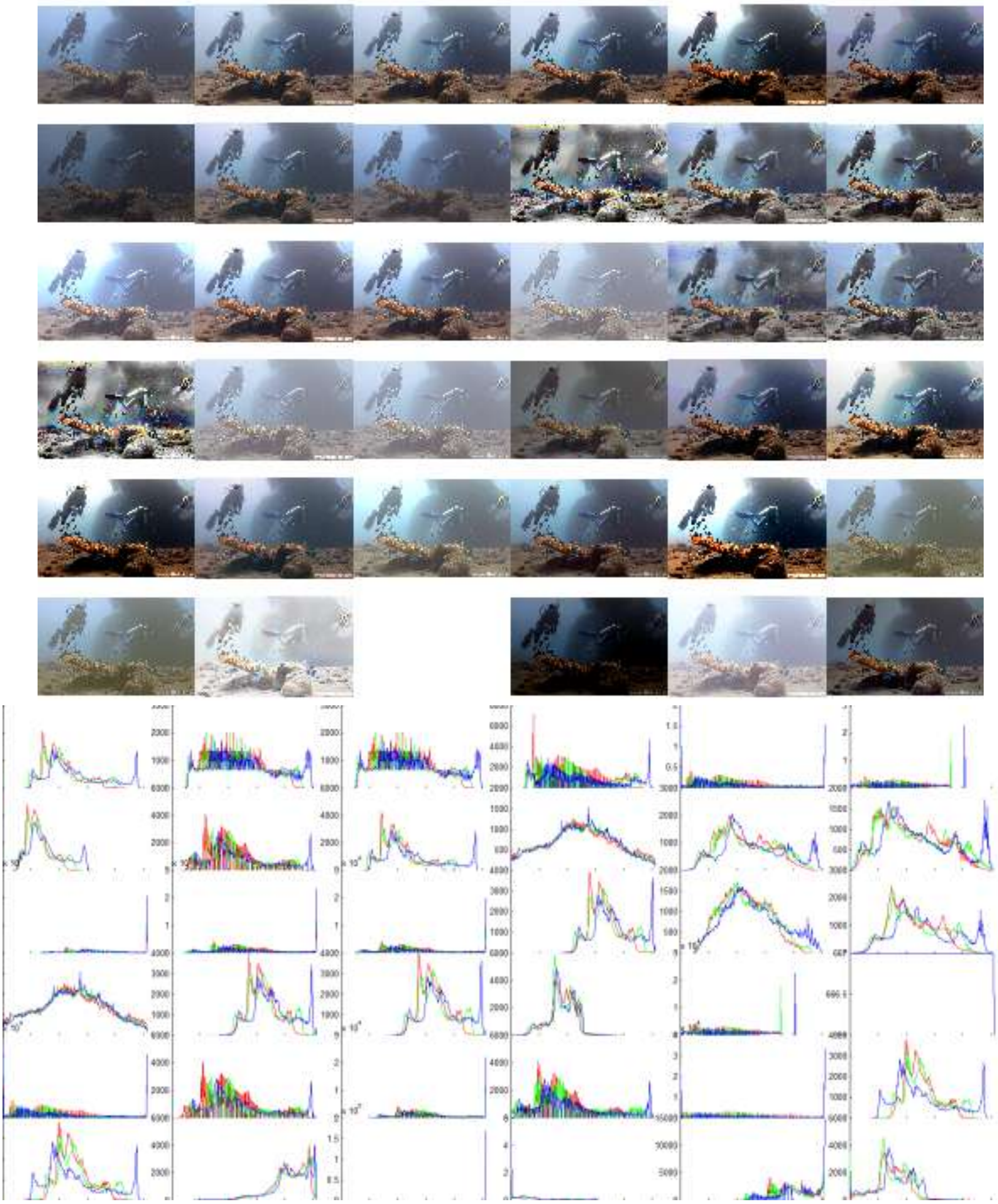

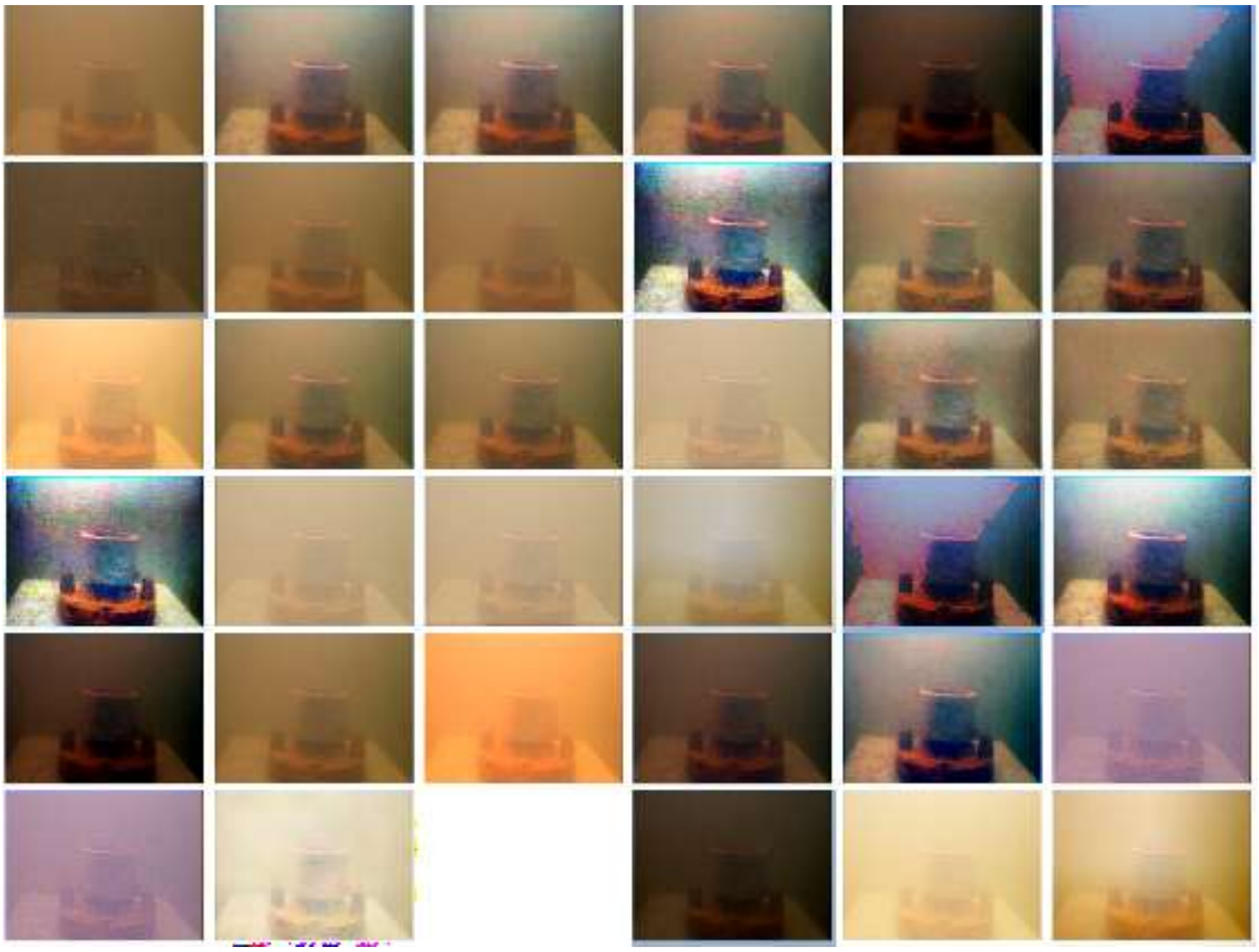

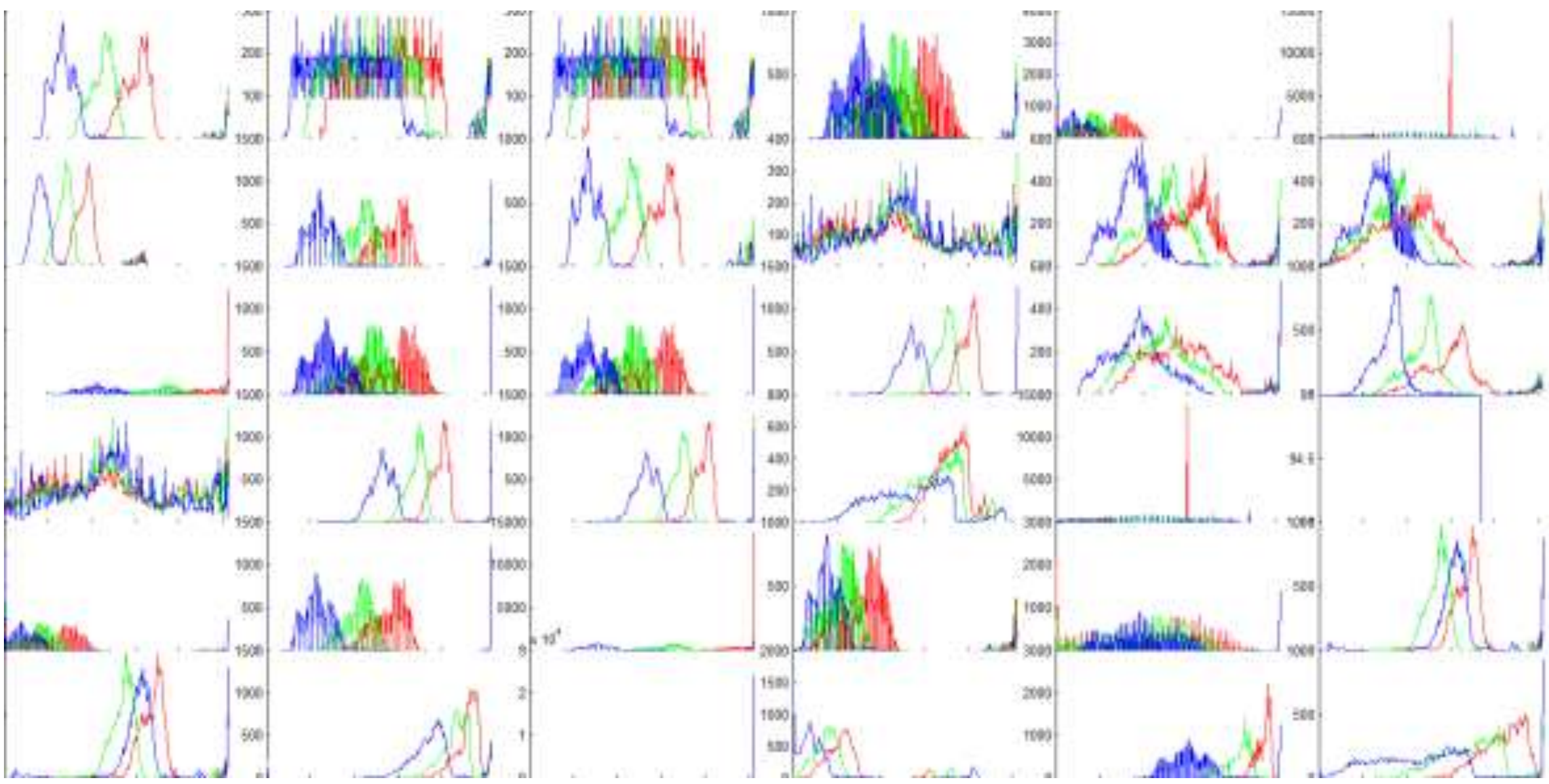

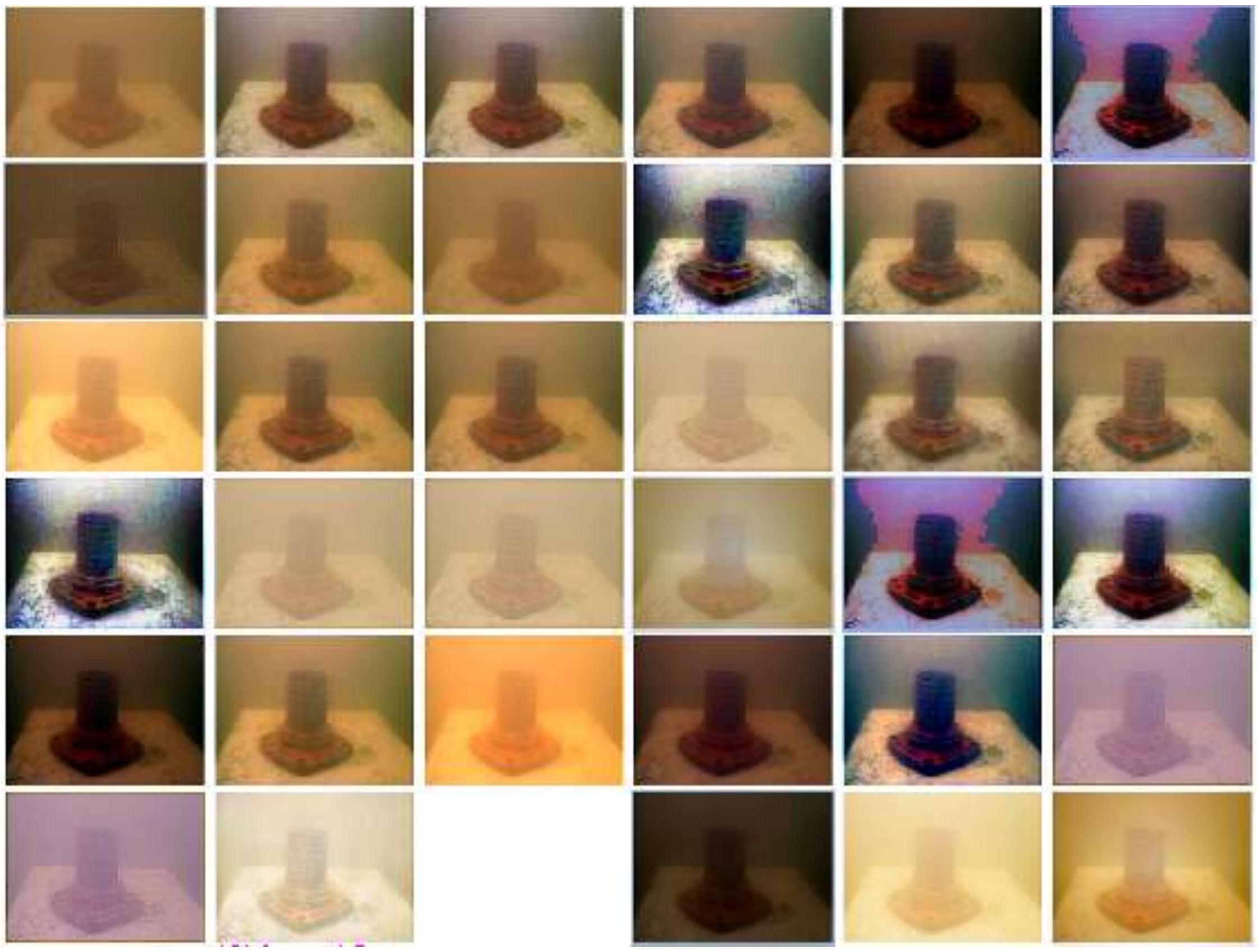

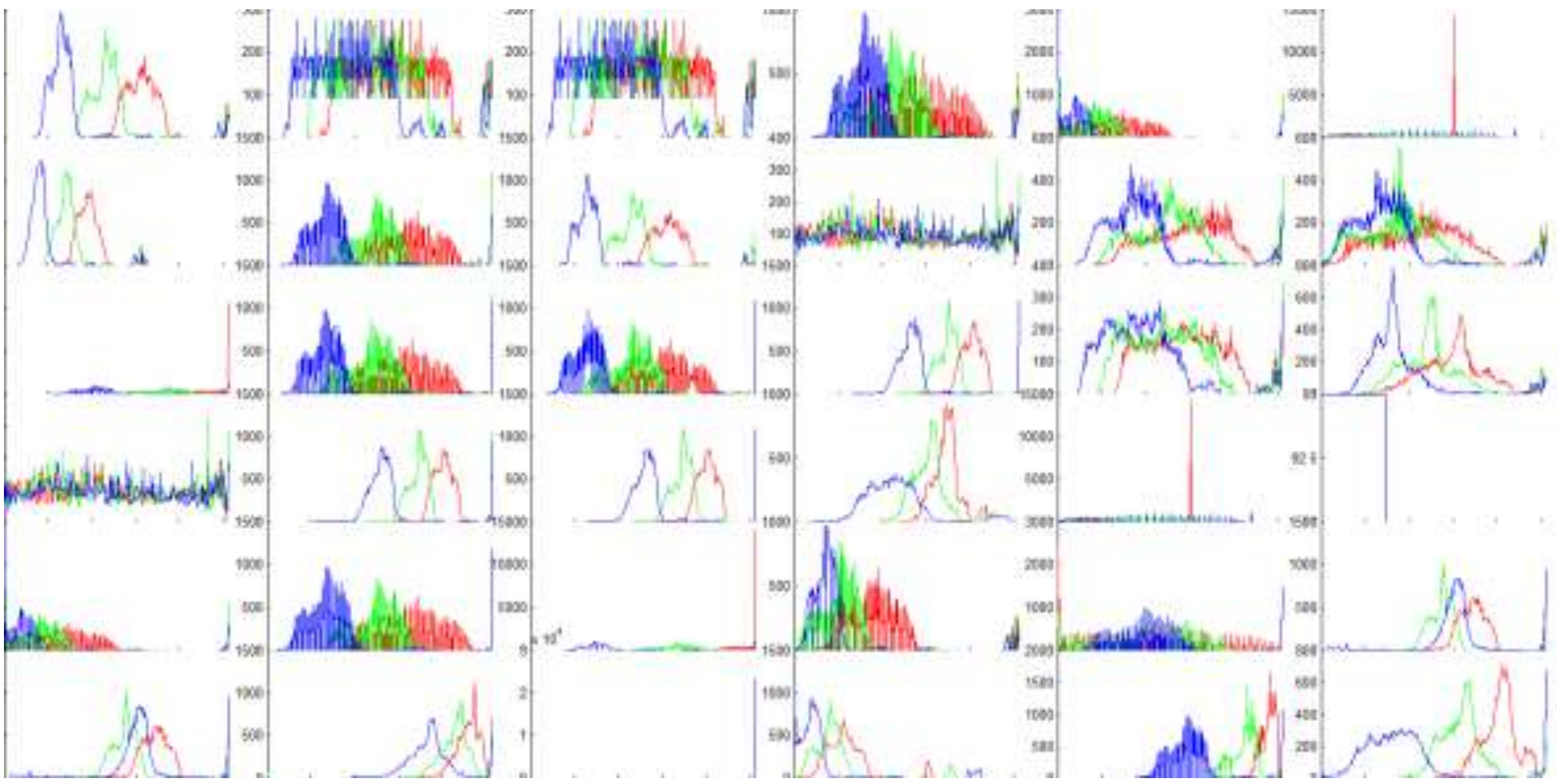

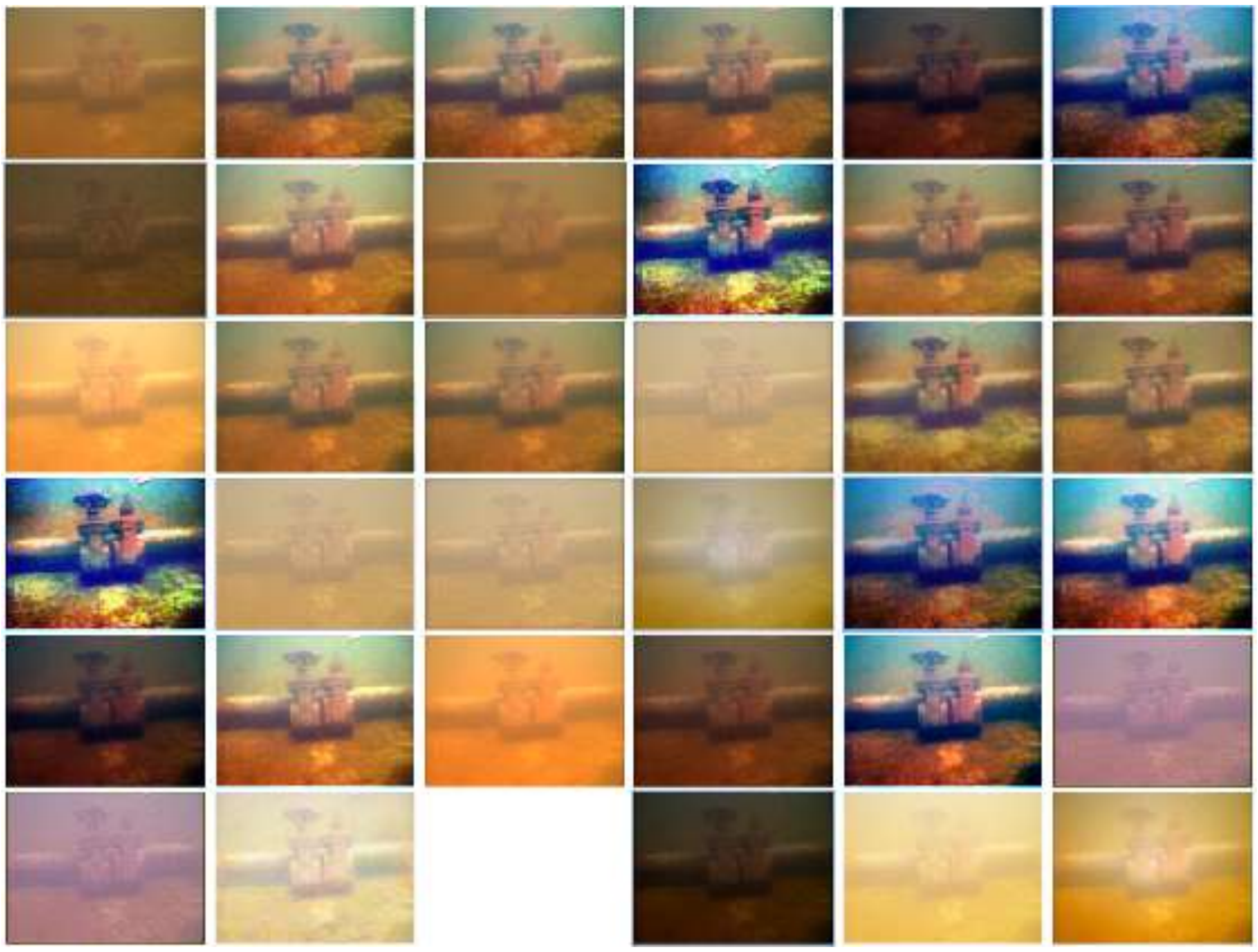

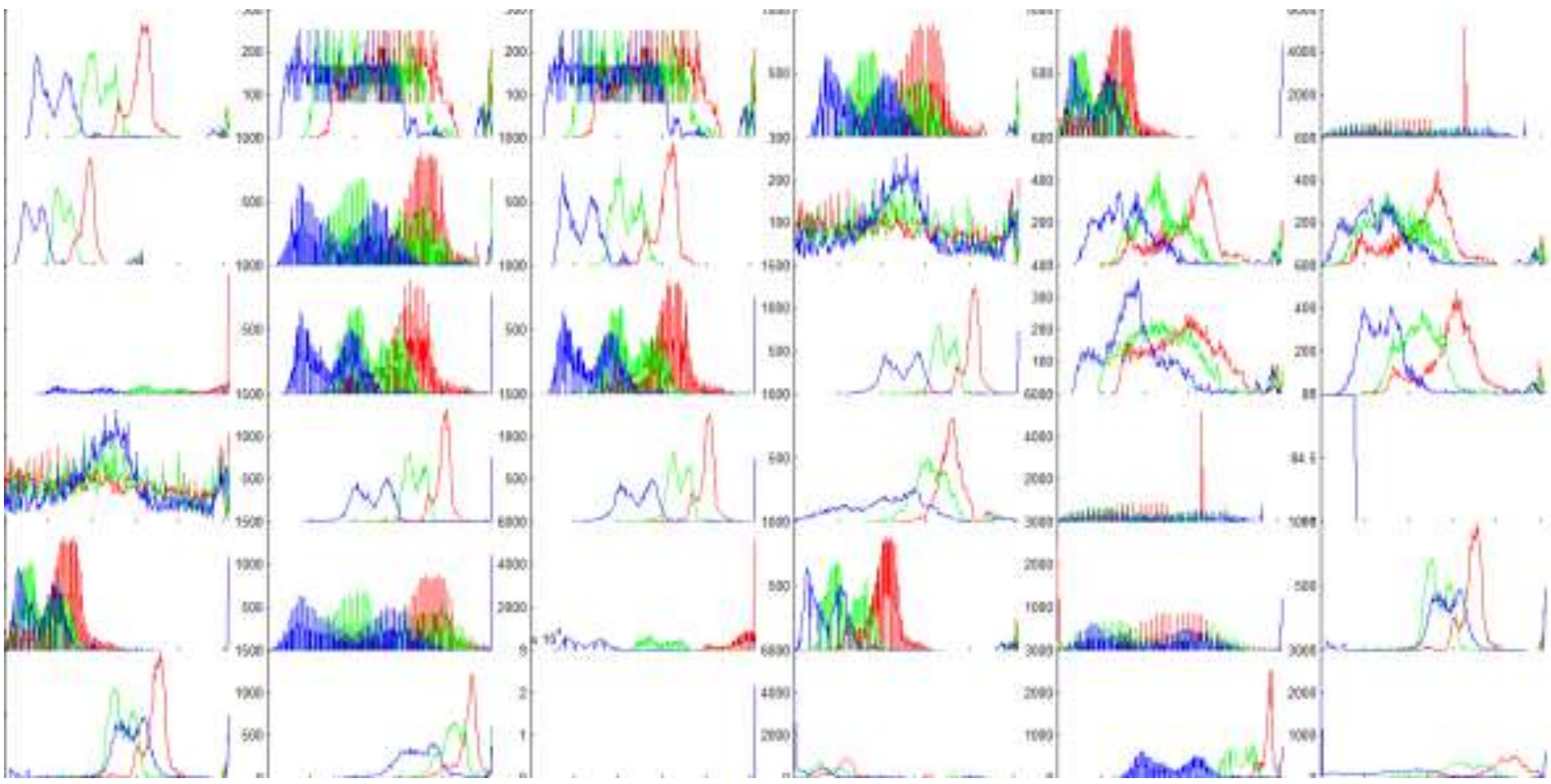

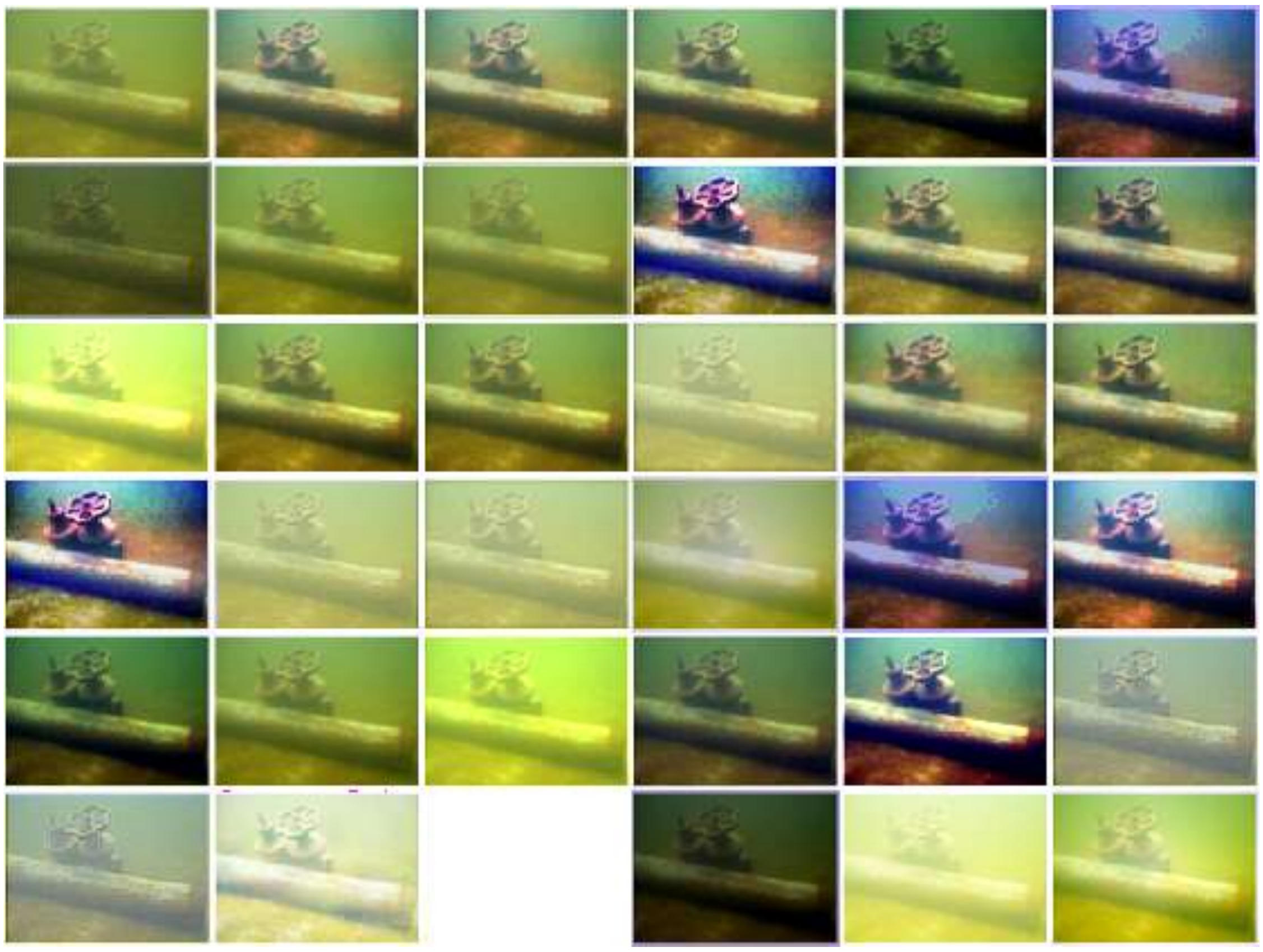

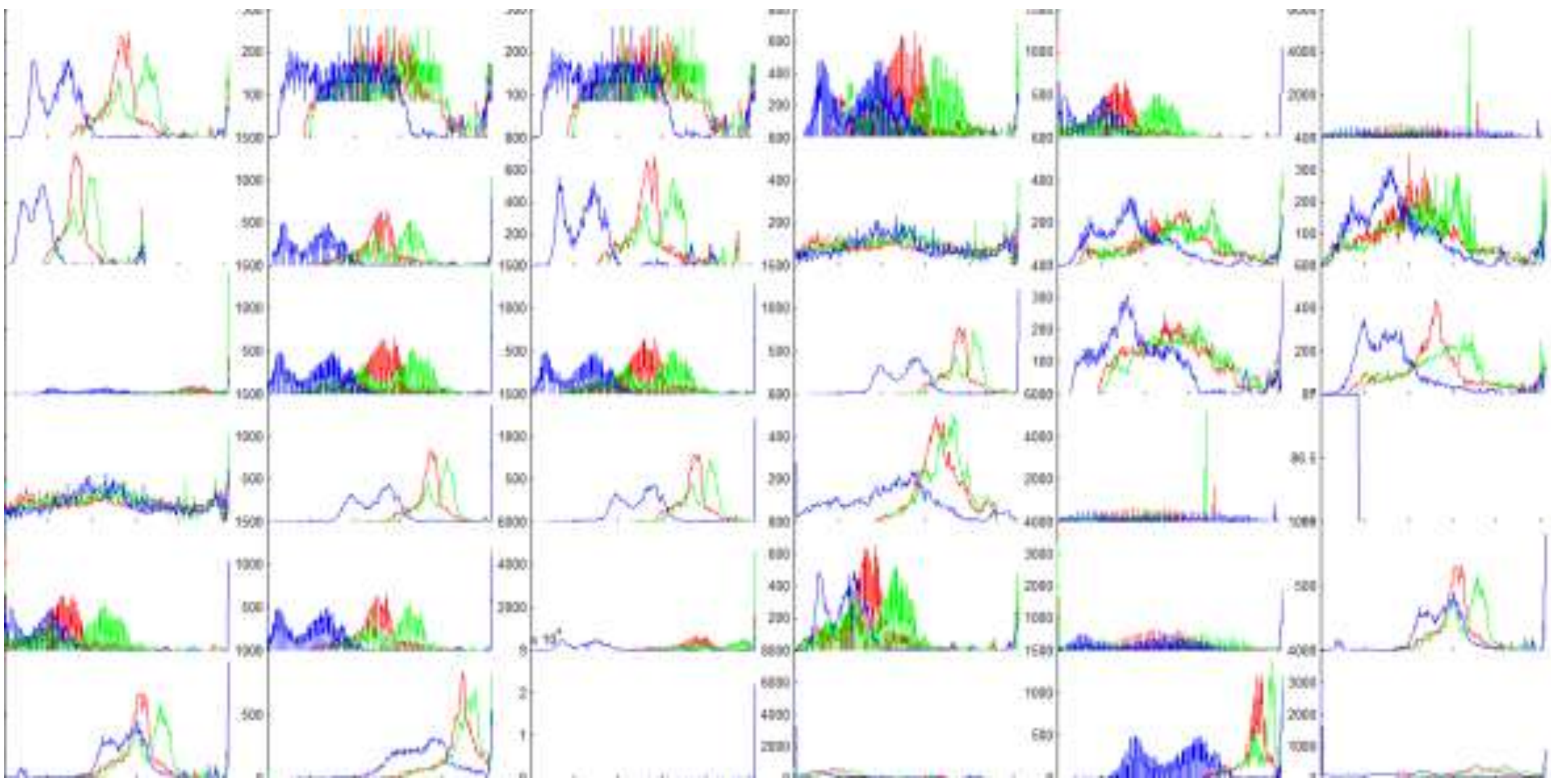

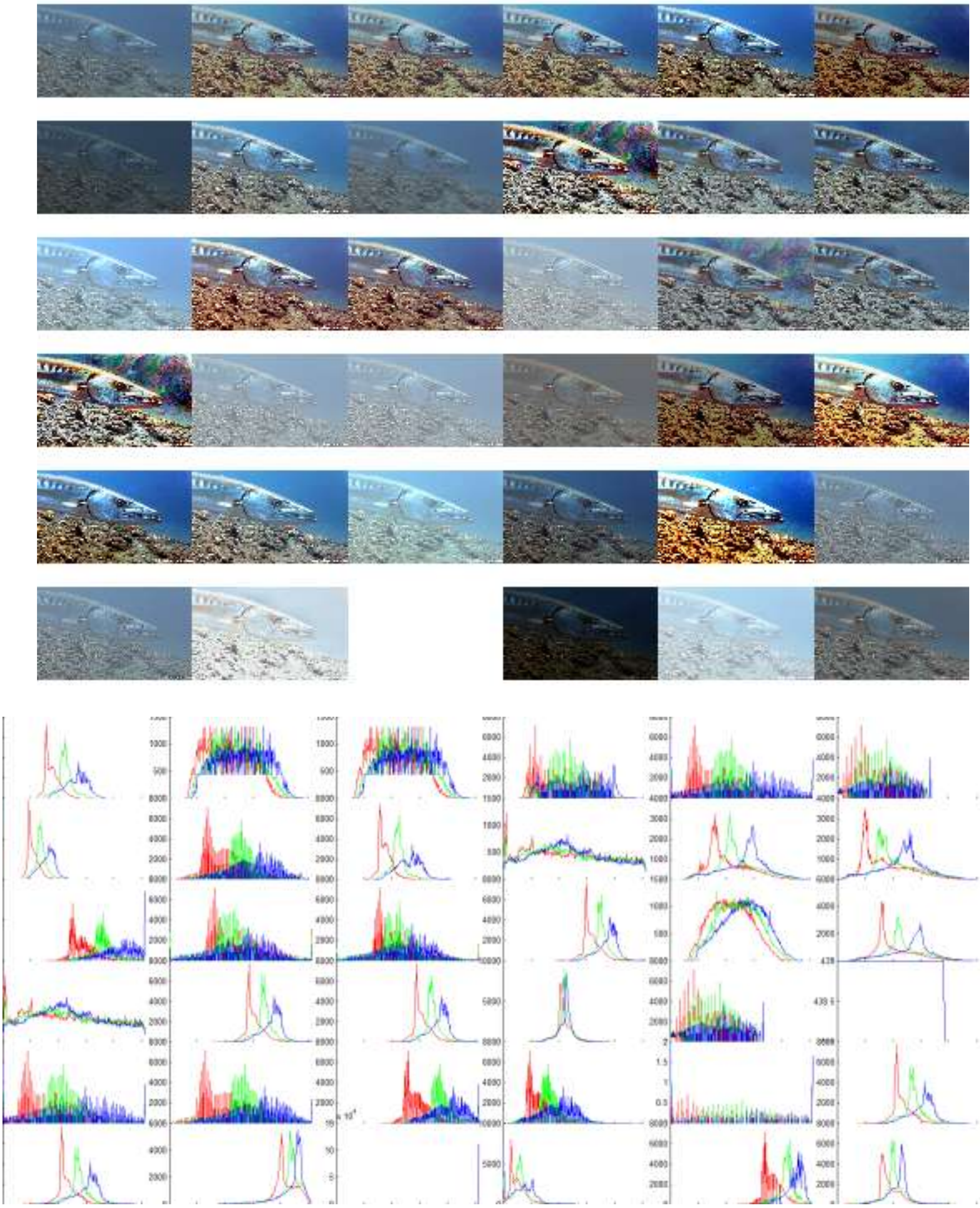

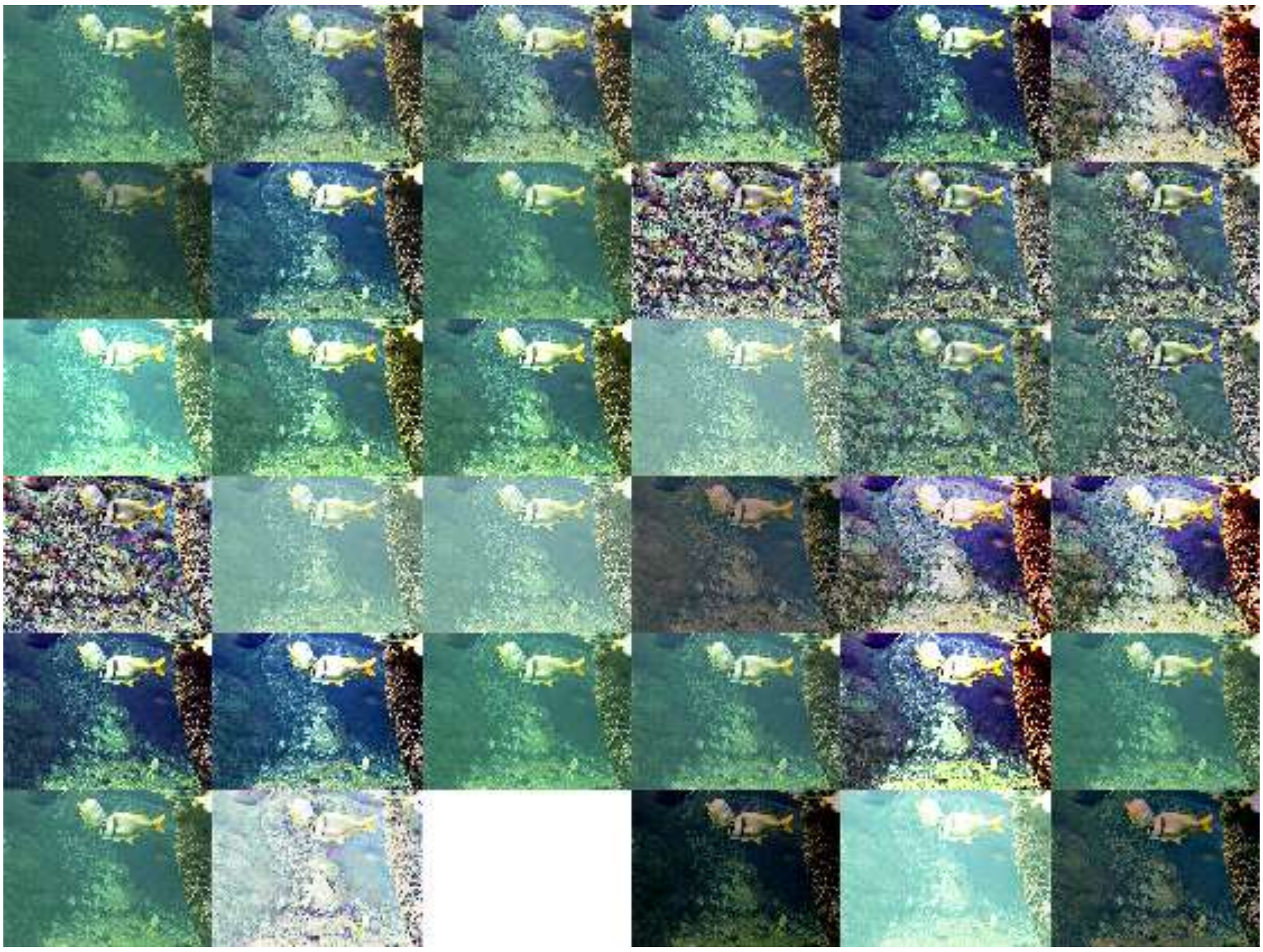

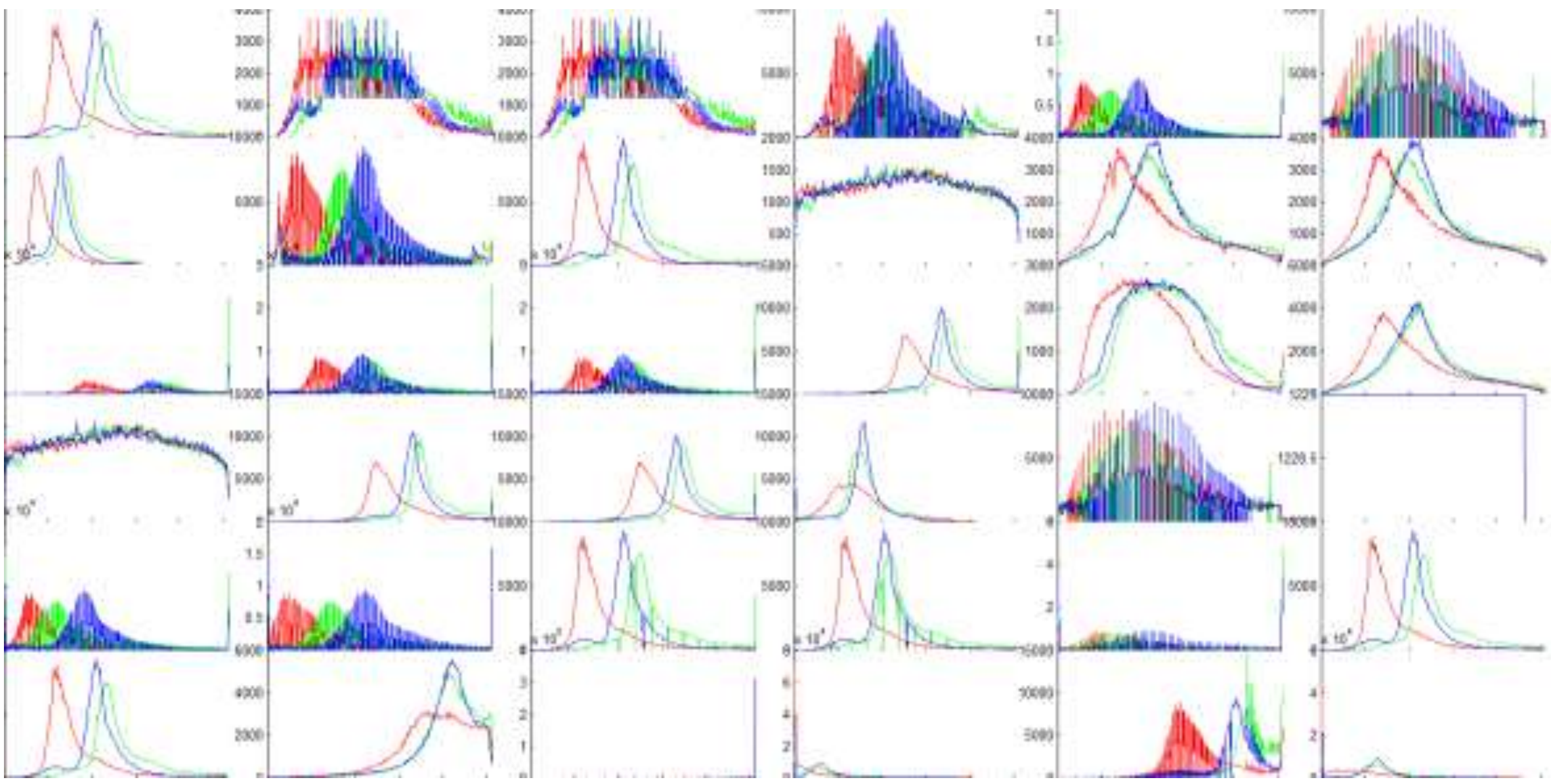

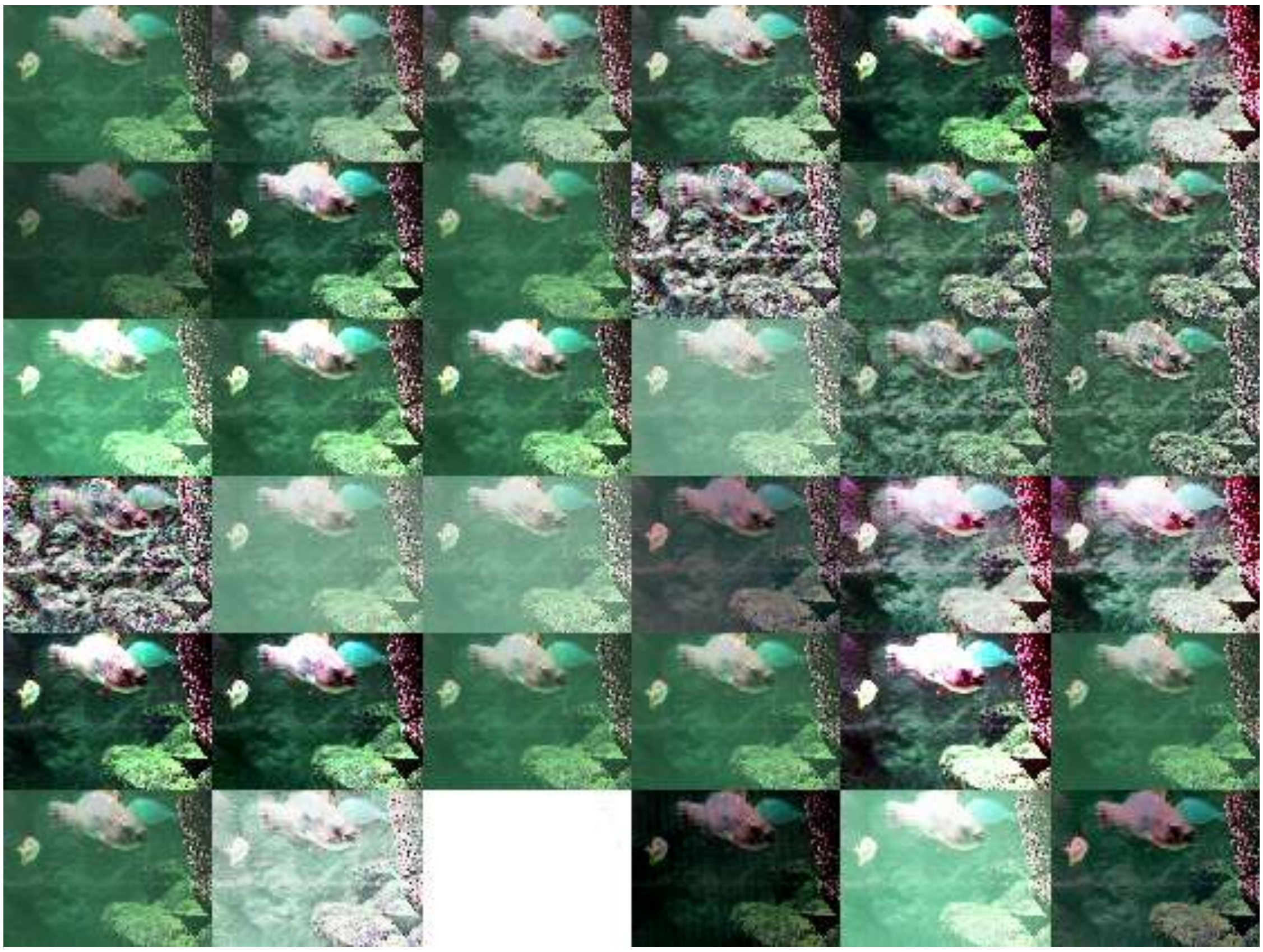

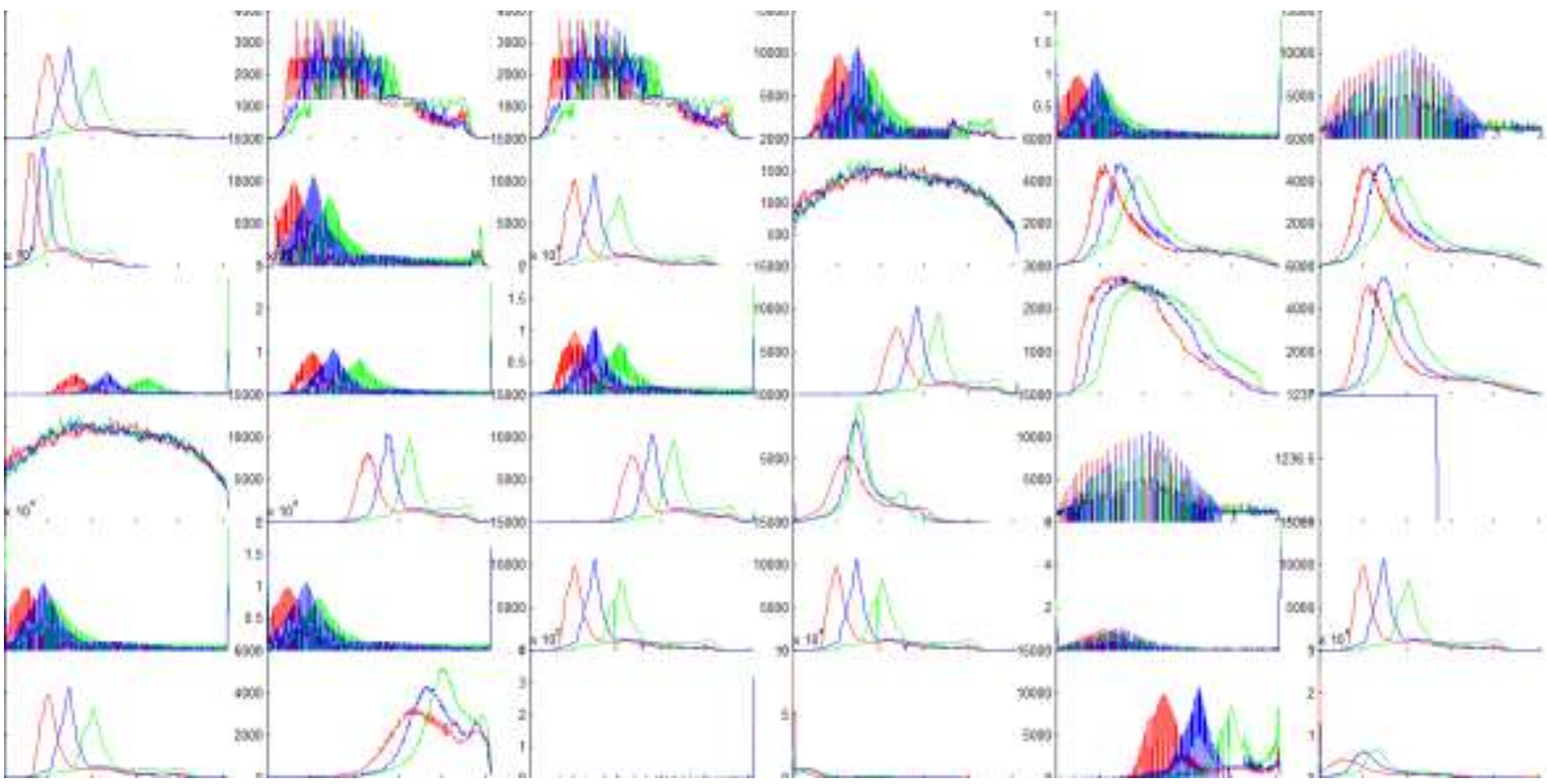

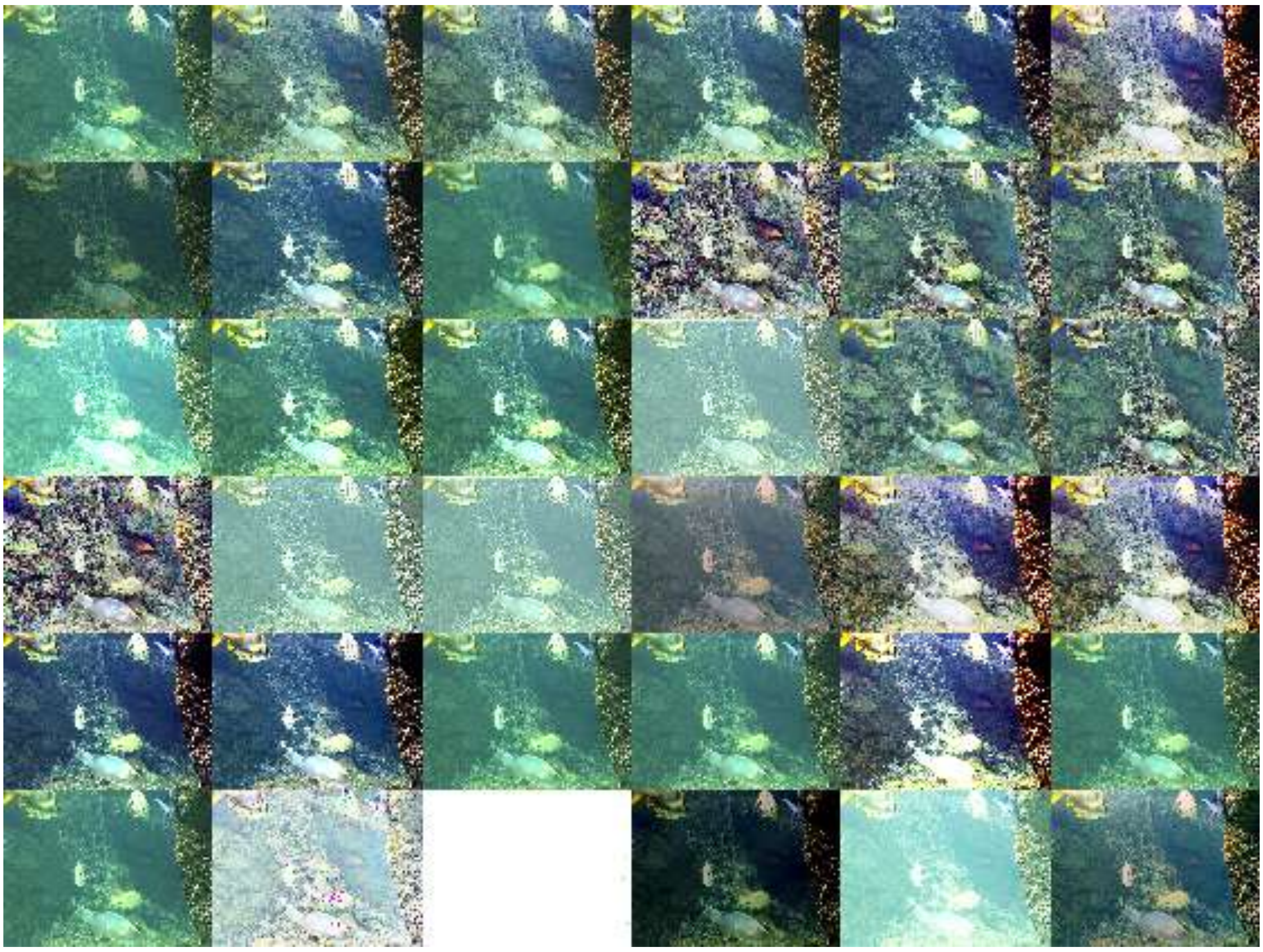

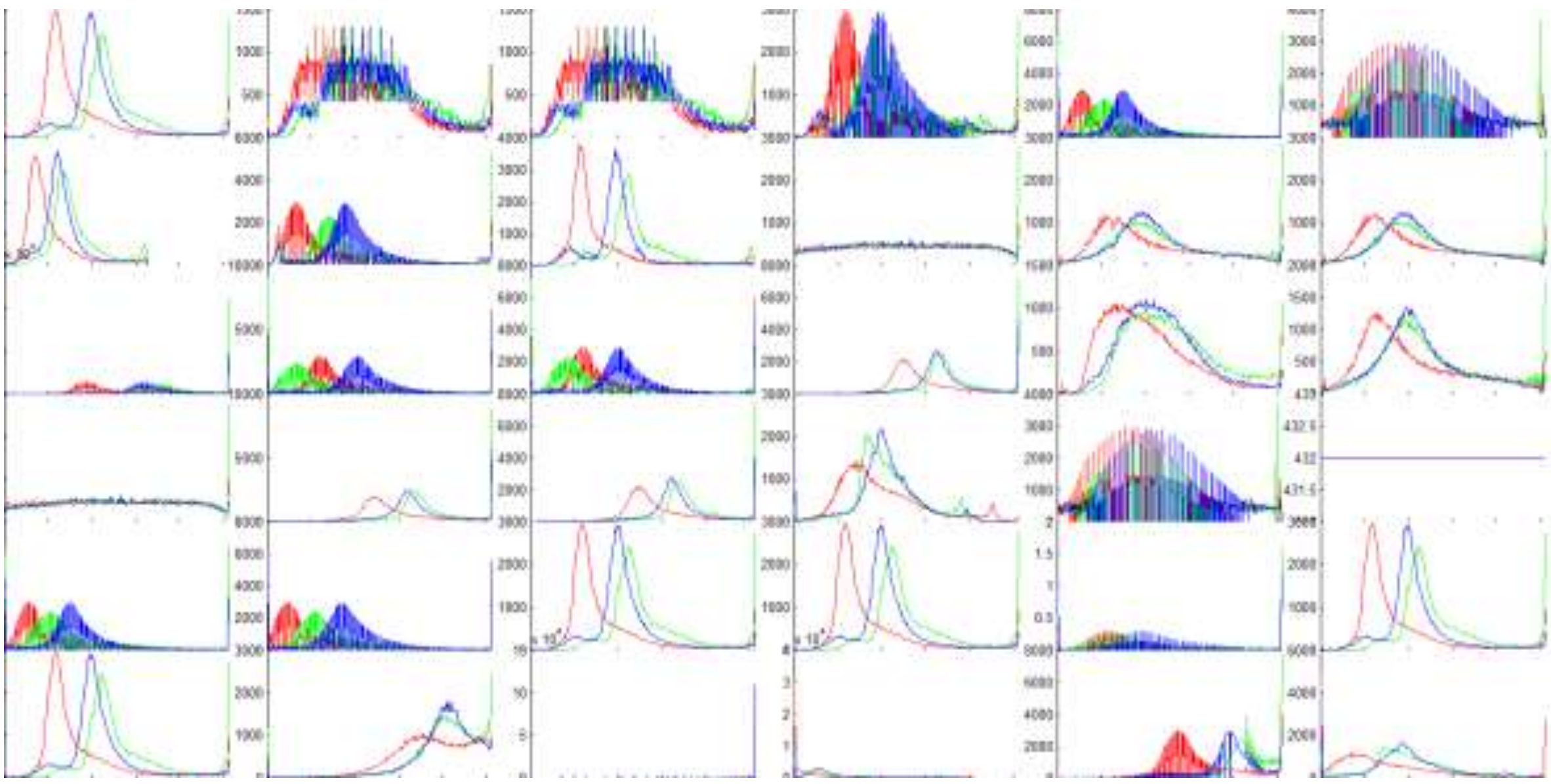

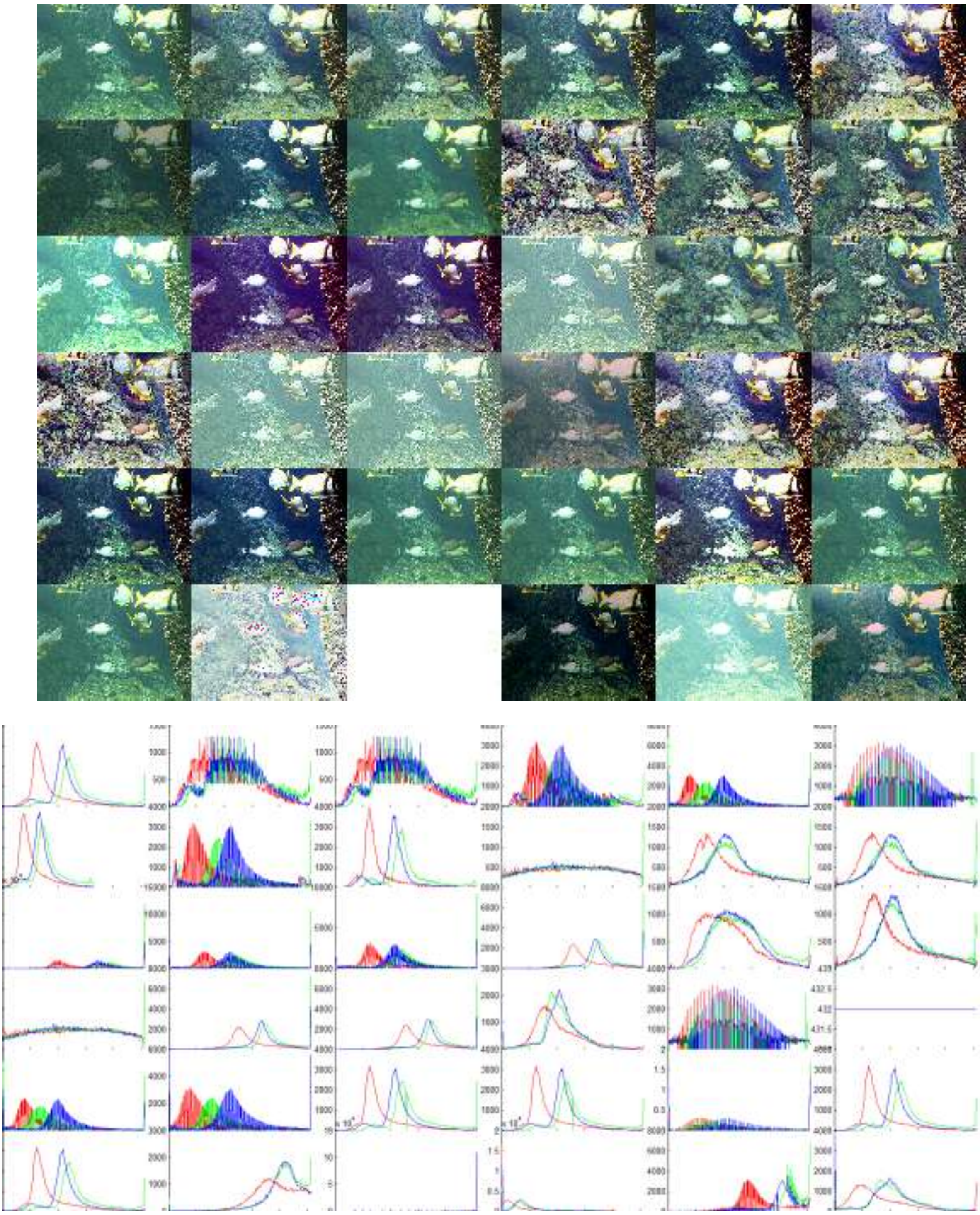

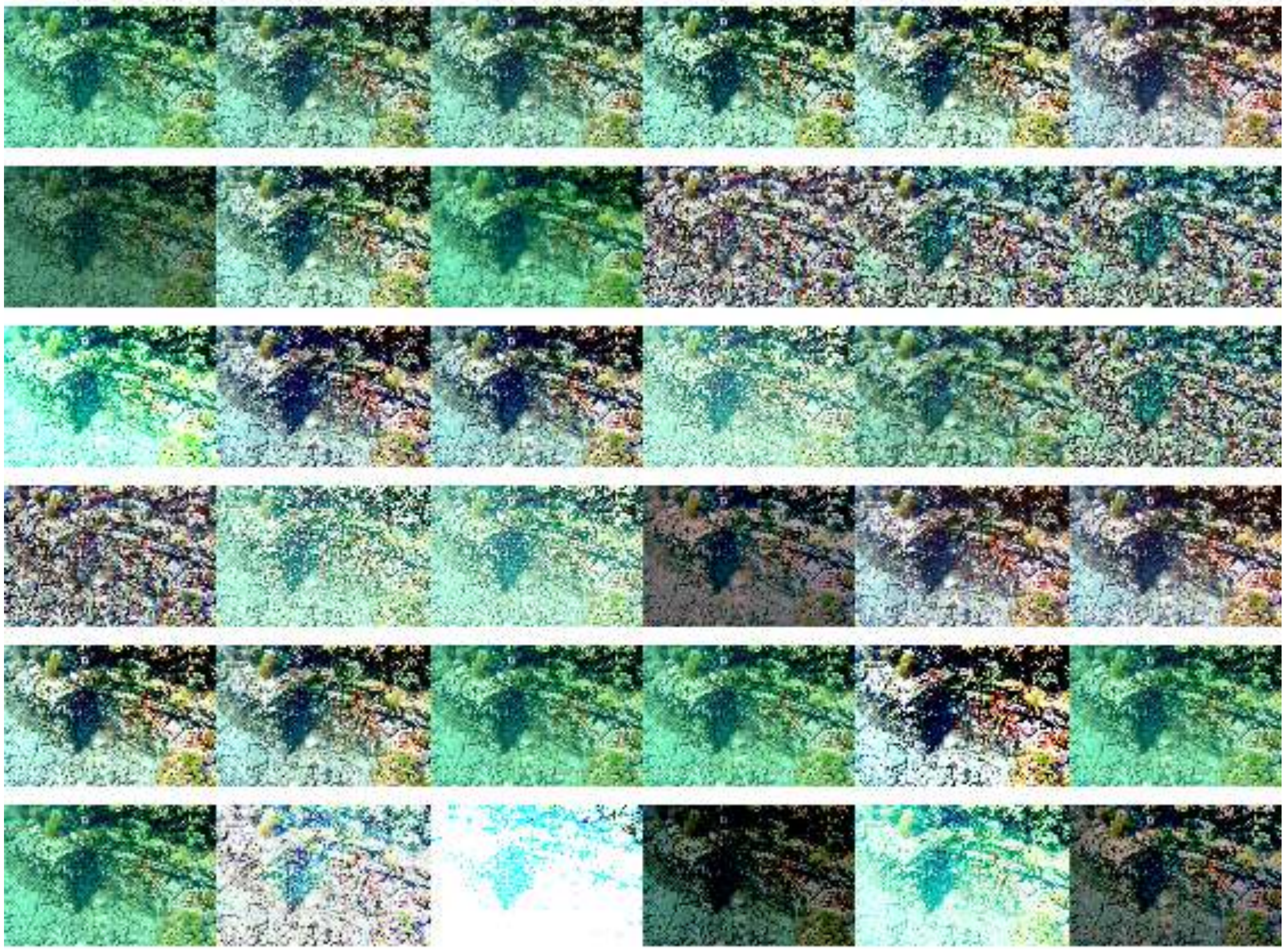

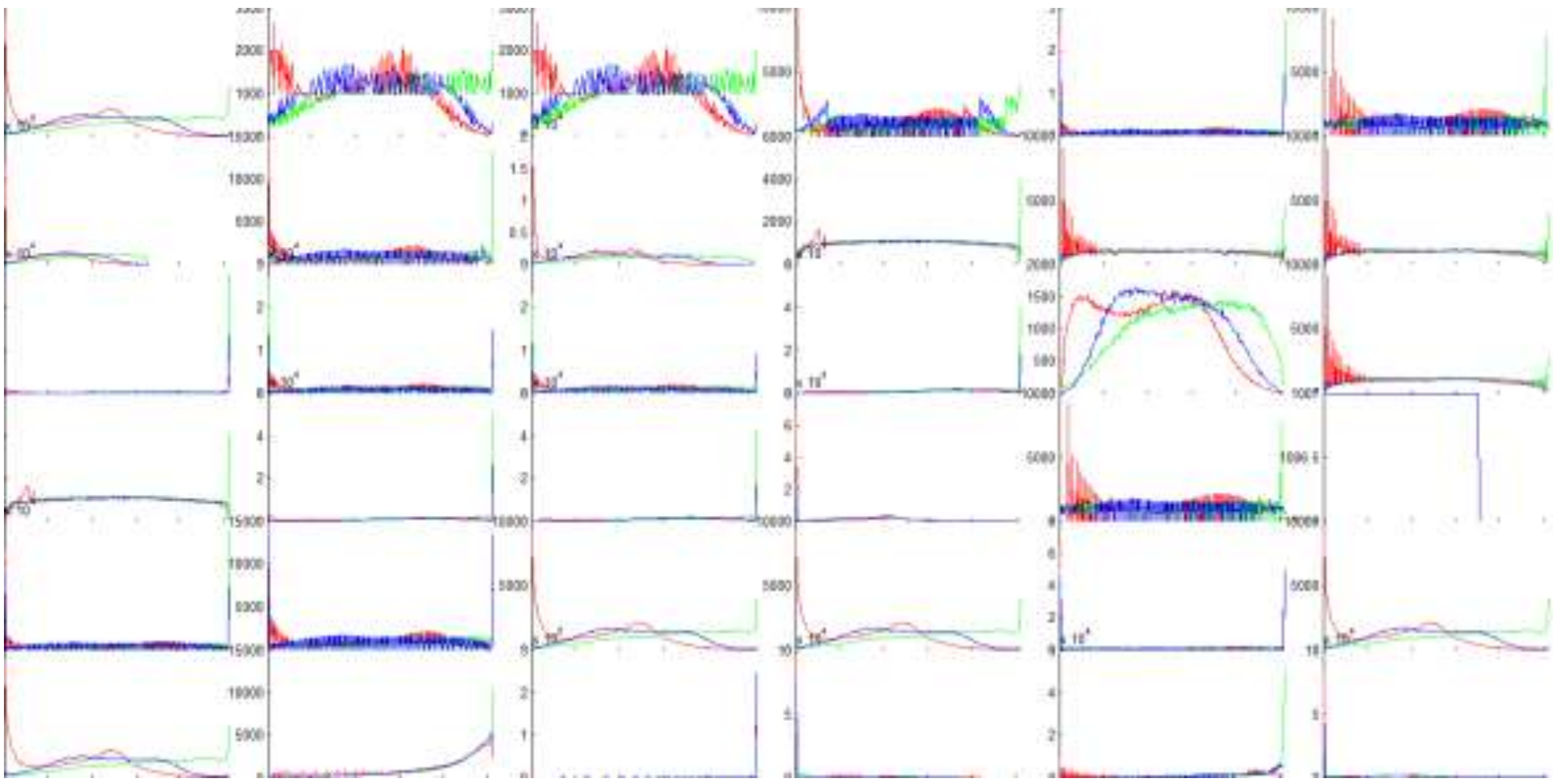

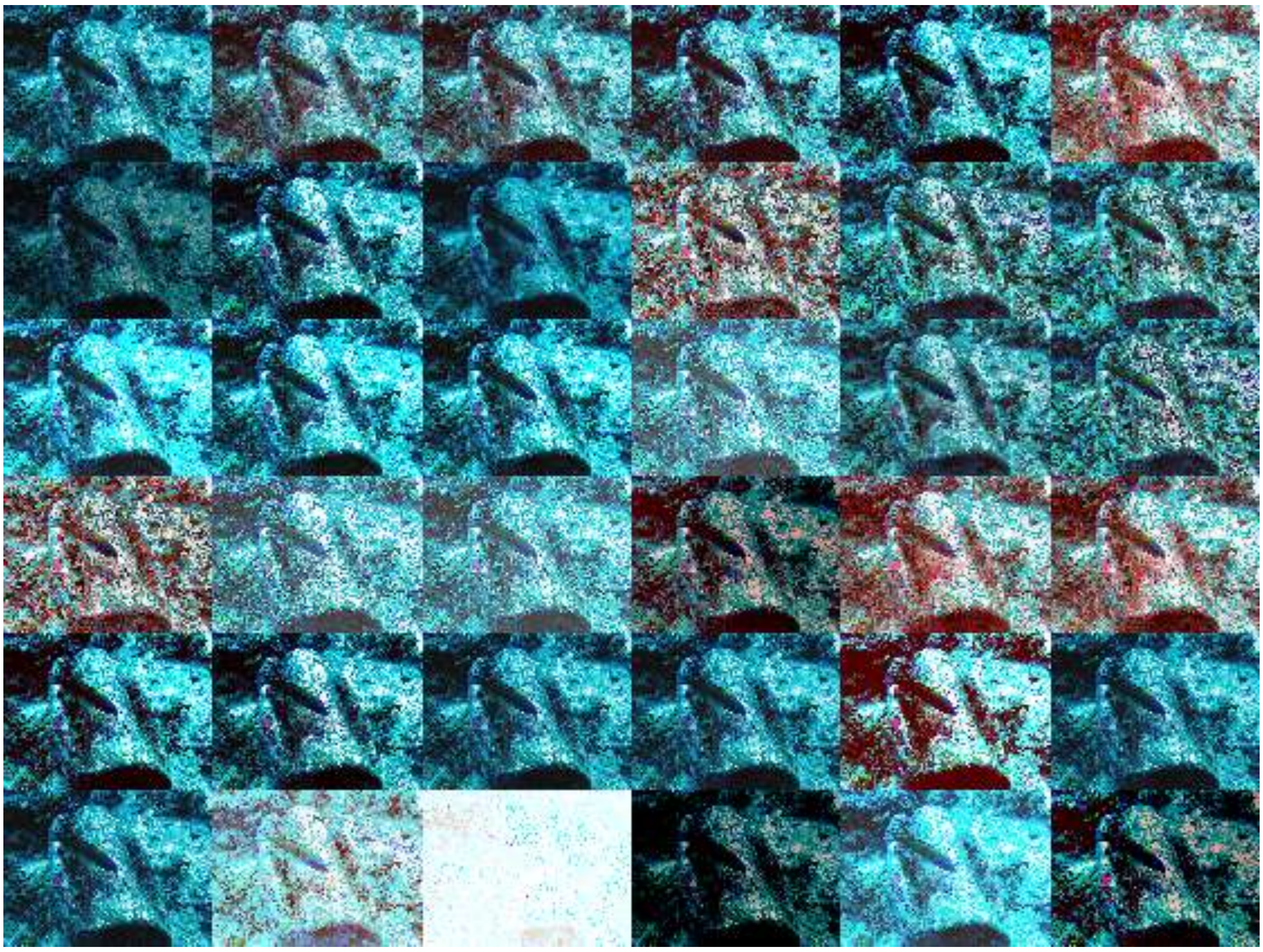

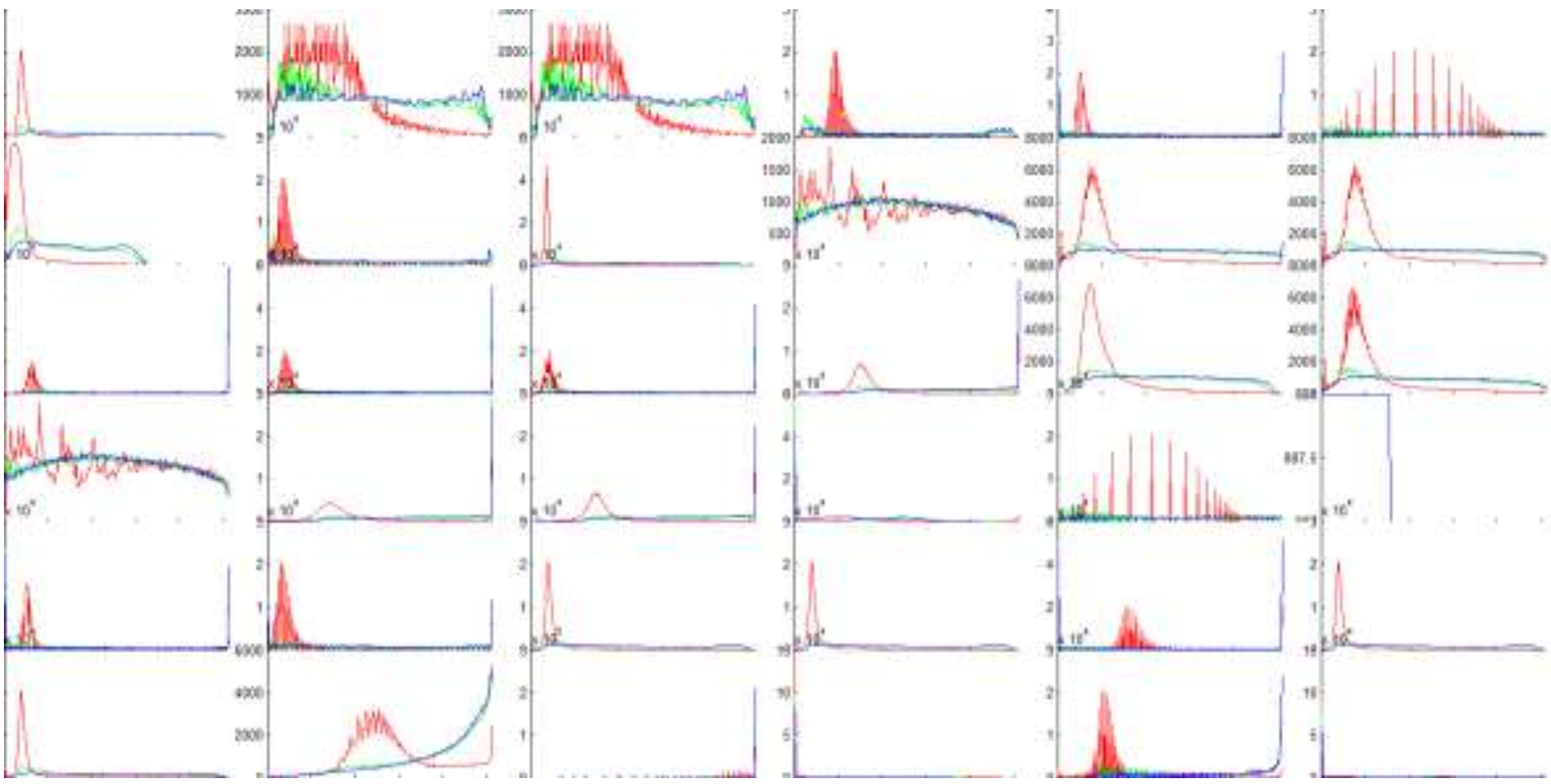

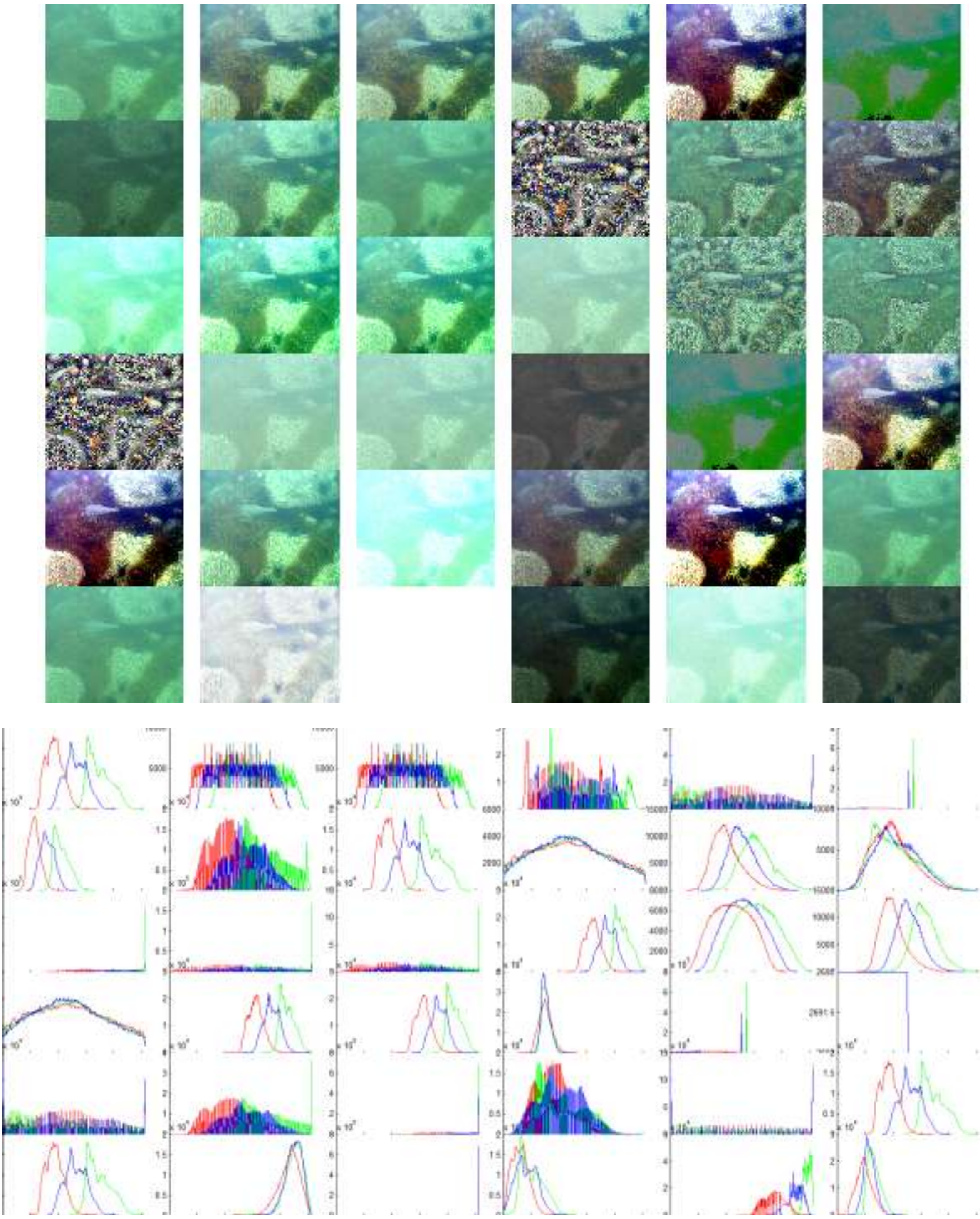

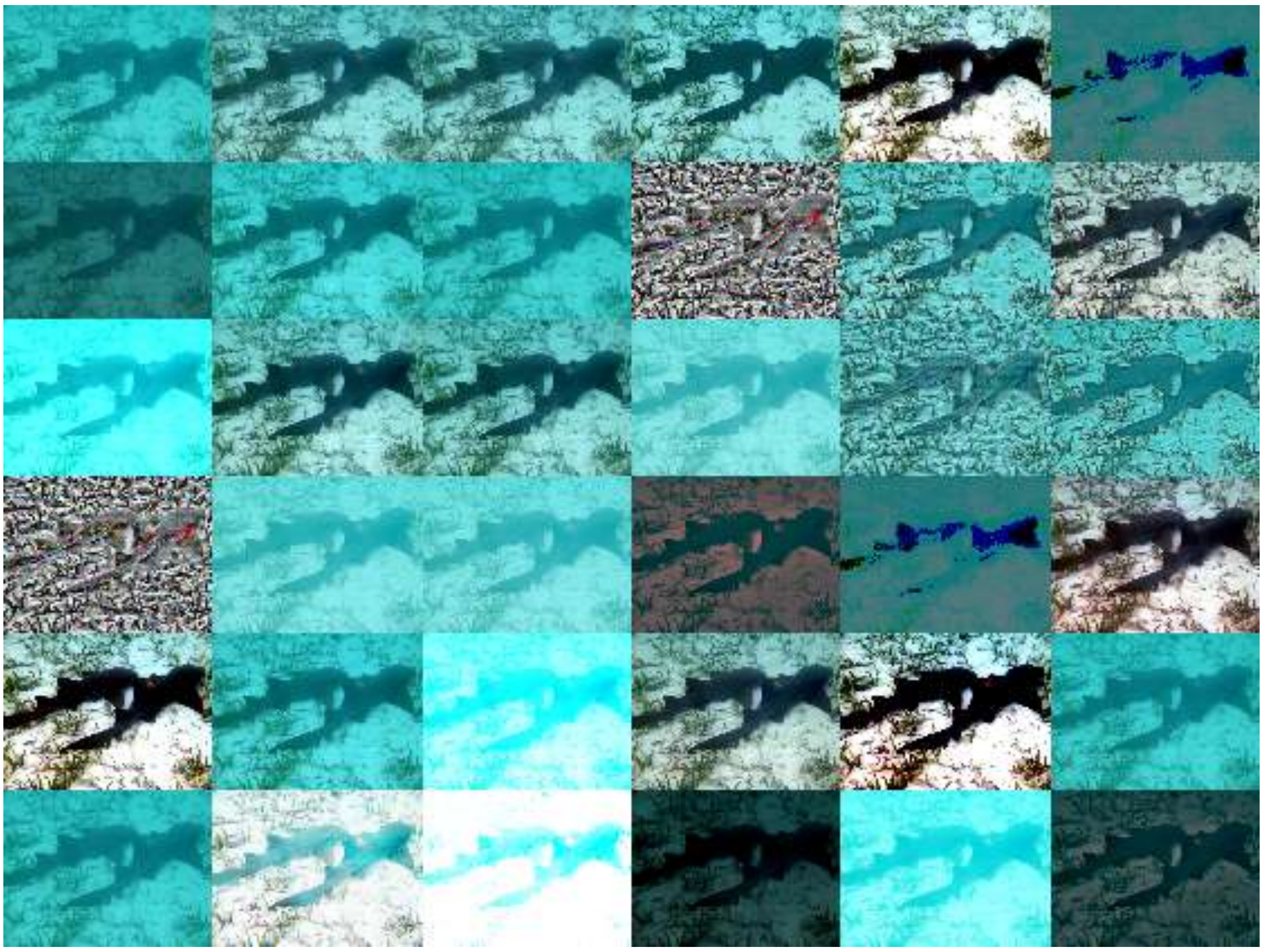

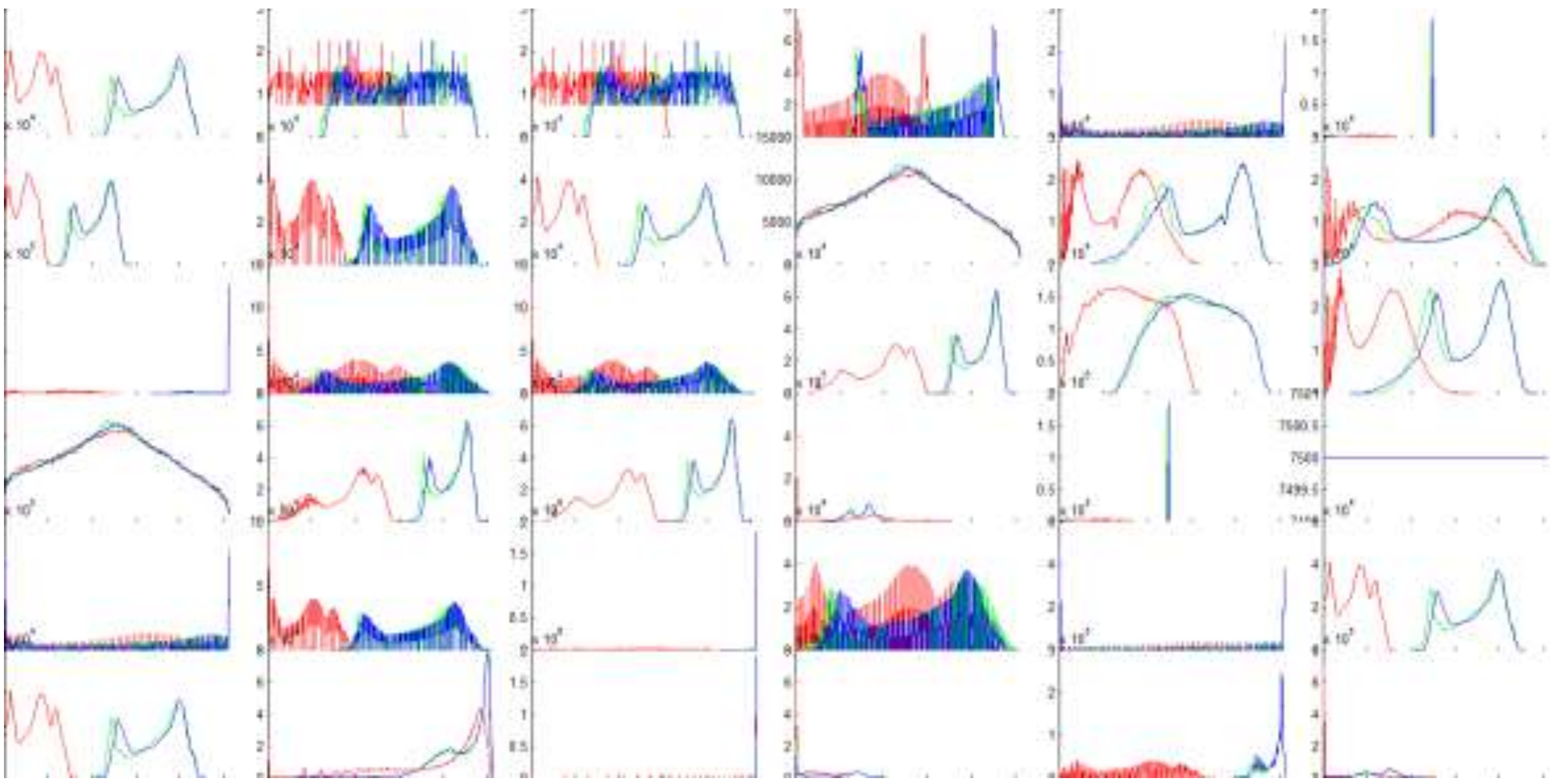

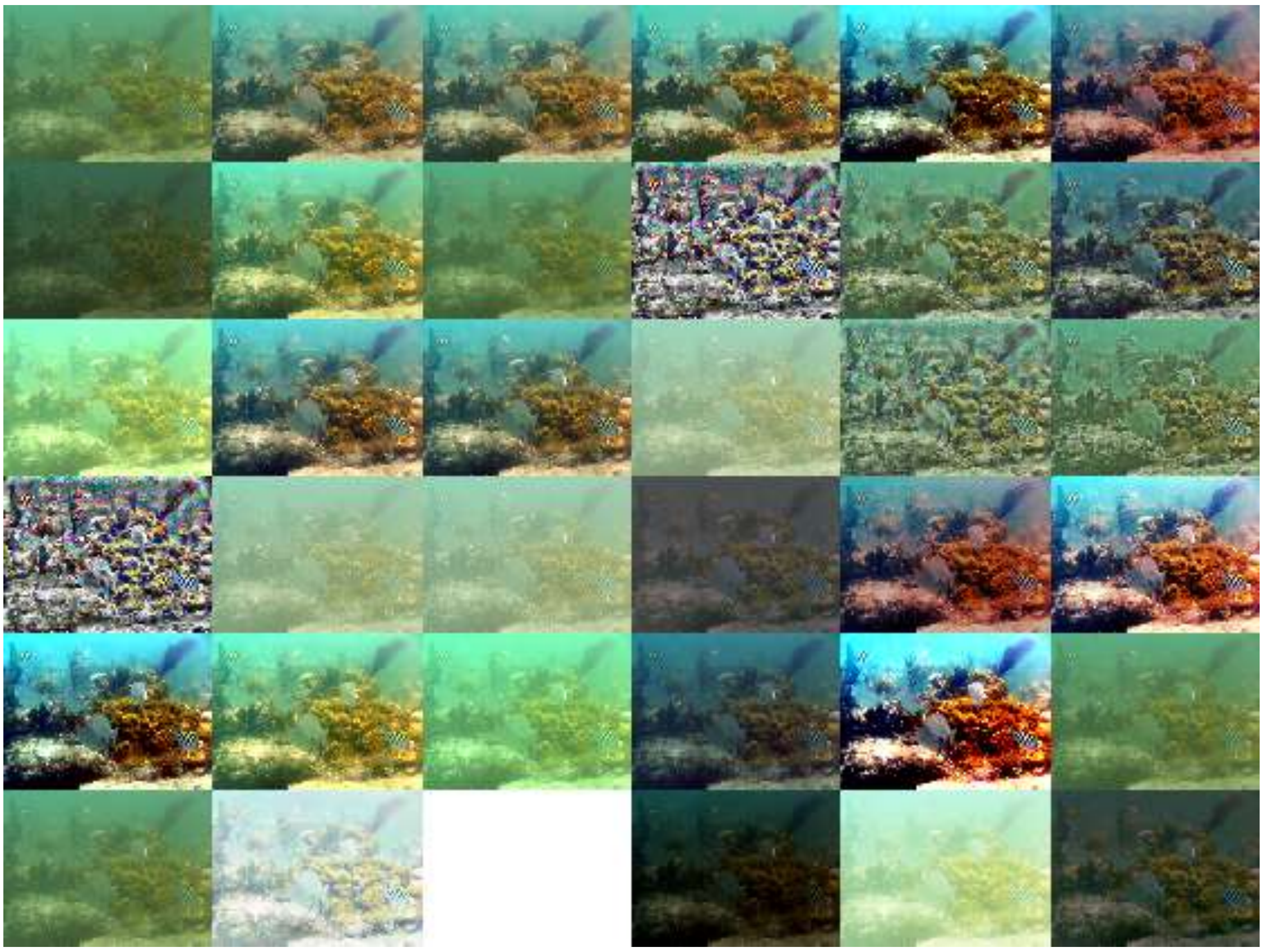

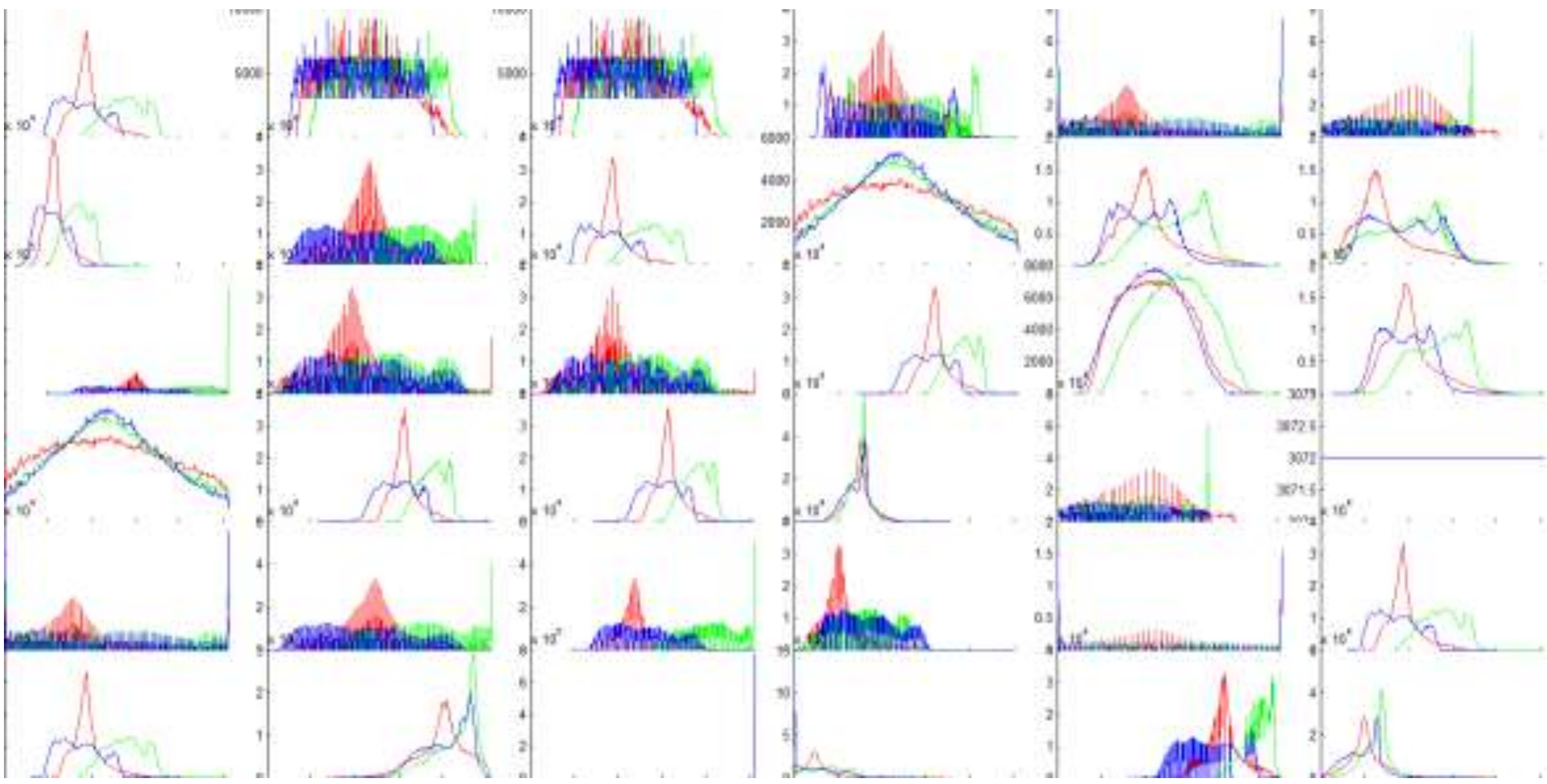

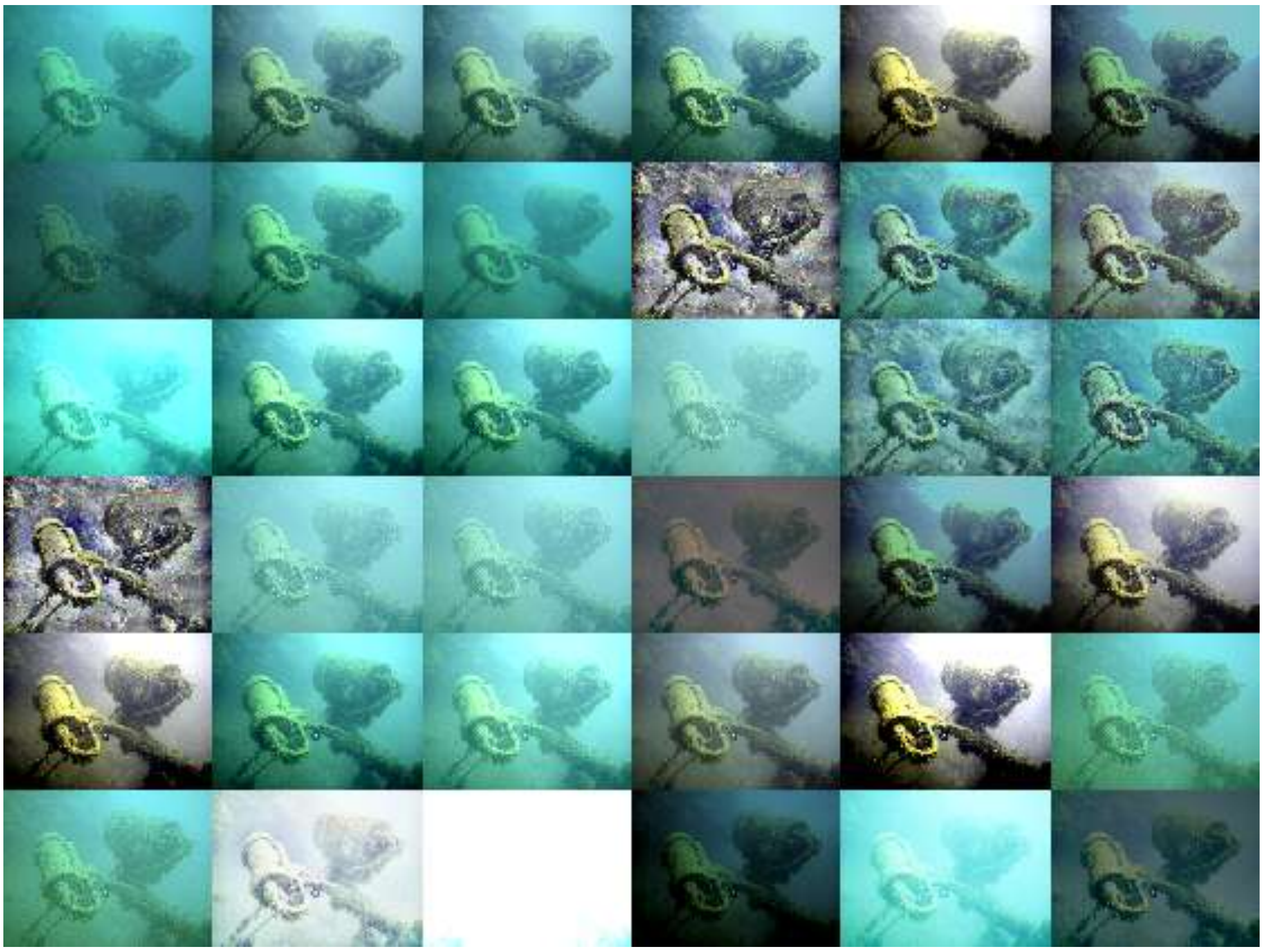

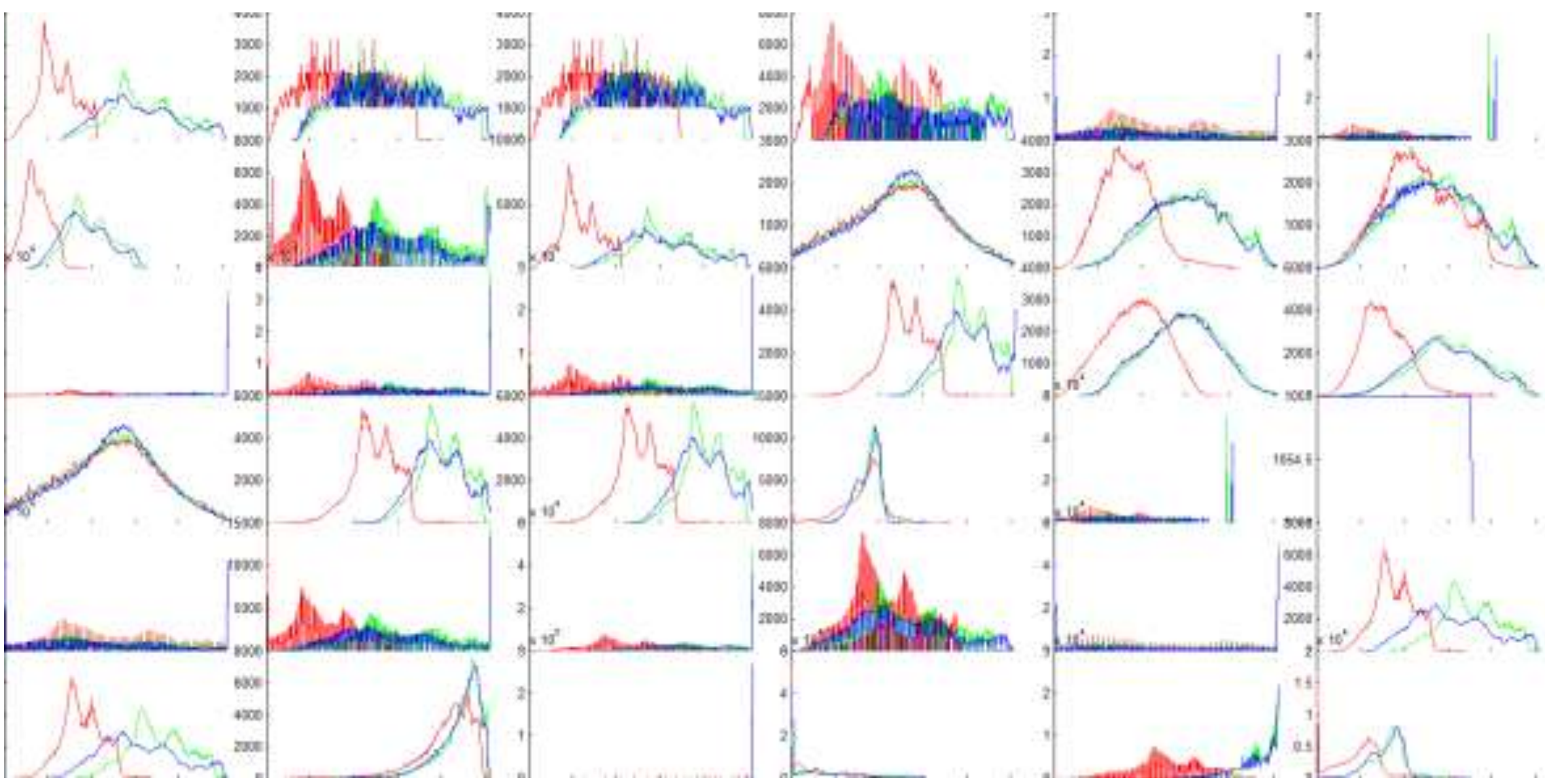

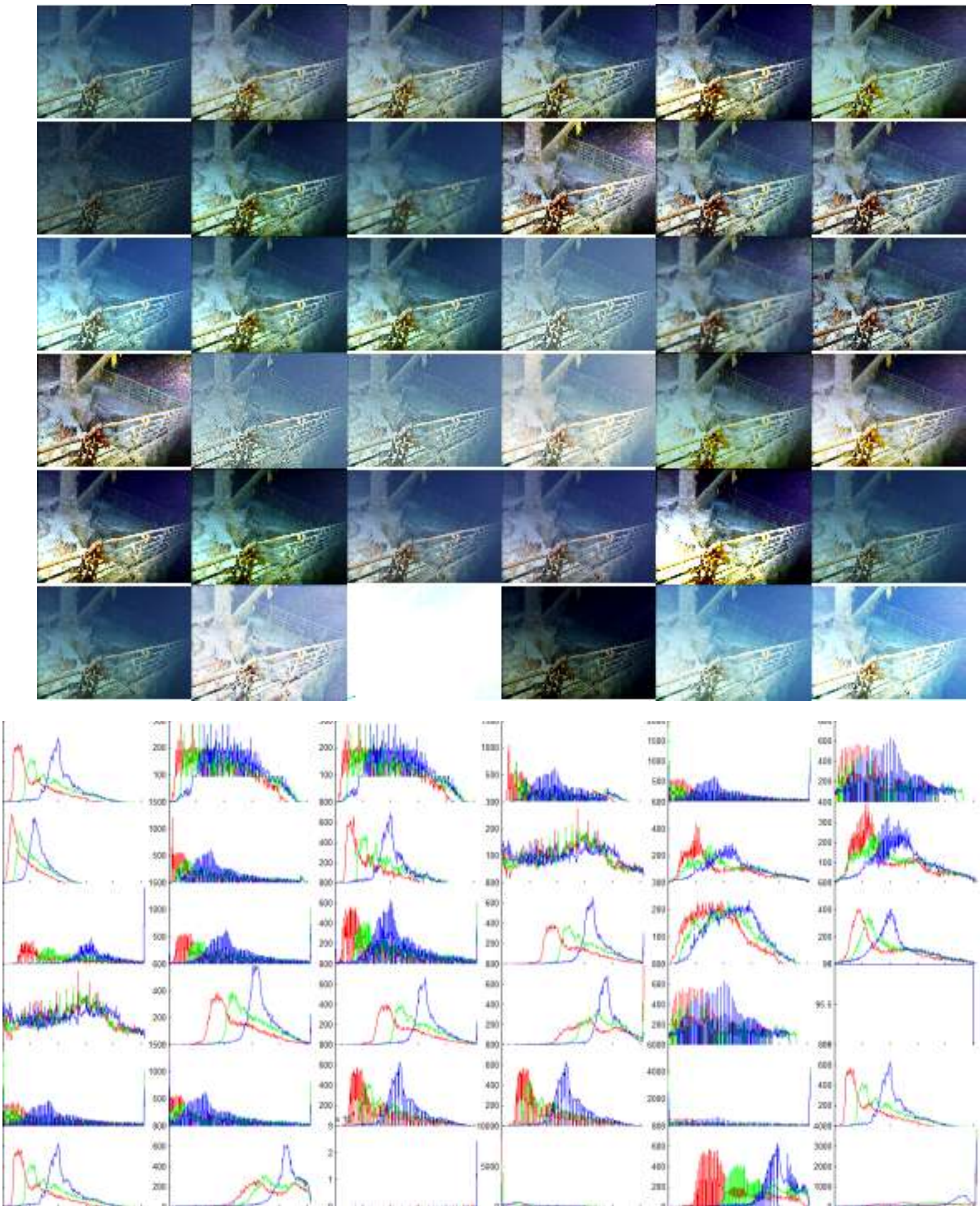

(a)

| | | | | | |
|---|---|---|---|---|---|
| Original image | PDE_HS | PDE_GOC2 | PDE_GOC3 | PDE_PWL | PDE_GHE |
| PDE_CE | PDE_CS | PDE_MINMAX | PDE_AHE | PDE_CLAHE | PA-2B |
| ADE1 | PA-1A | PA-1B | PA-1+HF | PA-2 | CLAHE |
| AHE | SHF | FDHF | MSR | GHE | HS |
| PWL | CS | GOC1 | GOC2 | GOC3 | PA-1 (few iter) |
| PA-1 (few iter) | GUM | SSR | SSAR | TC | MSRCR |

(b)

**Fig. 8** (a) Underwater images processed with various algorithms compared with the proposed approaches and corresponding histograms (b) key to figures

### 4.1 Illumination correction

In this section we compare the proposed algorithm with the CLAHE and Homomorphic filters and show that the proposed approaches lead to a wide range of results based on the amount of contribution is allowed by the anisotropic diffusion term and the contrast term. In Fig. 9, we observe the effects of parameter adjustments for PA-1 and PA-2.

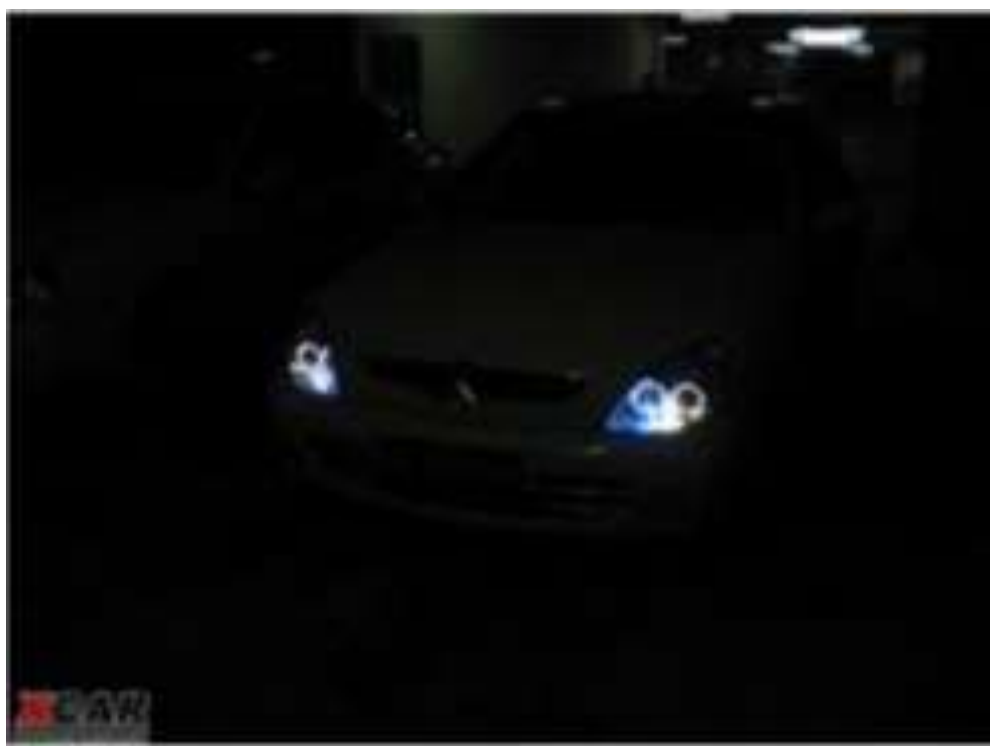

Original image

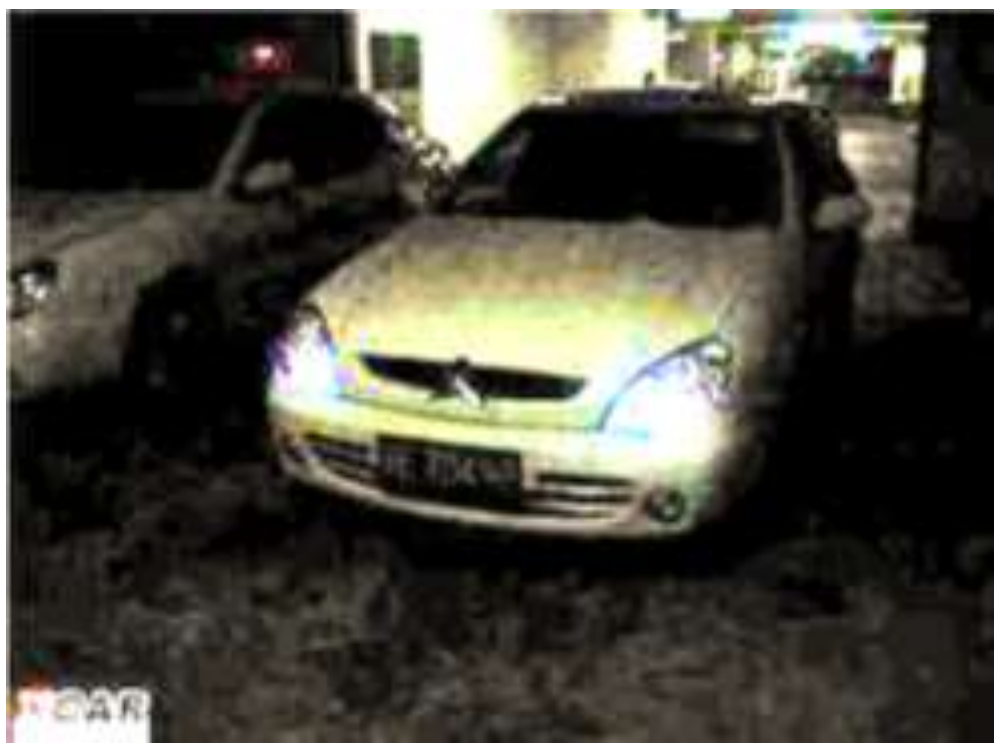

60 iterations using PA-1 dt = 1/8, lambda = 0.001

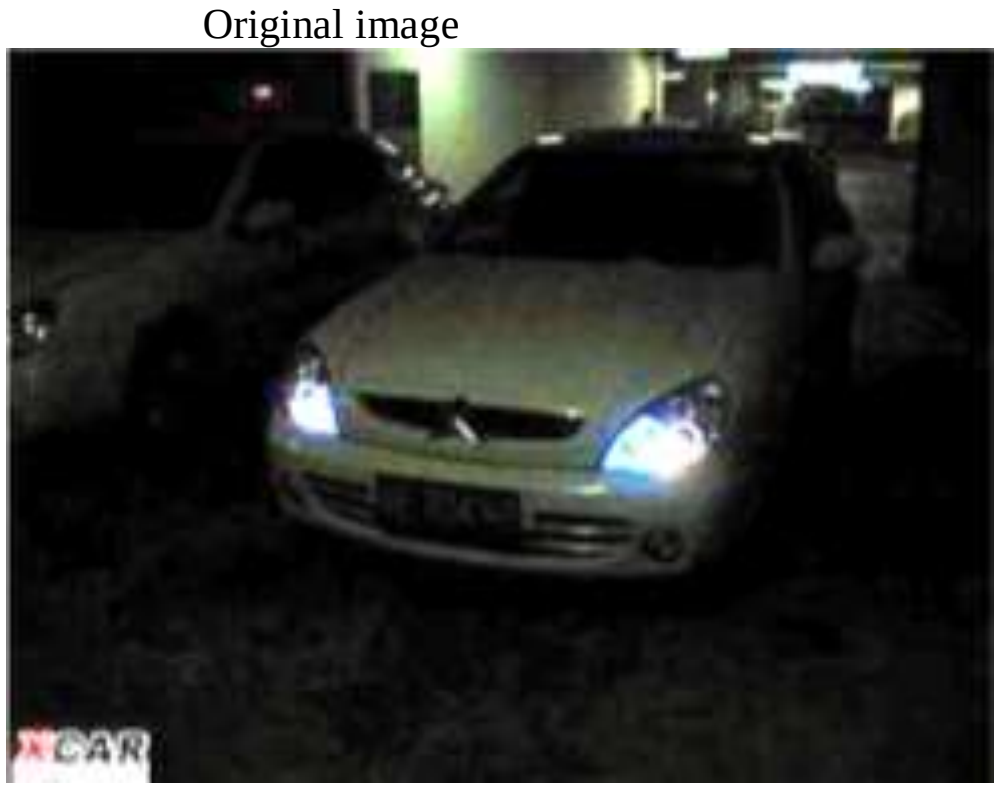

60 iterations using PA-1 dt = 1/4, lambda = 0.01

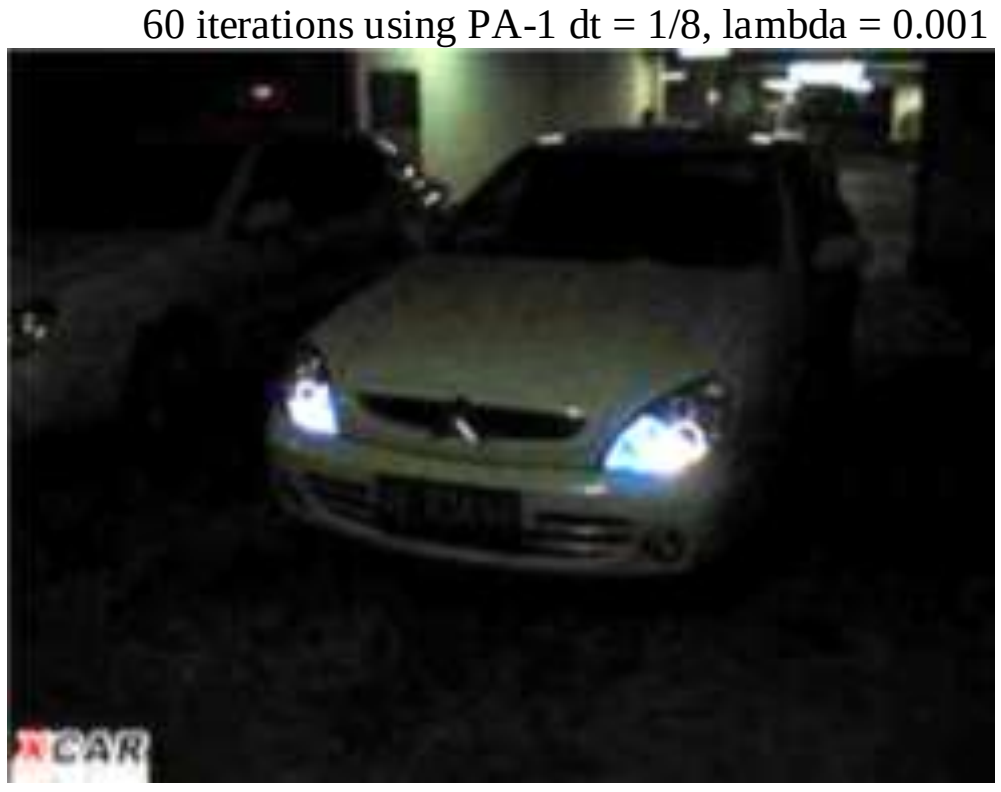

100 iterations using PA-1 dt = 1/4, lambda = 0.01

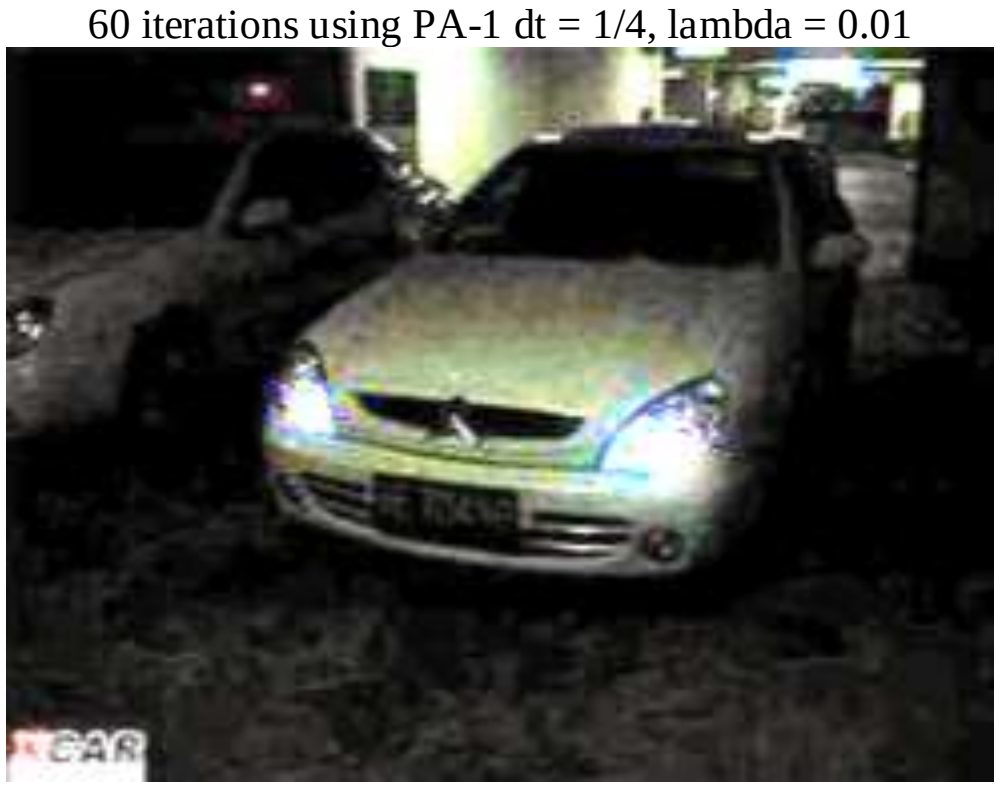

80 iterations using PA-1 dt = 1/4, lambda = 0.01

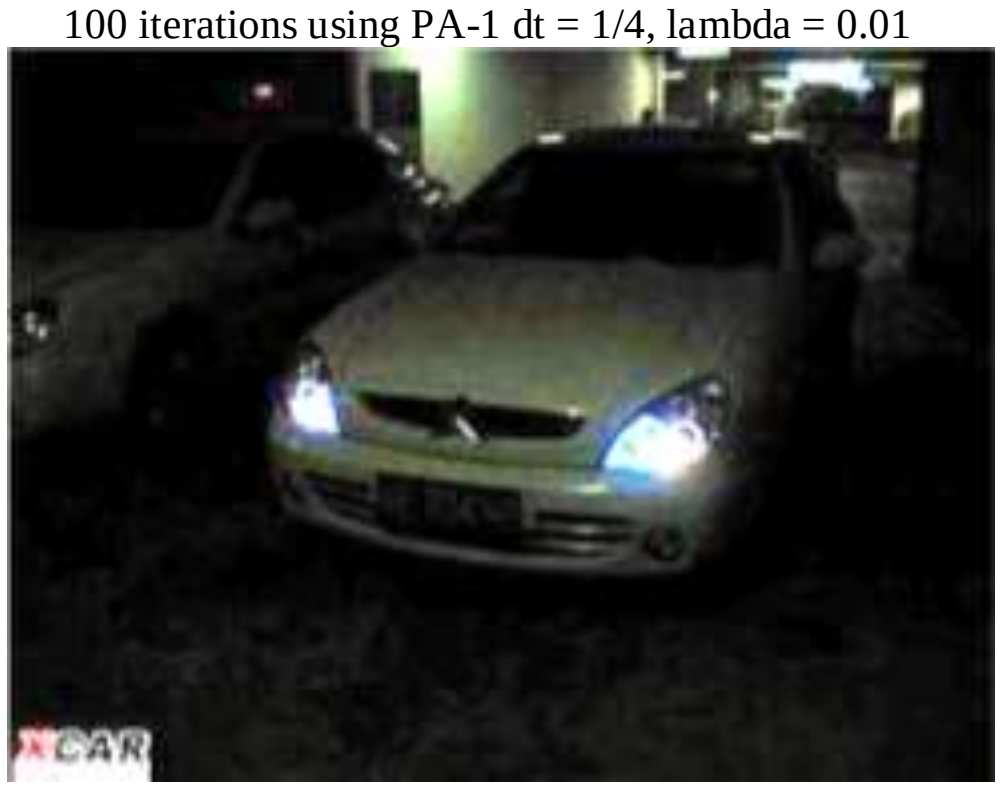

120 iterations using PA-1 dt = 1/4, lambda = 0.01

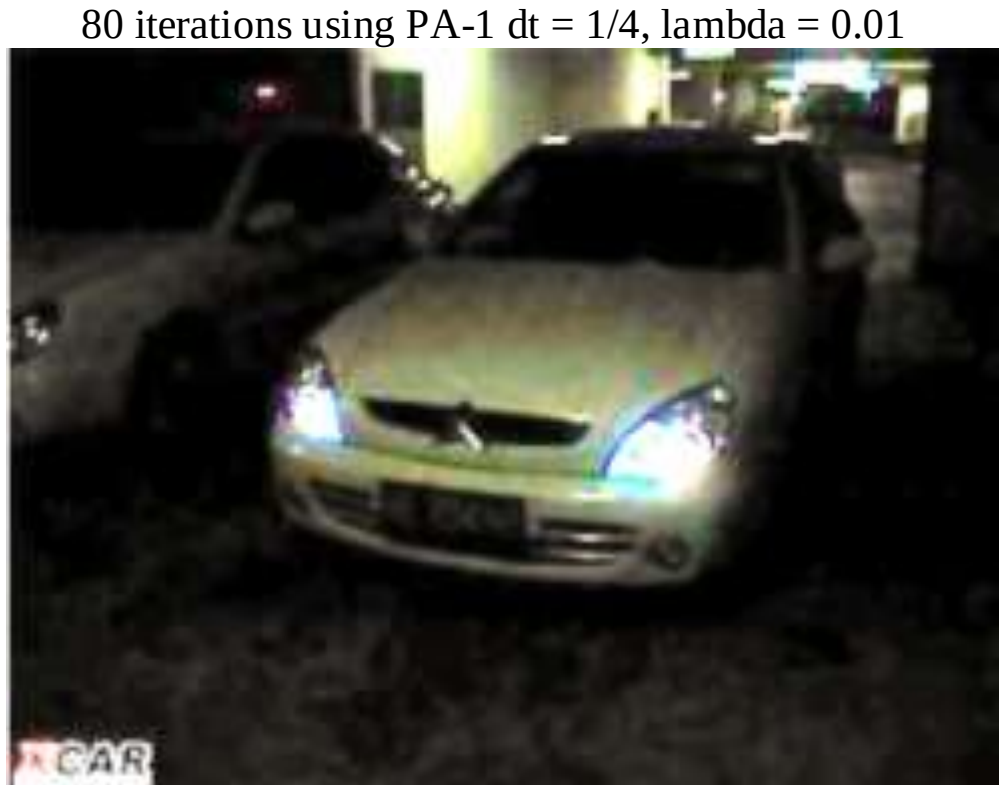

200 iterations using PA-2, dt = 1/4, lambda = 0.1

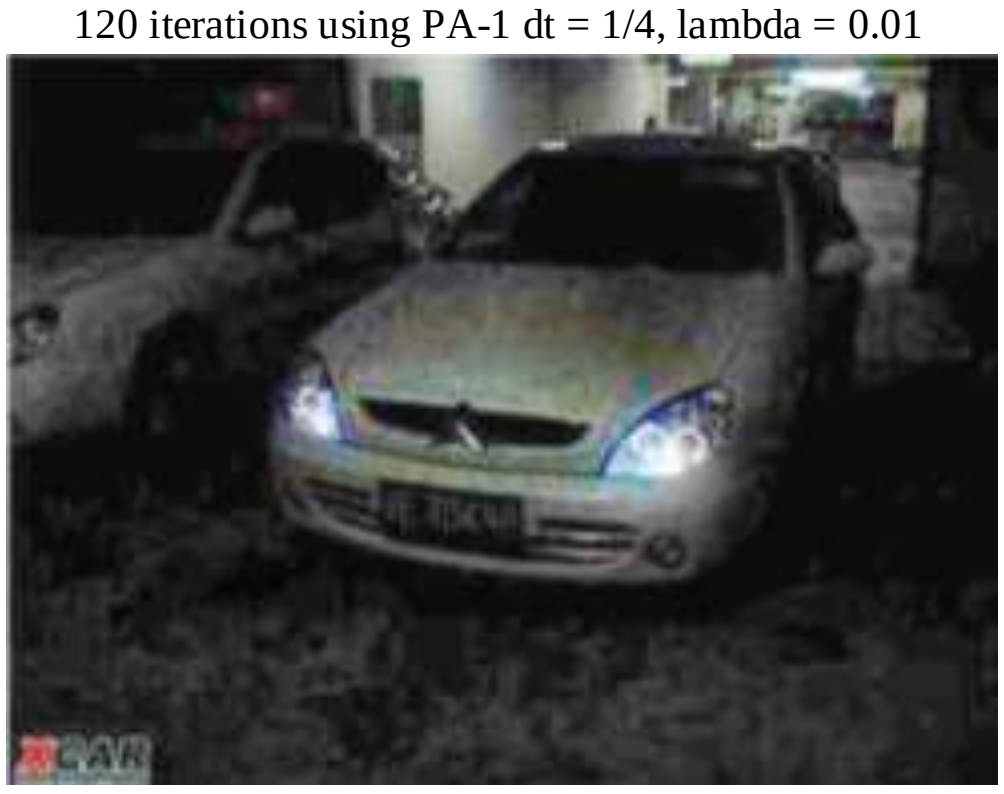

Using CLAHE

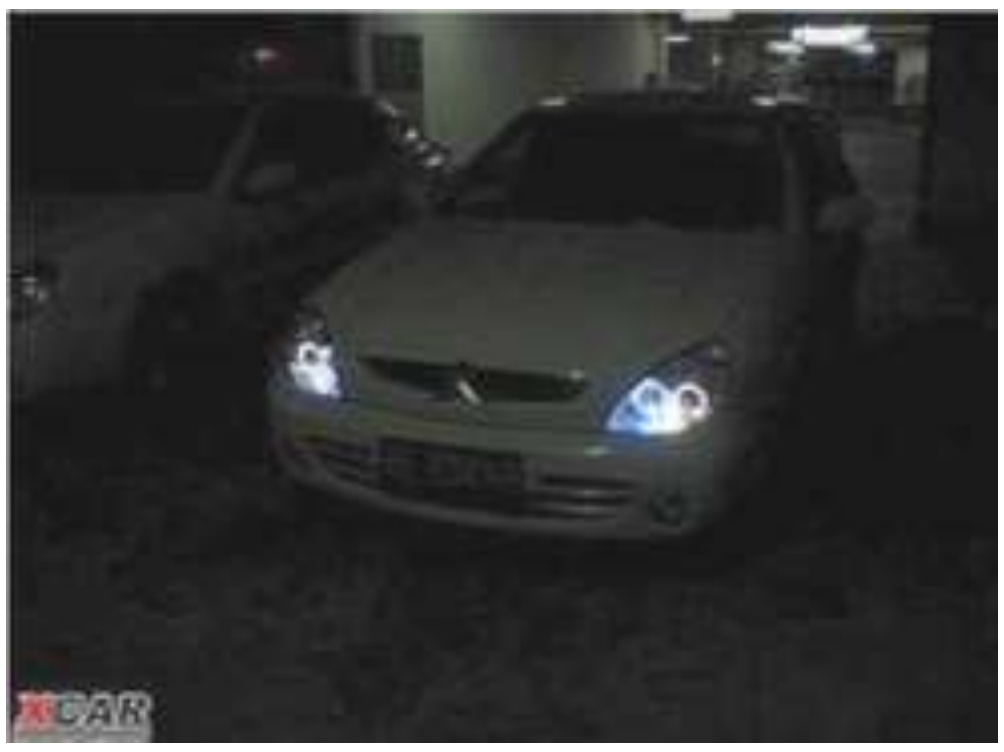

Using Frequency Domain Homomorphic filter

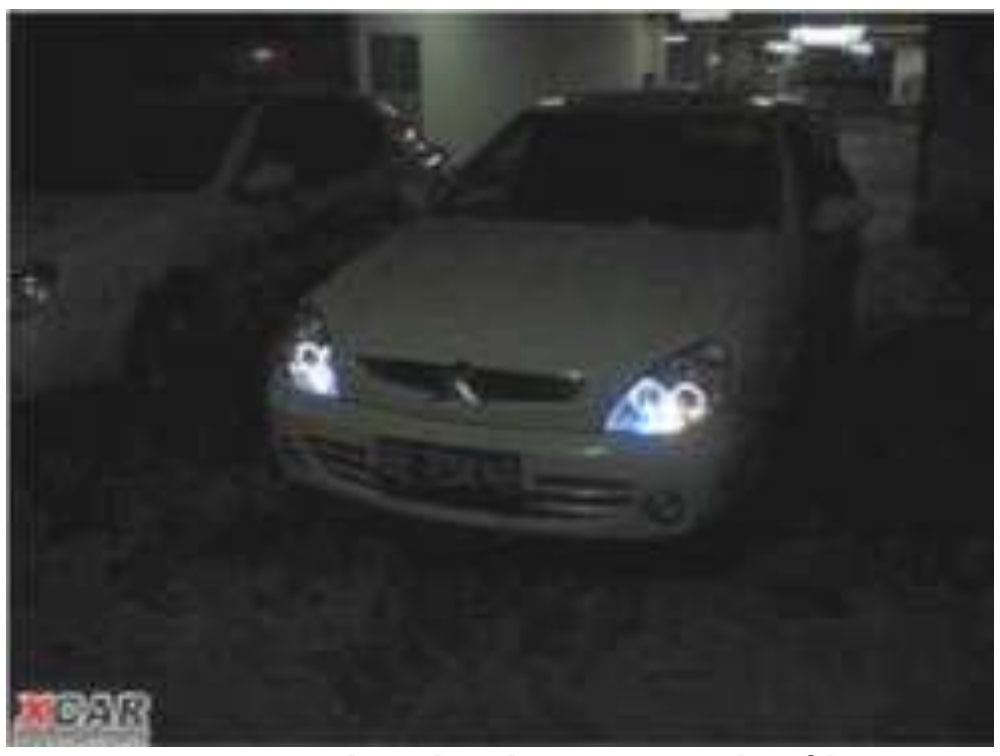

Using PA-1 combined with Homomorphic filter

Fig. 9A Natural dark image processed with various algorithms compared with the proposed approaches

Based on the results, the algorithm is very good for global contrast and colour enhancement but leads to whiting out of bright areas as evidenced in the saturation of the head lamp region in the car image. Additionally, the results are influenced by the amount of contribution allowed by the user and the number of iterations required. However its illumination correction potential is clearly observed and with more control over the contrast term, highlights are better managed. Note the increase in compression artifacts in the enhanced images with reduced contribution from the Ansiotropic Diffusion term.

**4.1.1 Illumination correction and colour enhancement**

Another point to note is the colour enhancement or retention observed in poorly illuminated images processed with PA-1 and PA-2 compared with the CLAHE and RGB Homomorphic filter methods. To better appreciate this we select another image to provide another instance of this ability of the proposed algorithms. Fig. 9B shows the results of using PA-2 compared with CLAHE and Homomorphic filter.

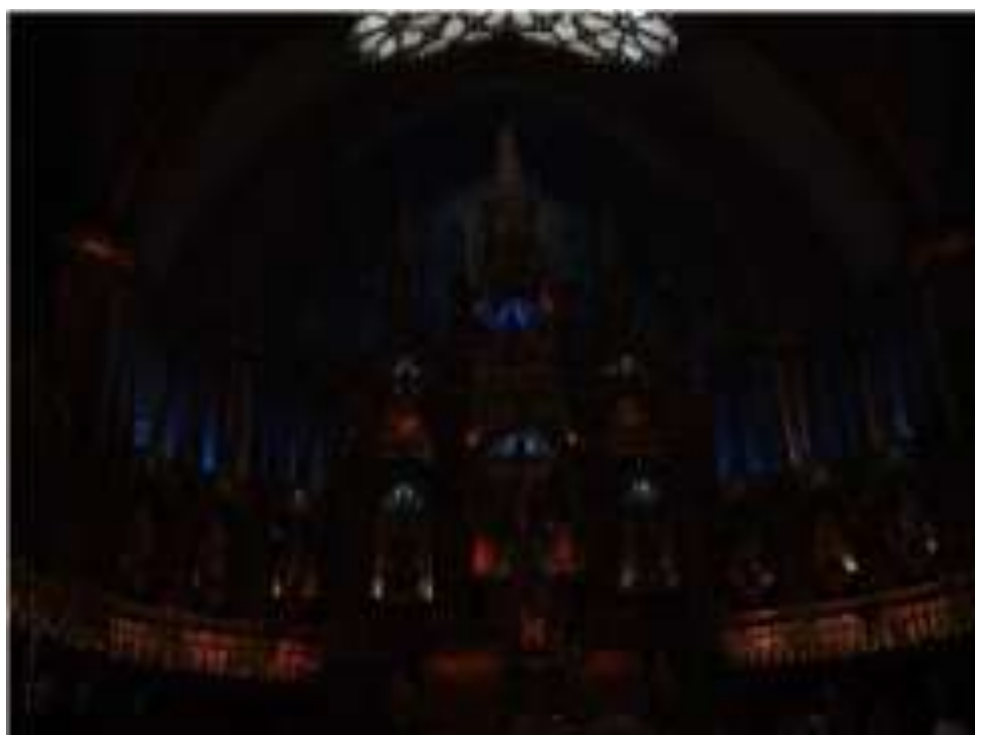

Original image

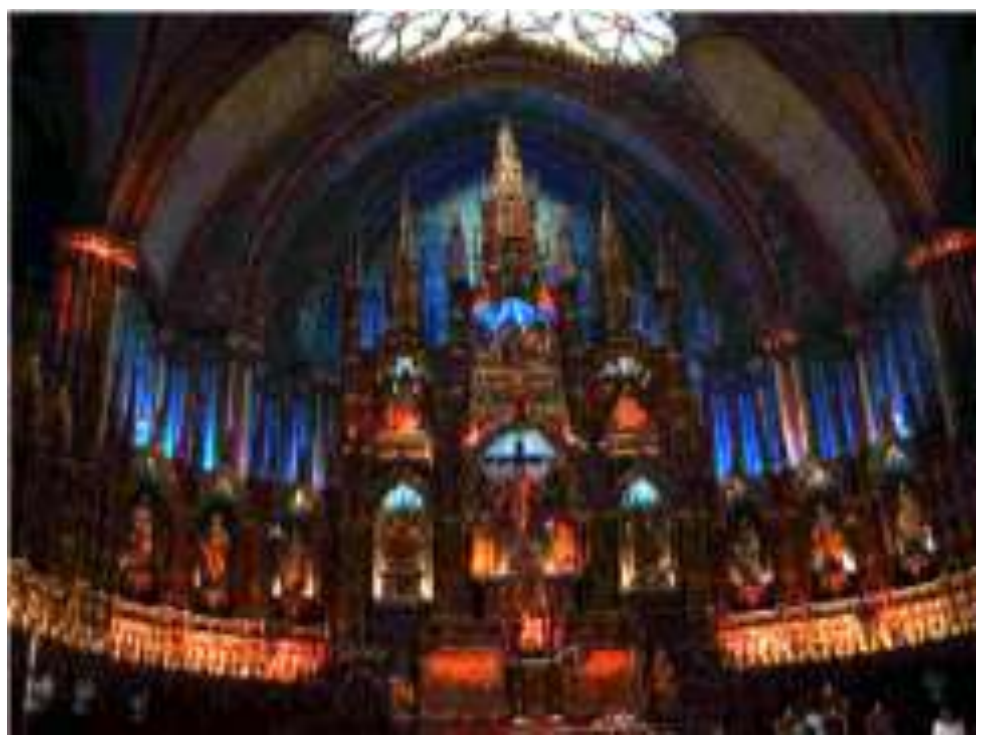

60 iterations using PA-1 dt = 1/4, lambda = 0.01

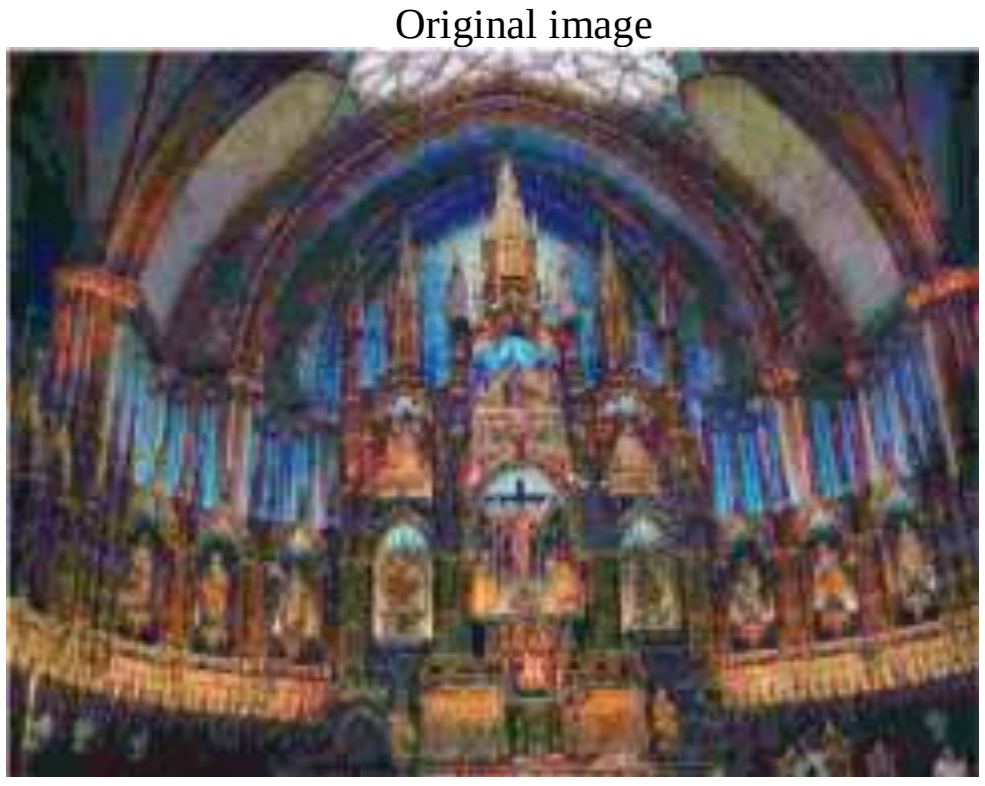

using CLAHE with optimum parameters

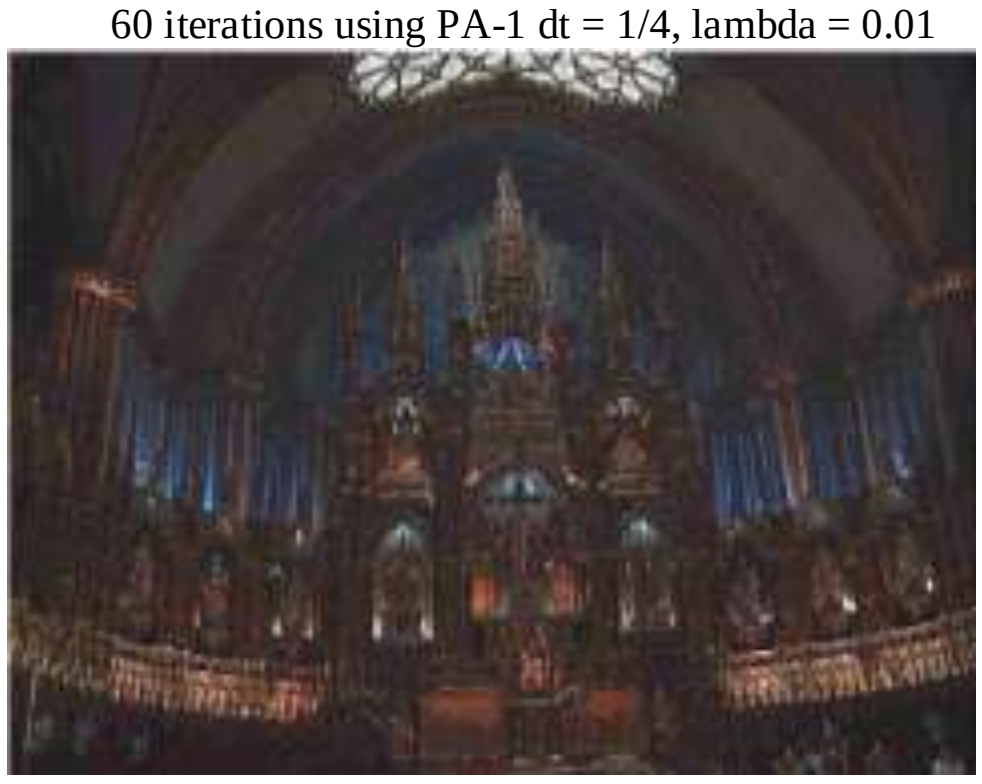

using Frequency Domain Homomorphic filter

**Fig. 9B** Natural dark image processed with various algorithms compared with the proposed approach

**4.2 Colour correction**

We subsequently test the algorithm for the ability to perform colour correction or restoration. We utilize an image that is degraded by excess blue hue. The results of the proposed algorithms (PA-1 and PA-2) are compare against the CLAHE and Homomorphic filters in Fig. 9C. The original image is also provided for comparison. Based on visual results, the output images produced by PA-1 and PA-2 are the closest in appearance to the original image. The CLAHE-based methods exhibit colour distortions in their outputs while the Homomorphic filter yields as sub-par image with distorted hue.

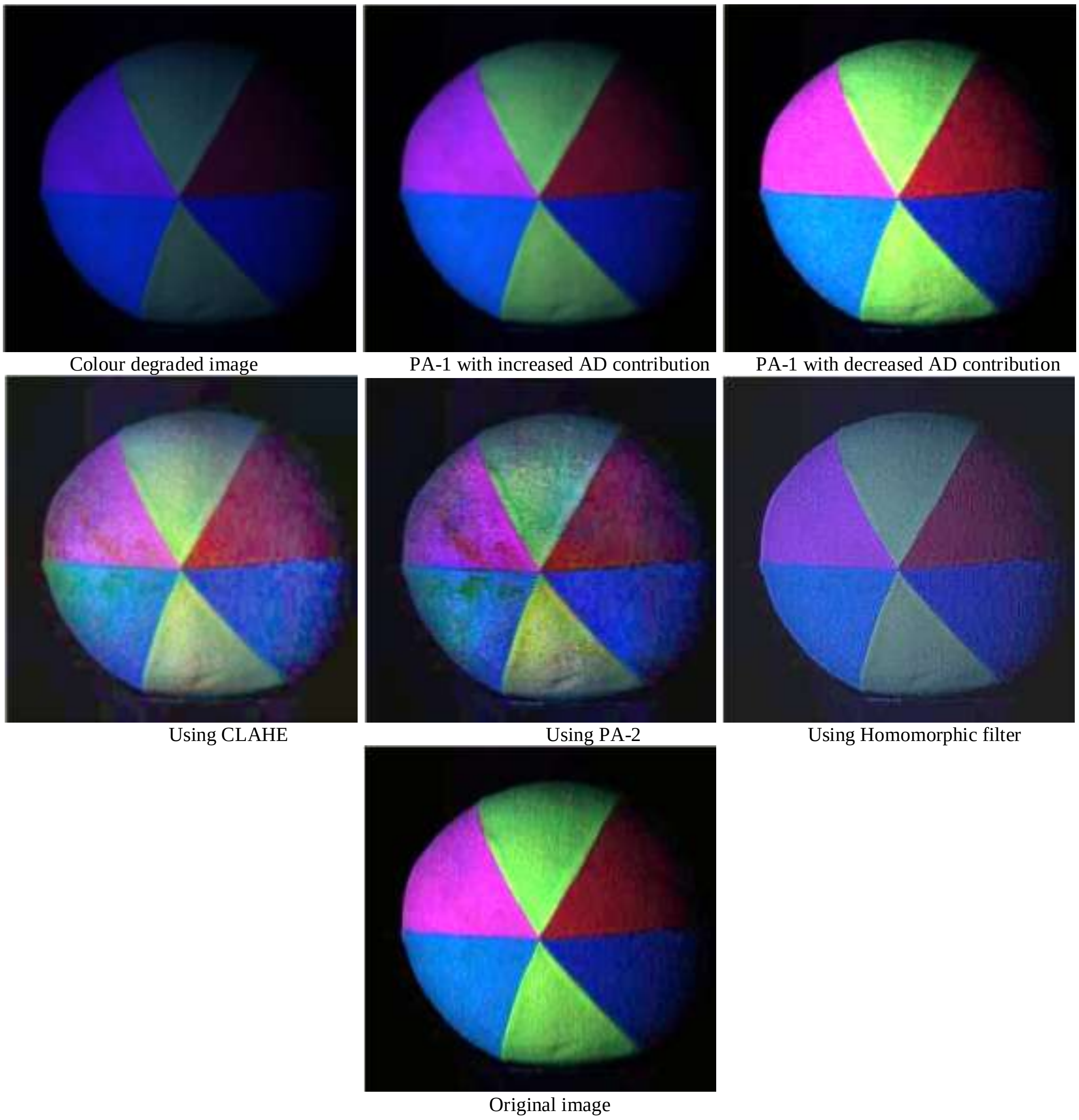

**Fig. 9C** Natural dark image with colour cast processed with various algorithms compared with the proposed approaches

### 4.3 Image smoothing, de-noising and contrast enhancement

This section deals with the image restoration and contrast enhancement aspect of PA-1. Based on the plots in Fig. 9D. The image is corrupted with Gaussian noise and is filtered with the Anisotropic Diffusion (AD) term of the algorithm. This feature is the standard AD algorithm for image denoising images corrupted with Gaussian noise. We also process the image using both the contrast and AD term in PA-1 and PA-2 (minimizing contribution from CLAHE term).

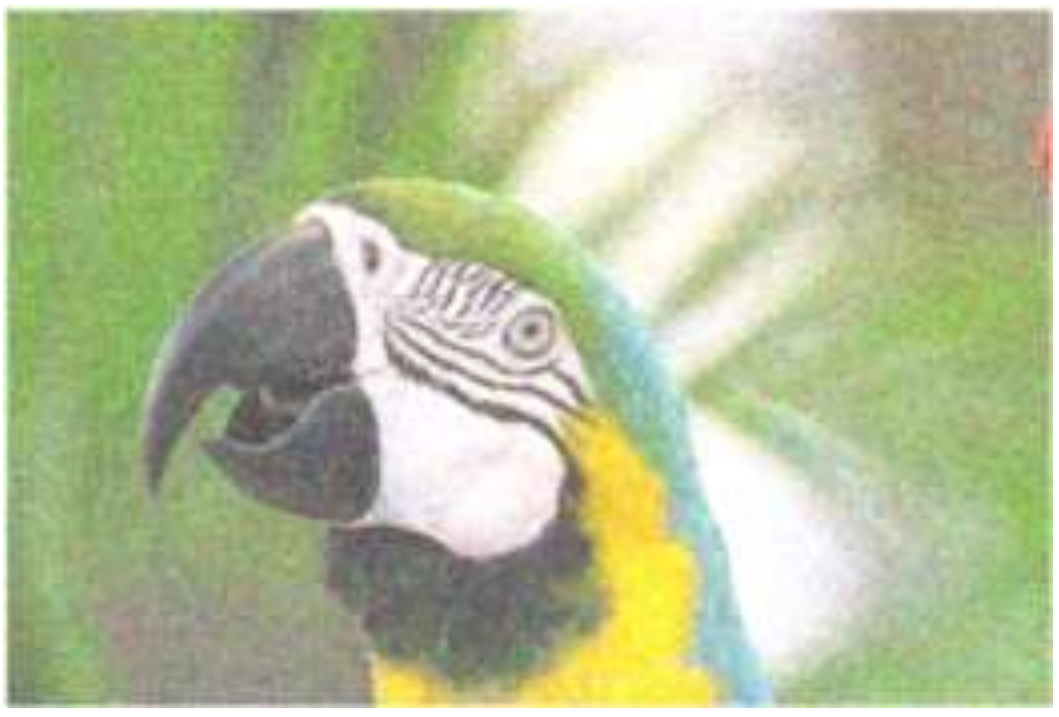
Image with Gaussian noise (variance = 0.25)

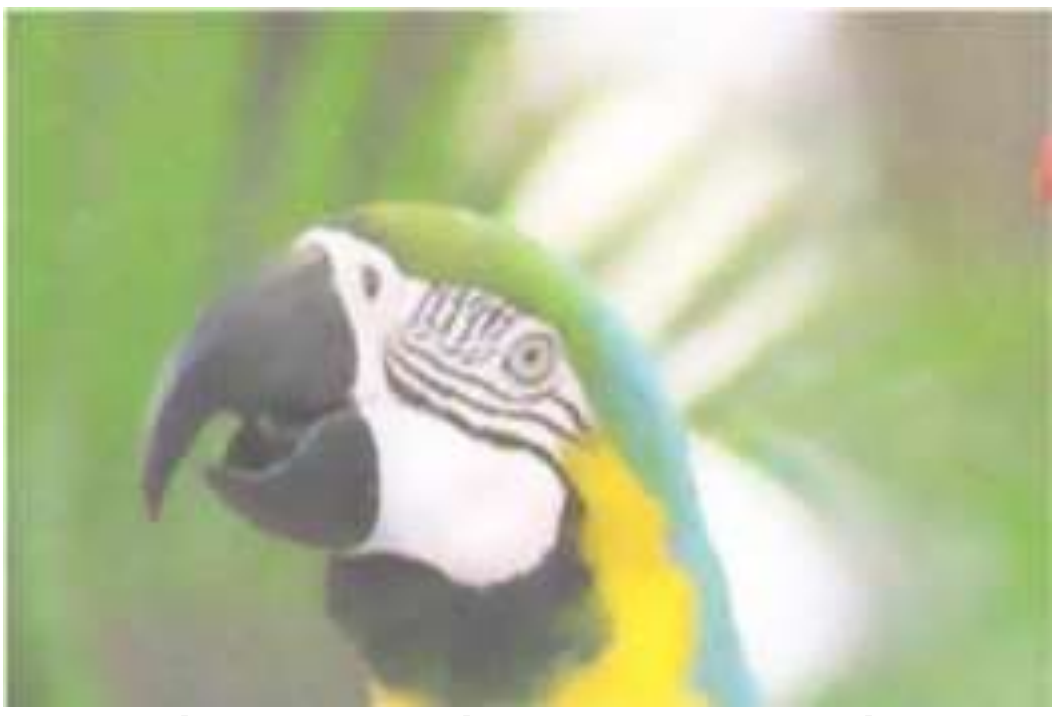
De-noised image using only AD term in PA-1 and PA-2

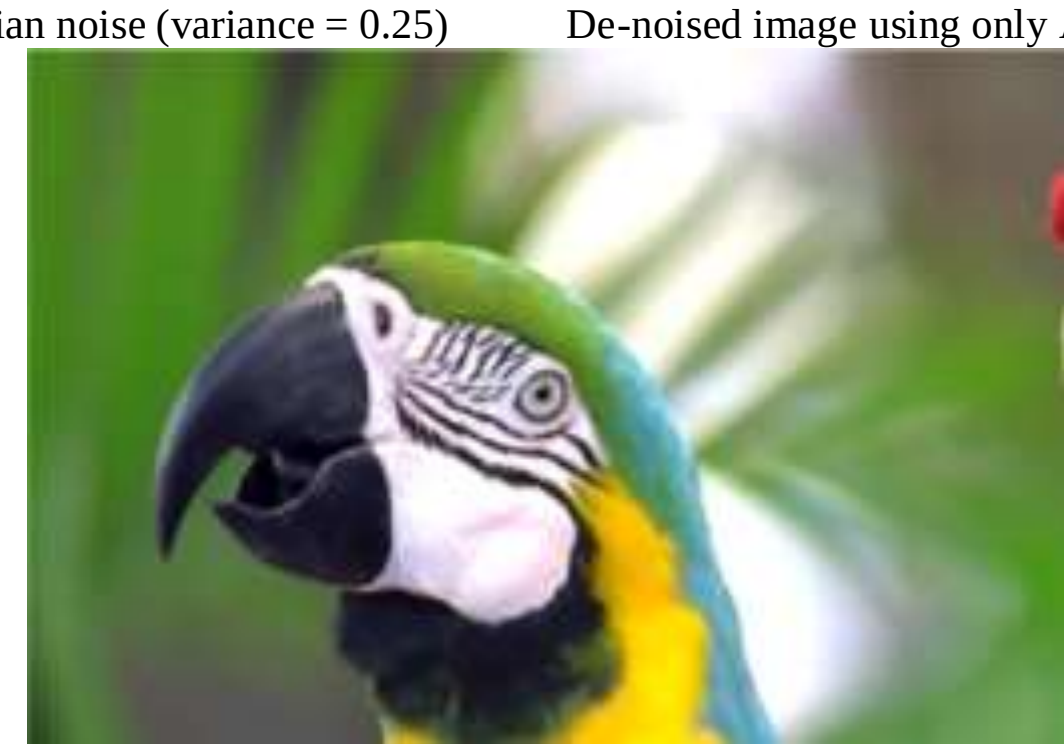
Colour and contrast enhancement using both AD and contrast term in PA-1 and PA-2

**Fig. 9D** Natural noisy image processed with various algorithms compared with the proposed approaches

**Conclusion**

This report has presented the theoretical formulation and detailed experimental verification of a group of algorithm based on PDEs proposed in the paper titled, “An image smoothing and enhancement algorithm for underwater images based on partial differential equations” [49].